\documentclass{article}

\usepackage{microtype}
\usepackage{graphicx}
\usepackage{subcaption}
\usepackage{booktabs} % for professional tables
\usepackage{xurl}
\usepackage{array}
\usepackage[svgnames]{xcolor}
\usepackage{float}
\usepackage{placeins}
\usepackage{listings}
\usepackage{tabularx}
\usepackage{multirow}
\usepackage{hyperref}

\usepackage[accepted]{icml2026}

\usepackage{amsmath}
\usepackage{amssymb}
\usepackage{mathtools}
\usepackage{amsthm}

\usepackage[capitalize,noabbrev]{cleveref}

\theoremstyle{plain}

\theoremstyle{definition}

\theoremstyle{remark}

\icmltitlerunning{Can Language Models Identify Shadow Trading Targets?}

\begin{document}

\twocolumn[
  \icmltitle{Can Language Models Identify Shadow Trading Targets? An NLP Evaluation of SEC Enforcement Theory}




  \icmlsetsymbol{equal}{*}

  \begin{icmlauthorlist}
    \icmlauthor{Sarah Wilson}{equal,Columbia}
    \icmlauthor{Michael MacKay}{equal,Columbia}
    \icmlauthor{Anthony Marello}{equal,Columbia}
    \icmlauthor{Trinav Bhattacharyya}{equal,Columbia}

  \end{icmlauthorlist}

  \icmlaffiliation{Columbia}{Columbia University}

  \icmlcorrespondingauthor{Sarah Wilson}{sw4104@columbia.edu}

  \icmlkeywords{Natural language processing, privacy, securities law, anonymity, insider trading, financial NLP, large language models, legal AI}

  \vskip 0.3in
]



\printAffiliationsAndNotice{\icmlEqualContribution\ --- \textit{Extended version of a poster presented at the AI for Law (AI4Law) Workshop, ICML 2026.}}

\begin{abstract}
  Shadow trading---trading in a peer firm's securities on the basis of material nonpublic information (MNPI) about an ``economically linked'' company---is a novel and contested theory of insider trading liability, first prosecuted in \textit{SEC v.\ Panuwat} (2023). Enforcing it requires identifying economically linked firms \textit{ex ante}, a determination the SEC makes only after the fact using mass market surveillance infrastructure. We ask whether NLP can do what the SEC's theory presumes insiders already know: identify peer firms ex ante from publicly mandated disclosures. Using a two-stage LLM pipeline applied to Item~7 (Management's Discussion and Analysis) sections of SEC 10-K filings, we score semantic similarity across 30 M\&A events spanning five industries and relate similarity to announcement-day abnormal stock returns. On the \textit{Panuwat} fact pattern itself the pipeline recovers Incyte among the closest peers, a useful sanity check on the one case with a known outcome. Across the full dataset, however, we find no association: pooling 217 peer observations, the within-event rank correlation between similarity and abnormal return is $+0.07$ (permutation $p = 0.37$), and the mean per-event Spearman correlation is $+0.05$ with a $95\%$ confidence interval of $[-0.08, +0.18]$---narrow enough to exclude any moderate relationship rather than merely failing to detect one. A case-level reading agrees: 14 of 30 events support the hypothesis, 12 contradict it, and 4 are ambiguous. We also find that Incyte fell outside the standard \$2B--\$10B mid-cap band on the day before the announcement, complicating the ``mid-cap oncology'' category the SEC invoked. These results are exploratory and bound to this pipeline, corpus, and return measure, but they put pressure on the empirical premise of shadow trading enforcement and bear on constitutional questions surrounding the SEC's financial surveillance infrastructure.
\end{abstract}

\section{Introduction}
\label{sec:intro}

On August 18, 2016, Matthew Panuwat, a business-development executive at Medivation, Inc., received an email from his CEO indicating that Pfizer would soon likely acquire the company. Seven minutes later, he purchased call options---not in Medivation, but in Incyte Corporation, a separate publicly traded biopharmaceutical company with no direct business relationship to Medivation. When the acquisition was announced the next day, Incyte's stock rose 7.7\%, and Panuwat sold his options for approximately \$107,000 in profit~\cite{panuwat2023}. 

The U.S.\ Securities and Exchange Commission (SEC) charged Panuwat with insider trading under a new theory it calls ``shadow trading'': that MNPI about Firm~A is material to any firm with which A is ``economically linked,'' such that trading in Firm~B's securities on the basis of such information constitutes securities fraud~\cite{panuwat2023}.

\paragraph{The enforcement asymmetry.}
Classical insider trading enforcement begins with a \textit{known event}: regulators detect abnormal volume in a stock, then subpoena the relevant broker's records to identify the trader. Shadow trading inverts this logic. The suspicious trade may occur in any security whose connection to the MNPI event is asserted only \textit{after} the fact. Regulators therefore cannot index on a single security; they must surveil trading activity \textit{across the entire market} to find which trades correlate with any given corporate event. Among the SEC's enforcement programs, this is unique: flash-crash reconstruction starts from a known event, classical and misappropriation insider trading start from a tagged trade in a particular security, but shadow trading is the only use case for which the SEC must search the corpus \textit{before} it can identify which trader to investigate. That is precisely what the Consolidated Audit Trail (CAT) enables: a comprehensive, queryable record of every order, cancellation, and execution in U.S.\ equities and options markets, covering more than 100 million investors~\cite{davidson2024}. Multiple legal challenges have characterized CAT as a ``dangerous dog''~\cite{peirce2019} and a potential Fourth Amendment violation~\cite{davidson2024,carpenter2018}; this paper's empirical results bear on those challenges directly.

\paragraph{The NLP question.}
If economic linkage can be identified \textit{ex ante} from public disclosures, targeted watchlists could replace mass surveillance. Because the SEC requires firms to disclose their risk-reward profiles in annual reports, Item~7 (Management's Discussion and Analysis) of the Form 10-K is an ideal corpus for testing this hypothesis: if two firms genuinely occupy the same strategic niche, their MD\&A sections should exhibit high semantic similarity, and if the shadow trading hypothesis is empirically grounded, those same firms should experience positive abnormal returns when an acquisition of one is announced. We operationalize this test using a two-stage LLM pipeline: a peer discovery stage that identifies candidate companies from filings in the SEC's EDGAR database, followed by a document-grounded similarity ranking stage that assigns semantic similarity scores. We then compute announcement-day abnormal returns for each peer firm and assess whether textual similarity predicts peer stock reactions across 30 acquisition events in five industries.

\paragraph{Contributions.}
\begin{enumerate}
    \item A fully documented two-stage LLM pipeline for long-context multi-document financial similarity analysis, including principled handling of Foreign Private Issuers (FPIs). The prompts, rubric, and per-event outputs are reproduced in full in the Appendix; because the underlying model is closed-weight, exact score reproduction is not guaranteed (Section~\ref{sec:limitations}).
    \item A dataset of 30 M\&A events (2013--2023) with LLM-derived similarity scores and manually computed abnormal returns, spanning biopharmaceuticals, consumer staples, technology, finance, and automotive industries.
    \item Empirical evidence that, for this pipeline and corpus, textual similarity in SEC filings does not predict announcement-day peer stock reactions: a pooled within-event rank correlation of $+0.07$ ($p = 0.37$) and a mean per-event Spearman correlation of $+0.05$, $95\%$ CI $[-0.08, +0.18]$, with $14/30$ events supportive under a case-level reading. Cross-industry variation is pronounced---Technology yields zero supporting cases across five acquisitions while Automotives yields zero contradictions---though with five events per sector we treat this as descriptive rather than established.
    \item A correction to the record on the \textit{Panuwat} fact pattern: Incyte's capitalization on the day before the announcement was \$14.3B, outside the standard \$2B--\$10B mid-cap band that the SEC's ``mid-cap oncology'' framing presupposes.
\end{enumerate}

This work addresses a high-stakes legal enforcement question where the answer---whether language can serve as a legal standard---has direct constitutional consequences for the largest government-mandated collection of personal financial data in U.S.\ history.

\section{Background}
\label{sec:background}

\subsection{Shadow Trading and \textit{SEC v.\ Panuwat}}
\label{sec:panuwat}

Under the classical theory of insider trading, liability attaches to a relationship between the trader and the \textit{issuer} of the traded security. \textit{United States v.\ O'Hagan}~\citep{ohagan1997} extended this to misappropriation: a person who trades on MNPI obtained from a source owed a duty of trust and confidence commits securities fraud even if they have no relationship with the issuer. \textit{SEC v.\ Panuwat}~\citep{panuwat2023} pushes this further still: the duty-source need not be connected to the security traded at all. Because Panuwat owed a duty to Medivation, and because the court found that Medivation and Incyte were ``economically linked'' mid-cap oncology companies, his MNPI about Medivation was deemed material to Incyte's securities. Table~\ref{tab:timeline} shows the case timeline.

\begin{table}[t]
    \small
    \centering
    \caption{Timeline of \textit{SEC v.\ Panuwat}.}
    \begin{tabular}{@{}lp{5.0cm}@{}}
    \toprule
    \textbf{Date} & \textbf{Event} \\
    \midrule
    Aug.\ 2016 & Panuwat trades Incyte options, 7 minutes after CEO email \\
    Aug.\ 2021  & SEC files civil complaint \\
    Jan.\ 2022  & Motion to dismiss denied \\
    Nov. 2023        & Motion for summary judgment denied \\
    Apr.\ 2024  & Jury finds Panuwat liable after 8-day trial \\
    Sep.\ 2024  & Civil penalty of \$321,197 imposed \\
    Nov. 2024   & Panuwat appeals (9th Cir.\ No.\ 24-6882) \\
    May 2025    & U.S.\ Chamber of Commerce files amicus brief \\
    Jun.\ 2026  & Ninth Circuit hears oral argument \\
    Aug.\ 2026  & Panel decision \textbf{pending} \\
    \bottomrule
    \end{tabular}
    \label{tab:timeline}
\end{table}

The key doctrinal problem is that neither firm mentioned the other in its annual report, and neither shared a direct business relationship. ``Economic linkage'' was established retroactively by the SEC's expert witness, who testified that Incyte's 7.7\% announcement-day return ``was not caused by normal fluctuations.'' The category itself---``mid-cap oncology''---is not a recognized industry classification, and the record fixes no capitalization threshold for it. The opinion describes Medivation as a mid-cap, oncology-focused biopharmaceutical company and returns repeatedly to the scarcity of mid-cap oncology assets, but nowhere states what capitalization range the term denotes~\citep{panuwat2023}. The label does substantive work---it is what makes Incyte comparable enough for MNPI about Medivation to be material to Incyte's securities---while remaining numerically undefined. The economic stakes are large: \citet{augustin2019} document positive abnormal options volume in approximately 25\% of U.S.\ takeovers between 1996 and 2012, and \citet{mehta2021} estimate single-event shadow-trading profits ranging from \$139{,}400 to \$678{,}000. Since \textit{Panuwat}, the only concluded enforcement action has been \textit{Bechtolsheim}, which settled for a \$923{,}740 civil penalty~\citep{bechtolsheim2024}; even so, roughly one in five public companies has now formally prohibited shadow trading in its insider-trading policy~\citep{corporatecounsel2025}. 

\subsection{The Consolidated Audit Trail and Fair Notice}
\label{sec:cat}

Because shadow trading liability may attach to any ``economically linked'' security, the theory requires whole-market retrospective surveillance rather than targeted investigation. The Consolidated Audit Trail (CAT)---created in 2012 and covering every equity and options order across all U.S.\ securities exchanges~\citep{secrule613}---is the infrastructure that makes this possible. Shadow trading is unique among the SEC's enforcement use cases in requiring this inversion; other programs proceed from a known event or a flagged trade in a particular security. Shadow trading thus transforms CAT from a regulatory tool into what critics call a ``digital general warrant,'' and raises a direct fair notice problem: for a regulatory prohibition to withstand due process scrutiny, a person of ordinary intelligence must be able to determine what conduct is prohibited before the fact~\citep{sessionsdimaya}. If the set of prohibited securities cannot be identified from public disclosures ex ante, an insider has no principled way to know which trades are off-limits. This paper tests that empirical premise directly.

\subsection{Text-Based Firm Similarity in Financial Disclosure} 
\label{sec:tnic} 
The SEC's ``economic linkage'' standard presupposes that similarity between firms is readable from their public disclosures. The foundational evidence for this comes from the Hoberg-Phillips Text-based Network Industry Classification (TNIC)~\citep{hobergphillips2016}, which demonstrates that pairwise firm similarity constructed from mandatory 10-K product descriptions captures competitive dynamics far better than static SIC codes. If TNIC-style analysis can classify industries, it should, in principle, identify the economically linked peers that shadow trading liability requires. 

However, TNIC relies on bag-of-words similarity over Item~1 (Business Description), which is semantically blind to synonymous terminology: ``oncology'' and ``cancer treatment'' would be treated as unrelated tokens. This creates the \textit{administrability gap}---if two firms describe the same strategic niche using different technical vocabularies, BoW similarity fails to detect the linkage that shadow trading liability presupposes.

Learned models have been applied to paired financial-report comparison before: \citet{koval2024comparing} train encoders to compare 10-K pairs for forecasting, establishing that document-scale report comparison is not itself novel. What long-context generative LLMs add is the ability to ingest complete Item~7 sections without chunking and to bridge vocabulary differences explicitly, applied here to the forward-looking MD\&A narrative rather than the backward-looking Item~1. Our contribution is not the comparison architecture but the use of it to operationalize a legal standard and test that standard against market outcomes. Whether closing the administrability gap is sufficient to make economic linkage reliably identifiable ex ante is the empirical question this paper addresses.

\section{Methodology}
\label{sec:methodology}

\subsection{Dataset Construction}
\label{sec:data}

We assembled 30 major public acquisitions across five industries: biopharmaceuticals (10 events), consumer staples (5), technology (5), finance (5), and automotive (5), spanning 2013--2023. Biopharmaceutical events are the primary test set, as shadow trading doctrine originated there; the remaining industries serve as a control group. Candidate events were generated by prompting Gemini~3.1~Pro for acquisitions structurally comparable to Pfizer/Medivation in target size and deal structure. From that pool, retained events met four criteria: the target was a U.S.-listed public company with an EDGAR filing history; the deal was an announced acquisition of the whole company rather than a divestiture or minority stake; the announcement date could be fixed unambiguously; and the sector had enough comparable public peers to populate a candidate list. This universe was not pre-registered and was not drawn by a mechanical screen from a deal database, so selection effects cannot be ruled out. Additionally, the candidate pool reflects which transactions are most prominent in the model's training data, which compounds the contamination concern discussed in Section~\ref{sec:limitations}. Events were fixed before abnormal returns were computed. For each acquisition target (Firm~A), we retrieved the most recent Form 10-K filed on EDGAR prior to the announcement date, using Item~7 (MD\&A) and Item~7A as the primary similarity corpus. For Foreign Private Issuers (FPIs) filing Form 20-F, we substitute with Item~5 (``Operating and Financial Review and Prospects'') and Item~11 (``Quantitative and Qualitative Disclosures About Market Risks'') accordingly.

\subsection{Two-Stage LLM Pipeline}
\label{sec:pipeline}

\begin{figure}[t]
    \centering
    \includegraphics[width=1\linewidth]{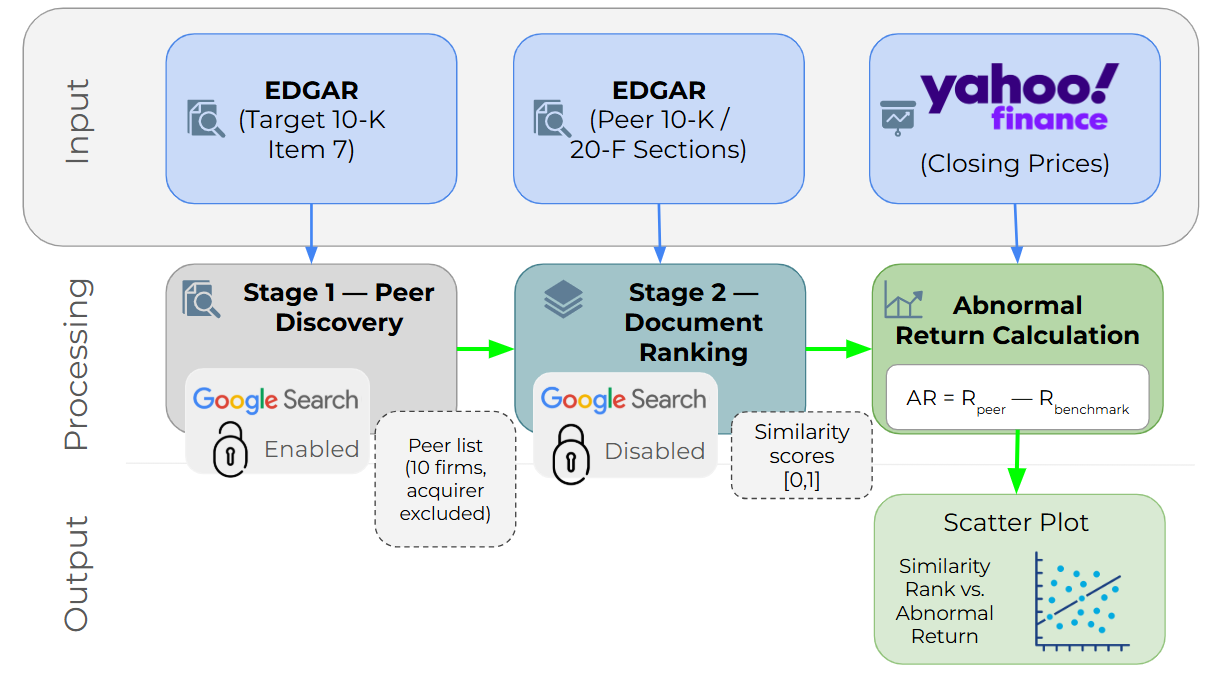}
    \caption{Augmented Hoberg-Phillips Framework}
    \label{fig:pipeline}
\end{figure}

Figure~\ref{fig:pipeline} illustrates our \textbf{Augmented Hoberg-Phillips (HP) framework}, which extends the classical HP bag-of-words approach~\cite{hobergphillips2016} by substituting a long-context LLM for cosine similarity over binary term vectors, and shifting from Item~1 (Business Description) to Item~7 (MD\&A) to better reflect the forward-looking narrative an insider would possess. We decompose the task into two sequential stages to avoid self-consistency bias: models anchored to prior numerical scores resist revision even when new evidence warrants it~\citep{huang2024largelanguagemodelsselfcorrect, harshavardhan2026selfanchoringcalibrationdriftlarge, khot2023decomposedpromptingmodularapproach}. Both stages are implemented in Google AI Studio using Gemini~3.1~Pro in isolated per-event sessions; full model configuration details are provided in Section~\ref{sec:model}.

\paragraph{Stage 1 --- Peer Discovery.}
The target firm's Item~7 is uploaded to identify ten peer companies as of the acquisition date. Google Search grounding is enabled to confirm that candidate firms were publicly traded on a U.S.\ exchange and had an active filing obligation at the time, and specified in the prompt. Section~\ref{sec:limitations} discusses how this bounds the ``identifiable from public filings'' framing. The model provides natural language dimensional justifications (Table~\ref{tab:searchrubric}) without numerical scores to mitigate anchoring bias. Key constraints include: no hindsight (market state at announcement only); reasonable insider standard; acquirer excluded; active 10-K/20-F filing obligation; and NYSE/NASDAQ trading status required. Full prompts are in Appendix~\ref{app:prompts}.

\begin{table}[h!]
    \centering
    \small
    \renewcommand{\arraystretch}{1.2}
    \caption{Stage 1 Dimensional Justification Rubric}
    \begin{tabular}{@{} p{2.2cm} p{1.8cm} p{3cm} @{}}
        \toprule
        \textbf{Dimension} & \textbf{Focus} & \textbf{Assessment Criteria} \\
        \midrule
        \textbf{Tech Overlap} & Product & Similarity of platforms or pipelines. \\
        \textbf{Strategic}    & Financial & Matching growth stages and revenue models. \\
        \textbf{Risk}         & External & Geographic exposure and headwinds. \\
        \bottomrule
    \end{tabular}
    \label{tab:searchrubric}
\end{table}

\paragraph{Stage 2 --- Document-Grounded Ranking.}
Item~7 sections for all identified peers are uploaded as a single multi-document prompt. The model assigns scores in $[0,1]$ based exclusively on the provided disclosures---mirroring the closed-corpus logic of HP~\citep{hobergphillips2016}---with Google Search disabled. Outputs are parsed into structured \texttt{.csv} tables. We weight Technical/Product Overlap most heavily (40\%) to reflect the SEC's emphasis on shared therapeutic mechanisms in \textit{Panuwat}; Financial (Strategic Stage) and Risk \& Market (Exposure) each receive 30\%, capturing the forward-looking narrative dimensions unique to Item~7 relative to Item~1.

\begin{table}[h!]
    \centering
    \small
    \caption{Stage 2 Similarity Score Rubric}
    \renewcommand{\arraystretch}{1.2}
    \begin{tabular}{@{} p{2.5cm} c p{3.5cm} @{}}
        \toprule
        \textbf{Dimension} & \textbf{Weight} & \textbf{Assessment Criteria} \\
        \midrule
        \textbf{Technical}   & 40\% & Core technology and product engines. \\
        \textbf{Financial}   & 30\% & Growth trajectories and margin profiles. \\
        \textbf{Risk/Market} & 30\% & Geographic focus and specific headwinds. \\
        \bottomrule
    \end{tabular}
    \label{tab:comparerubric}
\end{table}

\subsection{Model and Configuration}
\label{sec:model}

We select Gemini 3.1 Pro~\citep{gemini31pro} for its 1M-token context window (sufficient to ingest multiple complete 10-K sections without chunking), robust multi-document cross-referencing, and state-of-the-art reasoning performance (77.1\% on ARC-AGI-2; 94.3\% on GPQA Diamond). These capabilities matter because traditional bag-of-words approaches are semantically blind, creating the \textit{administrability gap} our study seeks to measure~\citep{chen-sarkar-2020-semantic}. The pipeline runs in Google AI Studio with Thinking Level set to High, enabling Chain-of-Thought reasoning over dense SEC disclosures~\citep{wei2023chainofthoughtpromptingelicitsreasoning}. Each acquisition event runs in an isolated session to prevent cross-contamination. Top-P is slightly lowered from 0.95 to 0.90 to reduce uncertain token probability mass while preserving domain-specific financial vocabulary~\citep{li2024dawndarkempiricalstudy}; all other parameters remain at default.

\subsection{Abnormal Return Calculation}
\label{sec:ar}

For each peer firm, we compute the announcement-day abnormal return:
\begin{equation}
    AR_i = R_i^{(0)} - R_{\text{benchmark}}^{(0)}
\end{equation}
where $R_i^{(0)} = (P_i^{(0)} - P_i^{(-1)}) / P_i^{(-1)}$ is the peer firm's raw return on announcement day (Day~0) relative to the prior trading day (Day~$-1$), and $R_{\text{benchmark}}^{(0)}$ is the same-day return of a sector-level ETF: XBI (biopharmaceuticals), IYK (consumer staples), XLK (technology), XLF (finance), and IYC (automotive). This directly mirrors the SEC economist's methodology at the \textit{Panuwat} trial. We compute the Spearman rank correlation ($\rho$) between the similarity score and $AR_i$ per event (Section~\ref{sec:spearman}) and generate industry-specific scatter plots (Appendices~\ref{app:cases} and~\ref{app:plots}).

\paragraph{Mid-cap validation.} We computed the market capitalization of every identified peer as of the day before its acquisition announcement, using contemporaneous share counts wherever they were retrievable.\footnote{Of 318 peer-event rows, 236 yielded a usable capitalization---209 with a contemporaneous share count and 27 falling back to the current count. Of the rest, 72 firms had been delisted with no retrievable price history, 9 traded under symbols since reassigned to unrelated companies, and 1 lacked share data. Because delisted firms skew small, the surviving sample is biased toward larger capitalizations, which makes the mid-cap shares reported here upper bounds. Prices are split-adjusted but not dividend-adjusted, so that price and share count are expressed on the same basis.} Only $30.9\%$ of peers fall within the generic \$2B--\$10B mid-cap band, and $30.9\%$ within S\&P's eligibility range for additions to the MidCap~400 as it stood at the time of the \textit{Panuwat} trade, approximately \$1.4B--\$5.9B~\citep{spdji2026methodology}. The two bands select overlapping but not identical sets of firms; that they capture the same share is coincidental, and the point is simply that the conclusion does not turn on which mainstream definition of ``mid-cap'' one adopts. Within the biopharmaceutical subset the shares are lower still, at $21.7\%$ and $24.6\%$ respectively.

This undercuts the ``mid-cap'' label as an ex ante constraint in two ways. First, the band captures a minority of the peer universe, so it does little to narrow which firms an insider would know to trade. Second, and more directly, \textit{Incyte itself falls outside both definitions}: on the day before the Medivation announcement its capitalization was \$14.3B---above the \$2B--\$10B band, and more than twice the \$5.9B ceiling S\&P then applied to MidCap~400 additions. ``Mid-cap oncology'' is not a recognized industry classification, and under neither mid-cap convention in force in 2016 does it contain the security actually traded. The two capitalization figures that do appear in the record point the same way. An investment bankers' chart grouped Medivation and Incyte among a small set of commercial oncology companies spanning roughly \$5B to \$75B in market capitalization---a range running from mid-cap through mega-cap---and the opinion elsewhere places the pair in the \$10 billion range while describing them as mid-cap assets~\citep{panuwat2023}. A \$10B-range firm sits at the very ceiling of the generic band and well above S\&P's 2016 threshold; neither figure in the record is consistent with mid-cap as the term was conventionally used at the time. Because the record fixes no threshold of its own (Section~\ref{sec:panuwat}), the classification also turns on which vintage of an external definition one applies: S\&P revises its eligibility ranges quarterly, and by 2026 the MidCap~400 addition band had risen to \$8B--\$22.7B---a range that \textit{would} admit a \$14.3B firm. Whether a 2016 trade fell inside the category must be assessed against the 2016 bands; a threshold drawn from a later period cannot establish what an insider could have known at the time. Whatever work the label does in characterizing the link between Medivation and Incyte, it is not the work of putting an insider on notice ex ante.

\paragraph{Reproducibility.} All prompts, per-event CSV outputs, and scatter plots are provided in the Appendix. Abnormal return calculations are implemented in Python using \texttt{yfinance} and \texttt{pandas}, with Stooq as a fallback for delisted tickers. Because Gemini~3.1~Pro is a closed-weight model, exact score reproduction is not guaranteed across API versions; we therefore treat LLM outputs as structured qualitative judgments and compute all financial metrics independently.

\section{Results}
\label{sec:results}

\subsection{Reference Case: Pfizer/Medivation (2016)}
\label{sec:groundtruth}

\begin{figure}[t]
  \centering
  \includegraphics[width=1\linewidth]{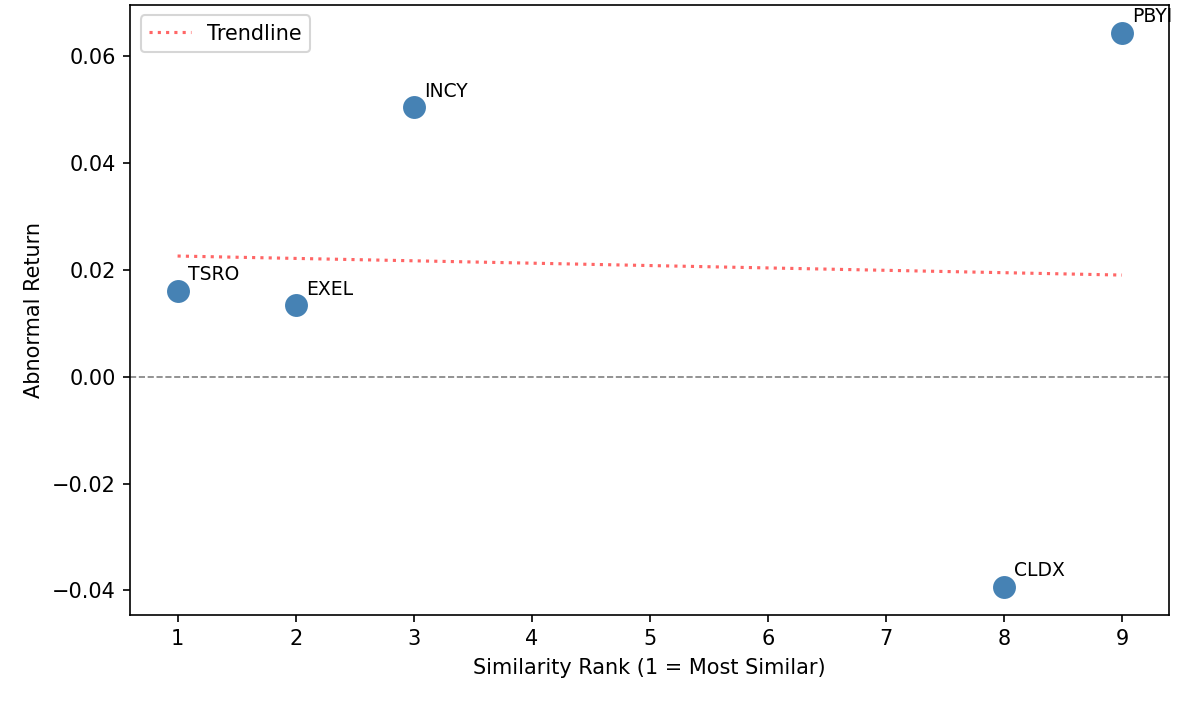}
  \caption{Pfizer/Medivation (2016): all three top-ranked peers (TSRO, EXEL, INCY) posted positive abnormal returns, which is what makes this case read as supportive under the procedure of Section~\ref{sec:fullresults}. The fitted trendline is negative but essentially flat---it falls $0.36$ percentage points across the full rank range ($R^2 = 0.002$)---and the rank correlation over the same peers is slightly negative ($\rho = -0.10$), driven by the ninth-ranked outlier PBYI. Neither statistic supports the hypothesis on its own.}
  \label{fig:pfizmed}
\end{figure}

We treat the Pfizer/Medivation acquisition---the factual basis of \textit{SEC v.\ Panuwat}---as a reference case, the one event in our dataset with a known legal outcome against which to sanity-check the pipeline. Figure~\ref{fig:pfizmed} shows the scatter plot of similarity rank versus abnormal return. The pipeline identifies Tesaro (TSRO, Rank~1), Exelixis (EXEL, Rank~2), and Incyte (INCY, Rank~3) as the most similar peers; all three show positive abnormal returns ($+1.60\%$, $+1.33\%$, and $+5.04\%$ respectively). Incyte---the stock Panuwat actually traded---is ranked third and posts the largest abnormal return \textit{among the top three ranked peers}. It is not the largest in the event overall: Puma Biotechnology (PBYI), ranked ninth, returned $+6.44\%$.

Three caveats bound what this case can establish. Incyte is recovered but is not top-ranked; the per-event rank correlation is in fact slightly negative ($\rho = -0.10$, Section~\ref{sec:spearman}) because of the PBYI outlier; and \textit{SEC v.\ Panuwat} has been widely reported since 2021, so a model with parametric knowledge of the case may associate Medivation with Incyte for reasons unrelated to the filing text we supply. We therefore read this case as a sanity check that the pipeline surfaces plausible oncology peers, not as validation of the methodology.

\subsection{Full Results: 30 Acquisition Events}
\label{sec:fullresults}

Table \ref{tab:results} summarizes all 30 events. Our interpretation of whether a case supports or contradicts the SEC's theory is not a mechanical byproduct of the OLS trendline alone. We applied the following decision procedure to each event: (i) compute the OLS trendline of abnormal return on similarity rank; (ii) inspect the signs of the abnormal returns for the three highest-ranked peers; (iii) where (i) and (ii) disagree, or where the trendline is driven by a single peer, weigh the absolute scale of the returns and the share of peers moving in each direction. An event is labeled \textit{supports} when the trend and the top-ranked reactions both point in the direction the shadow trading hypothesis predicts, \textit{contradicts} when they both point against it, and \textit{mixed} when they conflict and no reading dominates.

We state plainly that step (iii) involves judgment. The labels were assigned by the authors jointly, after inspecting the return data rather than under a rule fixed in advance, so they carry researcher degrees of freedom that a pre-registered rule would not. This is precisely why we do not rest the paper's quantitative claim on them. Section~\ref{sec:spearman} reports a fully mechanical rank-correlation statistic computed from the same per-event data, and readers who distrust the labels can read the headline result from that analysis alone; the two agree on the bottom line.

A ``negative'' trendline ({\color{blue}$\downarrow$}) indicates that higher similarity scores correlate with higher abnormal returns---the core requirement for the shadow trading theory. Conversely, a ``positive'' trendline ({\color{red}$\uparrow$}) indicates that higher similarity scores correlate with lower abnormal returns. However, if a trend is negative but the absolute returns for the top-ranked peers are negligible or negative, the case does not ``support'' a theory of predictable profit. Peer counts per event vary below ten for several reasons: firms identified in Stage~1 were subsequently excluded when their EDGAR filings lacked sufficient Item~7 content, or when historical price data was unavailable due to post-announcement delisting.

\begin{table*}[t]
    \small
    \centering
    \caption{Summary of all 30 acquisition events. \\
      Trend: {\color{blue}$\downarrow$} = negative (similarity predicts positive AR);
      {\color{red}$\uparrow$} = positive (similarity predicts negative AR). \\
      $^\dagger$Factual basis of \textit{SEC v.\ Panuwat}; used as a reference case (Section~\ref{sec:groundtruth}). \\
      Per-event rank correlations for these same events are reported in Table~\ref{tab:spearman}.}
    \setlength{\tabcolsep}{2pt}
    \begin{tabular}{@{}llcclcccl@{}}
    \toprule
    \textbf{Industry} & \textbf{Acquisition Announcement} & \textbf{Year} &
    \textbf{Peers} & \textbf{Trend} &
    \multicolumn{3}{c}{\textbf{Top-3 AR Direction}} &
    \textbf{Interpretation} \\
    \cmidrule(lr){6-8}
    \midrule
    \multirow{10}{*}{Biopharmaceuticals} 
    & Pfizer / Medivation$^\dagger$  & 2016 & 5  & {\color{blue}$\downarrow$} & $+$ & $+$ & $+$ & \color{blue}Supports (reference) \\
    & AbbVie / ImmunoGen             & 2023 & 8  & {\color{blue}$\downarrow$} & $+$ & $+$ & $+$ & \color{blue}Strongly supports \\
    & BMS / Karuna                   & 2023 & 8  & {\color{blue}$\downarrow$} & $+$ & $-$ & $\approx0$ & \color{blue}Supports \\
      & BMS / Mirati                   & 2023 & 7  & {\color{blue}$\downarrow$}   & $+$ & $-$ & $+$ & \color{red}Contradicts \\
      & BMS / MyoKardia                & 2020 & 8  & {\color{blue}$\downarrow$}   & $+$ & $\approx0$ & $-$ & \color{red}Strongly contradicts \\
       & BMS / Turning Point            & 2022 & 9  & {\color{blue}$\downarrow$} & $+$ & $+$ & $+$ & \color{blue}Supports \\
      & Gilead / Forty Seven           & 2020 & 7  & {\color{red}$\uparrow$}   & $-$ & $+$ & $\approx0$ & \color{red}Contradicts \\
      & Merck / Acceleron              & 2021 & 7  & {\color{blue}$\downarrow$}   & $\approx0$ & $-$ & $+$ & \color{red}Strongly contradicts \\
      & Merck / Pandion                & 2021 & 7  & {\color{blue}$\downarrow$} & $+$ & $+$ & $+$ & \color{blue}Strongly supports \\
      & Pfizer / Seagen                & 2023 & 7  & {\color{red}$\uparrow$}   & $\approx0$ & $-$ & $-$ & \color{red}Contradicts \\
    \midrule
    \multirow{5}{*}{Consumer Staples}
      & Amazon / Whole Foods         & 2017 & 9  & {\color{blue}$\downarrow$} & $-$ & $-$ & $\approx0$ & \color{blue}Supports \\
      & Danone / White Wave Foods              & 2016 & 6  & {\color{blue}$\downarrow$}   & $+$ & $\approx0$ & $\approx0$ & \color{Blue}Supports \\
      & J.M. Smucker / Hostess              & 2023 & 5  & {\color{blue}$\downarrow$} & $+$ & $+$ & $+$ & \color{blue}Supports \\
      & Kroger / Albertsons            & 2022 & 9  & {\color{blue}$\downarrow$}   & $+$ & $-$ & $+$ & \color{red}Contradicts \\
      & Reckitt / Mead-Johnson         & 2017 & 9  & {\color{red}$\uparrow$}   & $\approx0$ & $-$ & $-$ & \color{red}Contradicts \\
    \midrule
    \multirow{5}{*}{Technology}
      & AMD / Xilinx                   & 2020 & 9  & {\color{red}$\uparrow$}   & $+$ & $-$ & $-$ & \color{red}Contradicts \\
      & Avago / Broadcom          & 2015 & 5  & {\color{red}$\uparrow$}   & $-$ & $-$ & $+$ & \color{red}Contradicts \\
      & Cisco / Splunk                 & 2023 & 10 & {\color{blue}$\downarrow$}      & $-$ & $+$ & $+$ & Mixed \\
      & Microchip / Microsemi          & 2018 & 6  & {\color{red}$\uparrow$}   & $-$ & $-$ & $-$ & Mixed \\
      & SS\&C Technologies / DST Systems          & 2018 & 7  & {\color{red}$\uparrow$}   & $+$ & $\approx0$ & $\approx0$ & \color{red}Contradicts \\
    \midrule
    \multirow{5}{*}{Automotives}
    & Apollo / Tenneco      & 2022 & 9 &            {\color{red}$\uparrow$} & $-$ & $+$ & $-$ & Mixed \\
    & Goodyear / Cooper             & 2021 & 9 &
    {\color{blue}$\downarrow$}     &  $+$ & $+$ & $+$ & \color{blue}Supports \\
    & Tenneco / FDML                 & 2018 & 7 &
    {\color{blue}$\downarrow$}      &  $-$ & $+$ & $-$ & \color{blue}Supports \\
    & Warner / Delphi               & 2020 & 7 &
    {\color{red}$\uparrow$}         & $-$ & $\approx 0$ & $+$ & Mixed \\
    & Warner / Remy                 & 2015 & 8 &
    {\color{blue}$\downarrow$}      & $+$ & $+$ & $\approx 0$ & \color{blue}Strongly   supports \\
    \midrule
    \multirow{5}{*}{Finance}
    & BB\&T / SunTrust             & 2019 & 7 &
    {\color{red}$\uparrow$}     &  $+$ & $+$ & $+$ & \color{red}Contradicts \\
    & KeyCorp / FirstNiagara  & 2015 & 5 &
    {\color{blue}$\downarrow$}      &  $+$ & $-$ & $-$ & \color{blue}Strongly supports \\
    & People / United                 & 2019 & 6 &
    {\color{red}$\uparrow$}      & $-$ & $\approx 0$ & $\approx 0$ & \color{red}Contradicts \\
    & Sterling / Astoria               & 2017 & 5 &
    {\color{blue}$\downarrow$}         & $+$ & $-$ & $-$ & \color{blue}Supports\\
    & WSFS / Beneficial      & 2018 & 6 &            {\color{blue}$\downarrow$} & $+$ & $+$ & $+$ & \color{blue}Supports \\
    \midrule
    \multicolumn{4}{l}{\textbf{Summary}} &
      \multicolumn{5}{l}{\textbf{Supports: 14} \quad \textbf{Contradicts: 12} \quad \textbf{Mixed: 4}} \\
    \bottomrule
    \end{tabular}
    \label{tab:results}
\end{table*}

\subsection{Biopharmaceutical Results}
\label{sec:biopharma}
Our analysis begins with the Biopharmaceutical sector, the very crucible of the SEC's shadow trading doctrine. In the Pfizer/Medivation acquisition, the pipeline identified Incyte (the firm traded by Panuwat) as a highly similar peer that experienced a positive abnormal return (Figure~\ref{fig:pfizmed}). This shows that the similarity metric can surface a firm the SEC regarded as economically linked, subject to the caveats in Section~\ref{sec:groundtruth}.

However, when the same methodology is applied across ten major biopharmaceutical acquisitions, that relationship does not hold. As summarized in Table~\ref{tab:results}, the results are close to a coin flip: exactly half of the transactions (five out of ten) contradict the shadow trading hypothesis, and the mean rank correlation for the sector is $+0.12$ (Table~\ref{tab:spearman-industry}), not distinguishable from zero. In these contradictory cases, the firms with the highest 10-K semantic similarity to the target either experienced negligible price movement or negative abnormal returns on the announcement day.

For example, while transactions like AbbVie's acquisition of ImmunoGen generated the uniform sympathy rally the SEC expects (strong positive returns for the top three peers), deals like Pfizer's acquisition of Seagen and Gilead's acquisition of Forty Seven yielded fundamentally unaligned market reactions. In the Pfizer/Seagen transaction, the top three most semantically similar peers registered zero or negative abnormal returns. This directly refutes the premise that shared fundamental business traits—as disclosed in mandatory SEC filings—reliably dictate correlated market behavior.

These findings carry direct implications for NLP-based enforcement heuristics. As reported in Section~\ref{sec:ar}, only $21.7\%$ of the biopharmaceutical peers our pipeline identifies satisfy the standard \$2B--\$10B mid-cap definition, so the cohort is not structurally homogeneous in the way the \textit{Panuwat} framing suggests. Even so, a state-of-the-art LLM pipeline operating on the SEC's own mandated disclosures predicts peer stock reactions at chance level within this sector---suggesting that ``economic linkage,'' as operationalized through disclosure text by this pipeline, is an unreliable signal rather than a consistent market phenomenon.

\subsection{Cross-Industry Results}
\label{sec:crossindustry}
To determine whether the shadow trading phenomenon extends beyond the Biopharmaceutical sector, we expanded our NLP analysis to twenty additional acquisitions across four diverse industries: \textit{Consumer Staples, Technology, Automotives}, and \textit{Finance}. As detailed in Table~\ref{tab:results}, the expansion of the dataset further erodes the structural validity of the SEC's ``economic linkage'' theory, with results distributed across supporting, contradicting, and mixed outcomes.

The breakdown by sector reveals profound inconsistencies that undermine any notion of a universal, predictable market mechanism. For example, deals like AMD/Xilinx and Avago/Broadcom produced inverted trend lines where highly similar peers saw negligible or negative returns. Similarly, the Consumer Staples and Finance sectors exhibited near-random distributions of support and contradiction. In the BB\&T/SunTrust banking merger, a robust sympathy rally lifted the entire cohort, yet the magnitude of those gains completely decoupled from semantic similarity, yielding a strongly contradictory trend line. As shown in Table~\ref{tab:industry-summary}, Technology is the only sector with zero supporting cases while Automotives yields zero contradictions. One reading is that document-recoverable linkage is industry-contingent, with tight supplier ecosystems in automotive manufacturing producing spillovers that dispersed platform-based technology firms do not. We flag this as a hypothesis rather than a finding: each sector contributes five events, and the corresponding rank correlations (Table~\ref{tab:spearman-industry}) are not distinguishable from zero.

Across the full 30-event dataset, semantic similarity predicts peer stock movement direction in $14/30$ cases ($46.7\%$), counting the four mixed events as non-supporting. A two-sided binomial test against $H_0 = 0.5$ yields $p \approx 0.86$; excluding the mixed events instead gives $14/26$ and $p \approx 0.85$, so nothing turns on that convention. We stress that this is a \textit{failure to reject} rather than a demonstration of no effect: with $n=30$, a binomial test of this form has little power against modest departures from chance, and absence of evidence is not evidence of absence. The rank-correlation analysis in Section~\ref{sec:spearman} provides the sharper statement, bounding how large an association the data can accommodate. Taken together, the results suggest that ``economic linkage'' as our pipeline operationalizes it is not a stable, document-recoverable feature of firm relationships but an idiosyncratic outcome that varies across industries, deal structures, and market conditions.

\begin{table}[t]
    \small
    \centering
    \caption{Results by industry.}
    \setlength{\tabcolsep}{2pt}
    \begin{tabular}{@{}lcccc@{}}
        \toprule
        \textbf{Industry} & \textbf{Events} & \textbf{Supports} & 
        \textbf{Contradicts} & \textbf{Mixed} \\
        \midrule
        Biopharmaceuticals & 10 & 5 & 5 & 0 \\
        Consumer Staples   & 5  & 3 & 2 & 0 \\
        Technology         & 5  & 0 & 3 & 2 \\
        Automotives        & 5  & 3 & 0 & 2 \\
        Finance            & 5  & 3 & 2 & 0 \\
        \midrule
        \textbf{Total}     & \textbf{30} & \textbf{14} & \textbf{12} & \textbf{4} \\
        \bottomrule
    \end{tabular}
    \label{tab:industry-summary}
\end{table}

\subsection{Rank-Correlation Analysis}
\label{sec:spearman}

The labels in Table~\ref{tab:results} involve judgment (Section~\ref{sec:fullresults}). To give the central claim a form that does not, we compute for each event the Spearman rank correlation $\rho$ between the LLM similarity score and the announcement-day abnormal return, over the same per-event data reported in Appendices~\ref{app:cases} and~\ref{app:plots}. Throughout this section we adopt the convention that $\rho > 0$ means more similar peers earned higher abnormal returns---the direction the shadow trading hypothesis predicts. This is the opposite sign from the trendline arrows in Table~\ref{tab:results}, which regress return on similarity \textit{rank}: because rank~1 denotes the most similar peer, rank runs opposite to score, and the same underlying relationship therefore appears with the opposite slope.

Significance is assessed by a permutation test. Holding the similarity scores fixed, we reshuffle the abnormal returns among the peers of the same event and recompute $\rho$; the $p$-value is the fraction of reshuffles yielding a correlation at least as extreme as the one observed. This assumes only that, under the null, any return could equally well have attached to any peer---no distributional assumption is required. For events with $n \leq 8$ peers we enumerate all $n!$ reassignments exactly; for the eight events with nine or ten peers we sample $200{,}000$ at random. Per-event values appear in Appendix~\ref{app:spearman}.

\paragraph{Pooled result.} Events differ in peer count, so before pooling we convert each event's similarity scores and abnormal returns to within-event ranks and standardize them, placing every event on a common scale. The correlation across all $217$ standardized observations is $\rho_{\text{pooled}} = +0.066$ (permutation $p = 0.37$, reshuffling returns within event). Taking instead the $30$ per-event correlations as the unit of analysis, their mean is $\bar{\rho} = +0.046$ with standard deviation $0.348$; the corresponding $95\%$ confidence interval on that mean, formed as $\bar{\rho} \pm t_{29,\,0.975}\,s/\sqrt{30}$, is $[-0.084, +0.175]$. Seventeen of thirty events have $\rho > 0$, twelve have $\rho < 0$, and one is exactly zero (two-sided sign test $p = 0.58$). Only one event---AbbVie/ImmunoGen ($\rho = +0.81$, $p = 0.022$)---is individually significant at $\alpha = 0.05$, against the $1.5$ expected by chance alone across thirty tests.

\paragraph{What this rules out.} The confidence interval is the substantive result. It is centered near zero and narrow enough to exclude any association stronger than $\rho \approx 0.18$. This is a stronger statement than the binomial test supports: rather than merely failing to detect a relationship, the data are inconsistent with a moderate or large one. An enforcement standard resting on the premise that filing similarity identifies the firms that will move on an announcement requires an association far stronger than the upper end of this interval.

\paragraph{Agreement with the qualitative labels.} Of the $25$ events assigned a decisive label and a nonzero $\rho$, the sign of $\rho$ agrees with the label in $19$ ($76\%$). The six disagreements are informative rather than alarming: they arise where the OLS trendline and the rank correlation weigh outliers differently. Pfizer/Medivation is one such case---the trendline is negative, but $\rho = -0.10$, because Puma Biotechnology (rank~9) posted the single largest abnormal return in the event. That the mechanical statistic and the holistic labels disagree on individual events while converging on the same aggregate conclusion is the appropriate reading: the aggregate null is robust to how one scores any particular case.

\begin{table}[t]
    \small
    \centering
    \caption{Mean per-event Spearman correlation by industry. $\rho > 0$ is the direction the shadow trading hypothesis predicts.}
    \setlength{\tabcolsep}{3pt}
    \begin{tabular}{@{}lccc@{}}
        \toprule
        \textbf{Industry} & \textbf{Events} & \textbf{Mean } $\boldsymbol{\rho}$ & $\boldsymbol{\rho>0}$ \\
        \midrule
        Biopharmaceuticals & 10 & $+0.117$ & 6/10 \\
        Consumer Staples   & 5  & $+0.156$ & 4/5 \\
        Technology         & 5  & $-0.239$ & 1/5 \\
        Automotives        & 5  & $+0.085$ & 3/5 \\
        Finance            & 5  & $+0.036$ & 3/5 \\
        \midrule
        \textbf{Total}     & \textbf{30} & $\boldsymbol{+0.046}$ & \textbf{17/30} \\
        \bottomrule
    \end{tabular}
    \label{tab:spearman-industry}
\end{table}

The industry pattern in Table~\ref{tab:spearman-industry} is directionally consistent with the label-based breakdown---Technology is the only sector with a negative mean---but the per-industry samples are five events each, and none of these means is distinguishable from zero. We report them as description, not as evidence of sector-specific effects.

\section{Discussion}
\label{sec:discussion}

\paragraph{LLM similarity as an enforcement heuristic.}
By construction, our pipeline can bridge synonymous terminology that bag-of-words similarity treats as unrelated. In cases like AbbVie/ImmunoGen, textual similarity tracks announcement-day abnormal returns closely. But across the corpus the association is indistinguishable from zero ($\rho_{\text{pooled}} = +0.07$, $p = 0.37$; $14/30$ events supportive), and that is insufficient to serve as the basis for a legal standard: a standard right about as often as it is wrong provides neither the fair notice required by due process~\citep{sessionsdimaya} nor the precompliance-review predicate that \textit{Patel} requires for warrantless administrative access to business records~\citep{patel2015}. If some sector did exhibit a reliable signal, targeted watchlists would be preferable to whole-market collection on that ground alone---but our data do not identify such a sector. The per-industry correlations in Table~\ref{tab:spearman-industry} rest on five events each and none is distinguishable from zero, so we cannot recommend any sector as a candidate for a narrower list. A recent graph-based deep learning framework for shadow trading detection confirms the computational intensity of the problem~\citep{stenfors2025}, but scale alone does not resolve the constitutional defect---it accentuates it.

\paragraph{Failure mode analysis.} Qualitative inspection of contradicting cases reveals two recurring failure modes. The first is \textit{sector-wide contagion}: in BB\&T/SunTrust, all peers rallied regardless of similarity rank because the merger signaled regulatory approval of large-bank consolidation generally, swamping any firm-specific signal. The second is \textit{outlier dominance}: in BMS/MyoKardia, a single firm (CYTK, $+14.3\%$ AR) drove an apparent negative trendline while the remaining peers showed negligible returns, making the case technically ``contradicting'' by trendline but reflecting one idiosyncratic event rather than a systematic pattern. These failure modes suggest that even where 10-K similarity is high, macro-structural events and single-stock noise can overwhelm the announcement-day signal. Both are properties of the outcome measure rather than of the text, which is one reason we treat the one-day abnormal return as a limitation of the test rather than a verdict on the disclosures.

\paragraph{Why mass surveillance persists.}
The fundamental issue is not that similarity never tracks peer reactions---it does in some events---but that it does so \textit{unpredictably}. An insider running our pipeline before trading would face genuine uncertainty about whether a given trade crosses the shadow trading line. The SEC, by contrast, can observe the announcement-day return first and construct a textual similarity argument post hoc---which is essentially the \textit{Panuwat} fact pattern, where ``mid-cap oncology'' was an analyst category invoked retrospectively to explain Incyte's price reaction. In January 2026 the SEC approved an amendment eliminating names, addresses, birth years, and taxpayer identifiers from CAT reporting and directing the deletion of customer data already collected~\citep{caisamend2026}. That change does not disturb the search-first/identify-later inversion: replacing directly identifying data with pseudonymous identifiers addresses the objection to holding personal information in bulk, but leaves untouched the structural defect that the trigger for re-identifying a trader cannot be specified \textit{ex ante} from public disclosures.

\paragraph{Implications for live litigation.}
A state-of-the-art LLM, given the same mandated disclosures an insider would read, identifies shadow trading targets at a rate indistinguishable from chance in our sample. We are careful about what follows from this. Our result concerns one pipeline, one corpus, and one return measure; it does not establish that no method could recover economic linkage, nor that mass surveillance is legally required. What it does show is that the administrability premise---that a person of ordinary intelligence could determine ex ante which securities are off-limits---has not been demonstrated and is not supported by the most capable text-analytic tool we could bring to bear on the disclosures the doctrine itself points to. That is a question courts weighing \textit{Davidson v.\ Atkins}~\citep{davidson2026} and the appellate posture of \textit{Panuwat} will have to confront, and it is the contribution we claim.

\section{Related Work}
\label{sec:relatedwork}

\paragraph{Firm similarity from financial text.}
Prior work applying HP-style text similarity to financial disclosures has focused on market clustering, risk assessment, and portfolio optimization~\citep{hanley2019dynamic, chen-sarkar-2020-semantic}, and on evaluating LLMs applied to 10-K sections for question answering and information extraction tasks~\citep{araci2019finbert, wu2023bloomberggpt}. Closest to our setup, \citet{koval2024comparing} learn to compare pairs of financial reports for forecasting, showing that document-scale report comparison is an established task; their objective is predictive accuracy, whereas ours is whether a similarity ranking can discharge a legal identifiability requirement. None of this work evaluates whether textual firm similarity predicts announcement-day stock reactions, nor does it examine the legal implications of that relationship. We extend this lineage by applying long-context generative LLM reasoning to operationalize the SEC's ``economic linkage'' standard and test it empirically against market outcomes.

\paragraph{Shadow trading in finance and law.}
\citet{mehta2021} introduced ``shadow trading'' in the economics literature, estimating single-event profits of \$139{,}400--\$678{,}000; \citet{augustin2019} documented positive abnormal options volume in $\sim$25\% of U.S.\ takeovers between 1996--2012. \citet{brattle2025} discuss enforcement implications from recent SEC cases. \citet{stenfors2025} propose AMGIN, a graph-based surveillance framework that models the market as a spatio-temporal graph integrating sectoral ties and price co-movements; unlike our approach, AMGIN operates on market structure data rather than disclosure text and is designed as a regulator surveillance tool rather than an ex ante identifiability test. To our knowledge, this paper is the first to apply LLMs to mandatory SEC disclosures to operationalize economic linkage as a legal standard and evaluate whether that textual signal predicts peer stock reactions across industries.

\paragraph{Privacy and surveillance in securities markets.}
\citet{peirce2019} and \citet{barr2024} document privacy concerns with CAT; \citet{davidson2024} challenges its Fourth Amendment constitutionality~\citep{davidson2026}. \citet{carpenter2018} provides the Supreme Court framework for comprehensive digital surveillance, and in \textit{Chatrie v.\ United States}~\citep{chatrie2026} the Court extended it to an architecturally analogous bulk-then-unmask program, holding that querying a bulk repository for an individual's location history is a Fourth Amendment search because a person retains a reasonable expectation of privacy in those records. CAT has the same architecture: identifying a single trader means querying a record assembled from everyone. \citet{demontjoye2013} show that pseudonymization is a brittle defense---four spatio-temporal points identify 95\% of individuals in a 1.5M-person corpus---a result that bears directly on CAT's reliance on pseudonymous customer identifiers~\citep{acquisti2015privacy, solove2006taxonomy}.

\section{Limitations}
\label{sec:limitations}

Several limitations bound the scope of our conclusions, and two of them are severe enough that we state them before the rest.

\paragraph{No baselines.} We did not compare our pipeline against TNIC-style bag-of-words similarity~\citep{hobergphillips2016}, SIC/GICS industry peers, TF-IDF or embedding similarity, or random within-sector peer rankings. This is the most consequential gap in the study. Without such comparisons we cannot distinguish three explanations for our null result: that public filings do not encode the relevant linkage, that this particular LLM pipeline fails to extract it, or that announcement-day abnormal returns are too noisy a target for any similarity measure. Our claim concerns this pipeline, not NLP generally.

\paragraph{The similarity scores are not validated as measurements.} Each event was scored once. We report no run-to-run variance, no sensitivity to prompt or rubric wording, no second model, and no agreement with human expert judgments. LLMs used as evaluators are known to exhibit systematic biases and position or verbosity effects~\citep{wang2024fairevaluators}, and our rubric weights (40/30/30) were chosen to reflect the \textit{Panuwat} testimony rather than tuned or validated. Because Gemini~3.1~Pro is closed-weight, its outputs are also not guaranteed stable across API versions. We therefore treat the scores as structured qualitative judgments rather than deterministic measurements, and a stability study is the first thing we would add.

\paragraph{Statistical power and the meaning of the null.} A non-rejected null is not proof of no effect. Our binomial test on $14/30$ ($p \approx 0.86$) should not be read as demonstrating the absence of a relationship. The rank-correlation confidence interval in Section~\ref{sec:spearman}, $[-0.08, +0.18]$, is the defensible version of the claim: it bounds how large an association the data can accommodate. Industry-level statements rest on five events each and are descriptive only.

\paragraph{Construct validity of the outcome measure.} We equate ``economic linkage'' with a same-day price response, which is narrower than the legal concept of materiality to another issuer's securities. Linked firms might react over longer windows, through options markets, or through negative competitive channels. A perfect linkage measure could still fail to predict a one-day return dominated by liquidity and idiosyncratic noise; our negative result may therefore indict the proxy rather than identifiability. Relatedly, the expected sign is deal-dependent---in Amazon/Whole~Foods we score a case as supportive when similar peers fell \textit{further}, on the reasoning that the acquisition signaled competitive pressure---and we did not fix a sign convention by deal type in advance.

\paragraph{Peer discovery is not a clean filings-only condition.} Stage~1 ran with Google Search grounding enabled to confirm historical listing status, and we did not log the retrieved sources. Filing-derived evidence, search-derived evidence, and the model's parametric knowledge are therefore not fully separable, and for widely reported deals the model may recall the outcome. This bounds how strictly the ``identifiable from public filings'' framing should be read.

\paragraph{Sample and design.} The dataset covers 30 events across five industries spanning 2013--2023 and omits sectors such as energy, real estate, and healthcare services. The event set was assembled by the authors rather than drawn from a pre-registered universe, and the supports/contradicts/mixed labels were assigned after inspecting returns (Section~\ref{sec:fullresults}). Abnormal returns use sector ETFs rather than firm-specific market models---consistent with the SEC economist's approach in \textit{Panuwat}, but introducing benchmark risk for firms whose betas deviate from their sector. Finally, the market-capitalization figures in Section~\ref{sec:ar} exclude delisted firms and firms whose ticker symbols were later reassigned, which biases the surviving sample toward larger capitalizations.

\section{Conclusion}
\label{sec:conclusion}

We present an NLP-grounded empirical evaluation of the SEC's shadow trading enforcement theory. On the \textit{Panuwat} fact pattern our two-stage LLM pipeline recovers Incyte among the closest peers. Across 30 acquisition events, textual similarity in 10-K MD\&A sections shows no association with announcement-day peer reactions: the pooled within-event rank correlation is $+0.07$ ($p = 0.37$) and the mean per-event Spearman correlation is $+0.05$ with a $95\%$ confidence interval of $[-0.08, +0.18]$. Under a case-level reading, $14$ of $30$ events support the hypothesis and $12$ contradict it. The confidence interval is the operative result: it excludes any moderate association, rather than merely failing to find one.

We are deliberate about the scope of this claim. We tested one closed-weight model, one rubric, one filing section, and one abnormal-return measure, without baseline comparisons. Our evidence supports the conclusion that \textit{this} pipeline does not recover economic linkage from Item~7 text in a way that predicts announcement-day returns. It does not establish that no NLP method could, and a null of this kind cannot prove the absence of an effect---only bound its plausible size.

Future work should add the baselines this study lacks (TNIC-style bag-of-words similarity, industry-code and embedding peers), test multiple models and repeated runs to quantify scoring stability, expand the dataset, and combine multiple 10-K sections (Items~1, 1A, and~7) or supplementary filings (Form~8-K, Schedule~13D). A promising inversion of our design would begin from the market rather than the filings: scan a broad universe for the largest abnormal movers around each announcement, then ask whether those firms share characteristics that were publicly knowable beforehand. That reverses the direction of inference and would test identifiability without presupposing that our peer-discovery stage found the right candidates.

The broader stake is unchanged by these caveats. Shadow trading liability presumes that an insider can determine in advance which securities are off-limits, and our results give no support to that premise: the most capable text-analytic tool we could bring to bear on the disclosures the doctrine itself points to recovers no usable signal, while a linkage argument remains easy to construct once the price reaction is known. That asymmetry---cheap to assert after the fact, unavailable before it---is what the evidence assembled here documents, and it is what a defense of the doctrine's administrability now has to answer.

\section*{Impact Statement}

This paper evaluates whether NLP can operationalize the SEC's ``economic linkage'' standard for shadow trading enforcement---a legal theory with direct consequences for the largest government-mandated collection of personal financial data in U.S.\ history. We identify two categories of broader impact.
\paragraph{Beneficial impacts.} Our findings provide courts, policymakers, and legal scholars with an empirical test of the predictability premise underlying shadow trading enforcement. If economic linkage cannot be reliably identified ex ante from public disclosures---as our results suggest---then targeted watchlists cannot replace the mass surveillance architecture of the Consolidated Audit Trail, and constitutional challenges to that architecture are empirically grounded. This has direct relevance to pending litigation (\textit{Davidson v.\ Atkins}; \textit{Panuwat} at the Ninth Circuit) and to ongoing debates about the scope of the SEC's administrative authority. The dataset and two-stage pipeline documented in the Appendix may also benefit researchers studying NLP for financial regulation, legal AI, and text-based firm similarity.
\paragraph{Potential risks.} A sophisticated actor could in principle adapt this approach to look for trading opportunities in sectors where the signal appears stronger. We note four mitigating factors: (1) all inputs---10-K filings---are already publicly available on EDGAR; (2) the overall association is indistinguishable from zero, and the apparent sector differences rest on five events each and are not statistically distinguishable from chance, so there is no validated high-signal sector to exploit; (3) any trading on the basis of publicly available disclosure text would not constitute insider trading under any current theory of liability; and (4) the pipeline is not a trading tool and should not be used as one---the results here are exploratory and provide no reliable basis for trading, portfolio construction, or enforcement targeting. We emphasize this last point because a negative result about administrability could be misread as a positive result about predictability in the cases that did line up; it is not. A secondary risk is that our findings could be read as endorsing shadow trading. We emphasize the contrary: our empirical result that the enforcement theory lacks a reliable ex ante basis raises constitutional questions about the \textit{infrastructure} used to enforce it, not the underlying prohibition on trading on material nonpublic information.

\section*{Acknowledgments}
\label{sec:acknowledgements}

We are grateful to Professor Sebastian Zimmeck (Wesleyan University) and Alex Abdo (Knight First Amendment Institute, Columbia University) for their support and feedback throughout this project. This research was developed as part of the interdisciplinary seminar on Anonymity \& Privacy at Columbia Law School and Columbia Engineering. We would also like to thank Professor John C.\ Coffee, Jr.\ (Columbia Law School) for insightful discussions on insider trading doctrine and the development of the paper's legal arguments.


\bibliography{example_paper}
\bibliographystyle{icml2026}

%%%%%%%%%%%%%%%%%%%%%%%%%%%%%%%%%%%%%%%%%%%%%%%%%%%%%%%%%%%%%%%%%%%%%%%%%%%%%%%
%%%%%%%%%%%%%%%%%%%%%%%%%%%%%%%%%%%%%%%%%%%%%%%%%%%%%%%%%%%%%%%%%%%%%%%%%%%%%%%
% APPENDIX
%%%%%%%%%%%%%%%%%%%%%%%%%%%%%%%%%%%%%%%%%%%%%%%%%%%%%%%%%%%%%%%%%%%%%%%%%%%%%%%
%%%%%%%%%%%%%%%%%%%%%%%%%%%%%%%%%%%%%%%%%%%%%%%%%%%%%%%%%%%%%%%%%%%%%%%%%%%%%%%
\newpage
\appendix
\onecolumn

\section{Prompt Templates}
\label{app:prompts}

We define our multi-stage prompting architecture in Table~\ref{tab:prompts}, which outlines the system instructions and the dimensional rubrics provided to the model.

% Deliberately not a float: as a table environment this was deferred to the
% bottom of the page and left half a column blank above it. \captionof gives
% the same numbering and caption style with no placement logic at all.
\begingroup
    \centering
    \footnotesize
    \renewcommand{\arraystretch}{1.4}
    \setlength{\tabcolsep}{3pt}
    \captionof{table}{LLM Prompt Specifications.}
    \label{tab:prompts}
    \begin{tabularx}{\columnwidth}{@{} l >{\raggedright\arraybackslash}X @{}}
        \toprule
        \textbf{Type} & \textbf{Prompt Content} \\
        \midrule
        \textbf{System} & You are a securities analyst and computational linguist specializing in the analysis of SEC regulatory filings. You assess economic similarity between public companies based solely on their formal disclosures, applying the standard a sophisticated reasonable investor would use when evaluating whether two firms occupy the same market niche. \\
        \midrule
        \textbf{Stage 1} & I have uploaded the 10-K Item 7 (MD\&A) for [TARGET\_COMPANY], covering fiscal year [YEAR]. This was the primary disclosure available when [ACQUIRER] announced its acquisition of [TARGET\_COMPANY] in [MONTH\_YEAR]. \smallskip \par
        Using only this document, identify 10 publicly traded companies that were most similar to [TARGET\_COMPANY] as of [MONTH\_YEAR]. \smallskip \par
        Constraints: \par
        - Use No Hindsight: Base your assessment only on the state of the market as of [MONTH\_YEAR] \par
        - Reasonable Insider Standard: Prioritize firms that an employee at [TARGET\_COMPANY] would realistically view as a peer or benchmark company \par
        - Do not include [ACQUIRER] in your results \par
        - SEC Filing Requirement: All 10 companies must have an active SEC filing obligation (10-K or 20-F). \par
        - Price Data Availability: All 10 companies must have been actively traded on a major US exchange (NYSE, NASDAQ). \smallskip \par
        For each, provide: A) Name/Ticker; B) Dimensional Justification (B1: Technical, B2: Strategic, B3: Risk); C) Validation Document. \\
        \midrule
        \textbf{Stage 2} & Using only the content of these uploaded documents, assign each candidate a Total Similarity Score (0.00 to 1.00) based on: \smallskip \par
        A) Technical/Product Overlap (40\%): Core technology or product engines. \par
        B) Financial/Strategic Stage (30\%): Revenue growth and commercialization milestones. \par
        C) Risk \& Market Exposure (30\%): Geographic focus and competitive headwinds. \smallskip \par
        Additionally, apply dual-track analysis: \par
        D) Noun-Based Technical Comparison (HP Logic): Identify specific mechanism-of-action nouns. \par
        E) Semantic Bridging: Resolve instances where companies describe identical strategies using different terminology. This is critical to assessing the ``Administrability Gap.'' \\
        \bottomrule
    \end{tabularx}
    \par
\endgroup
\bigskip

\begin{table}[H]
\centering
\scriptsize
\renewcommand{\arraystretch}{1.02}
\setlength{\tabcolsep}{2pt}
\caption{Stage 1 Output: Pfizer/Medivation}
\resizebox{0.65\textwidth}{!}{%
\begin{tabular}{@{} l l p{2.6cm} p{2.6cm} p{2.6cm} @{}}
\toprule
\textbf{\#} & \textbf{Company (Ticker)} & \textbf{B1: Technical} & \textbf{B2: Strategic} & \textbf{B3: Risk} \\
\midrule

1 & Incyte (INCY) & Targeted oncology and hematologic malignancies overlap (e.g., lymphoma, myeloma). & Single-product revenue model with major global collaboration (e.g., Jakafi/Novartis). & Exposure to gross-to-net pricing dynamics and reliance on ex-US partners. \\

2 & Exelixis (EXEL) & Small-molecule oncology therapies overlapping prostate and breast cancer. & Transition to fully integrated commercial model with internal oncology sales force. & Inventory volatility and R\&D expense tied to late-stage trials. \\

3 & Tesaro (TSRO) & PARP inhibitor targeting ovarian and breast cancer; direct overlap with MDV3800. & Scaling SG\&A for commercial launch; growth via in-licensing. & Manufacturing scale-up risks for PARP inhibitors. \\

4 & Clovis Oncology (CLVS) & PARP inhibitor (rucaparib) for ovarian and prostate cancers. & Focus on genetically defined populations using companion diagnostics. & FDA setbacks and clinical/regulatory volatility. \\

5 & Seattle Genetics (SGEN) & Antibody-based oncology therapies for lymphoma and breast cancer. & Single-product revenue base reinvested into label expansion; co-development partnerships. & Pricing pressure and complex collaboration revenue accounting. \\

6 & Ariad Pharmaceuticals (ARIA) & Targeted therapies for hematologic and solid tumors. & Transition to independent commercialization with specialized sales teams. & High exposure to drug pricing pressure and advocacy scrutiny. \\

7 & Puma Biotechnology (PBYI) & Targeted breast cancer therapy (neratinib) overlapping TNBC strategy. & Growth via in-licensed assets from large pharma. & Large contingent liabilities tied to milestone payments. \\

8 & Merrimack (MACK) & Targeted oncology therapies for solid tumors (breast, pancreatic). & Early-stage commercialization with ex-US licensing partnerships. & Manufacturing scale-up and clinical trial cost expansion risks. \\

9 & Immunomedics (IMMU) & Antibody therapies targeting TNBC; direct indication overlap. & Advancing assets into Phase 3 with companion diagnostics. & Regulatory volatility and clinical trial design risk. \\

10 & Celldex Therapeutics (CLDX) & Antibody-based immuno-oncology therapies for breast cancer and lymphoma. & Mixed internal discovery and in-licensing strategy. & High IPR\&D impairment risk and clinical failure exposure. \\

\bottomrule
\end{tabular}%
}
\label{tab:stage1_pfizer_medivation}
\end{table}

\vspace{0.5em}

\begin{table}[H]
\centering
\scriptsize
\renewcommand{\arraystretch}{1.02}
\setlength{\tabcolsep}{2pt}
\caption{Stage 2 Output: Pfizer/Medivation}
\resizebox{0.65\textwidth}{!}{%
\begin{tabular}{@{} l l c c c c p{5.2cm} @{}}
\toprule
\textbf{Rank} & \textbf{Company} & \textbf{Tech} & \textbf{Stage} & \textbf{Risk} & \textbf{Total} & \textbf{Semantic Justification} \\
\midrule

1 & TSRO & 0.38 & 0.25 & 0.25 & \textbf{0.88} & Identical PARP inhibitor terminology and breast/ovarian cancer focus; overlapping PD-1 immuno-oncology pipeline narrative. \\

2 & EXEL & 0.30 & 0.28 & 0.28 & \textbf{0.86} & Matching U.S. profit-sharing and ex-U.S. royalty structures; identical mCRPC clinical focus. \\

3 & INCY & 0.25 & 0.29 & 0.28 & \textbf{0.82} & Single-product revenue with ex-U.S. partner; identical Medicare Part D gross-to-net pricing dynamics. \\

4 & CLVS & 0.35 & 0.20 & 0.25 & \textbf{0.80} & Direct PARP inhibitor overlap; mirrored IPR\&D impairment and FDA-related clinical setbacks. \\

5 & SGEN & 0.25 & 0.28 & 0.25 & \textbf{0.78} & Identical U.S./ex-U.S. commercialization split and healthcare pricing pressure exposure. \\

6 & ARIA & 0.20 & 0.25 & 0.25 & \textbf{0.70} & Similar reliance on third-party data and exposure to global pricing and reimbursement pressures. \\

7 & MACK & 0.20 & 0.22 & 0.20 & \textbf{0.62} & Parallel ex-U.S. licensing strategy used to fund internal oncology R\&D. \\

8 & CLDX & 0.30 & 0.10 & 0.15 & \textbf{0.55} & Immuno-oncology and TNBC overlap; weaker commercial-stage alignment. \\

9 & PBYI & 0.25 & 0.15 & 0.15 & \textbf{0.55} & Breast cancer focus with similar build-out of internal oncology sales infrastructure. \\

10 & IMMU & 0.28 & 0.10 & 0.15 & \textbf{0.53} & Strong TNBC overlap but penalized for pre-commercial financial stage. \\

\bottomrule
\end{tabular}%
}
\label{tab:stage2_pfizer_medivation}
\end{table}

\begin{table}[H]
\centering
\scriptsize
\renewcommand{\arraystretch}{1.02}
\setlength{\tabcolsep}{2pt}
\caption{Stage 1 Output: Pfizer / Seagen (2023)}
\resizebox{0.65\textwidth}{!}{%
\begin{tabular}{@{} l l p{2.6cm} p{2.6cm} p{2.6cm} @{}}
\toprule
\textbf{\#} & \textbf{Company (Ticker)} & \textbf{B1: Technical} & \textbf{B2: Strategic} & \textbf{B3: Risk} \\
\midrule

1 & Genmab (GMAB) & Direct overlap in ADCs and monoclonal antibodies; co-development of TIVDAK targeting tissue factor. & 50/50 co-commercialization and profit-sharing structure identical to Seagen. & Identical product-specific risks (TIVDAK adoption, eye toxicity management, global rollout). \\

2 & Exelixis (EXEL) & Overlap in targeted oncology; ADC pipeline targeting tissue factor (XB002). & Large-scale commercial oncology firm with \$1B+ revenue and global partnerships. & Exposure to U.S. pricing pressure (Medicaid rebates, 340B) and oncology competition. \\

3 & ADC Therapeutics (ADCT) & Pure-play ADC company targeting hematologic malignancies (e.g., CD30-expressing lymphomas). & Recently transitioned to commercial-stage biotech with FDA accelerated approvals. & Manufacturing complexity and geographic commercialization via partners. \\

4 & ImmunoGen (IMGN) & Core ADC platform using monoclonal antibodies to deliver cytotoxic payloads. & Hybrid revenue model (product + licensing/royalties) similar to Seagen. & ADC-specific manufacturing, safety, and clinical execution risks. \\

5 & MacroGenics (MGNX) & Monoclonal antibodies, bispecifics, and ADCs targeting HER2-positive cancers. & Early-commercial oncology firm with licensing and internal pipeline development. & Competitive pressure in HER2 oncology and commercialization cost burden. \\

6 & Blueprint Medicines (BPMC) & Targeted oncology therapies (kinase inhibitors); overlap via precision oncology. & Commercial-stage biotech with expanding global infrastructure. & Medicare/340B pricing exposure and specialty distribution dependence. \\

7 & Zymeworks (ZYME) & HER2-targeted ADCs and bispecific antibodies overlapping Seagen’s pipeline. & Pre-commercial with heavy reliance on licensing deals and milestone funding. & Competitive dynamics in HER2 space and pipeline execution risk. \\

8 & Incyte (INCY) & Broad oncology pipeline (JAK inhibitors, antibodies); limited ADC overlap. & Large commercial biotech with significant royalty revenue streams. & Pricing reform exposure (IRA), global partner reliance, and reimbursement risk. \\

9 & Mersana Therapeutics (MRSN) & ADC platform using proprietary payload technologies (e.g., microtubule inhibitors). & Pre-commercial biotech relying on partnerships to fund R\&D. & High clinical risk and early-stage financial uncertainty. \\

10 & Sutro Biopharma (STRO) & Next-gen ADCs with linker-warhead architecture; strong platform similarity. & Early-stage firm focused on partnerships and pipeline development. & Manufacturing scale-up and pre-commercial execution risk. \\

\bottomrule
\end{tabular}%
}
\label{tab:stage1_pfizer_seagen}
\end{table}

\begin{table}[H]
\centering
\scriptsize
\renewcommand{\arraystretch}{1.02}
\setlength{\tabcolsep}{2pt}
\caption{Stage 2 Output: Pfizer / Seagen (2023)}
\resizebox{0.65\textwidth}{!}{%
\begin{tabular}{@{} l l c c c c p{5.2cm} @{}}
\toprule
\textbf{Rank} & \textbf{Company} & \textbf{Tech} & \textbf{Stage} & \textbf{Risk} & \textbf{Total} & \textbf{Semantic Justification} \\
\midrule

1 & GMAB & 0.38 & 0.26 & 0.28 & \textbf{0.92} & Exact match via TIVDAK co-development; identical product, revenue structure, and geographic commercialization risk profile. \\

2 & EXEL & 0.30 & 0.28 & 0.28 & \textbf{0.86} & Shared targeted oncology focus; identical tissue-factor ADC targeting and nearly identical gross-to-net pricing exposure. \\

3 & ADCT & 0.38 & 0.18 & 0.26 & \textbf{0.82} & Pure-play ADC model targeting hematologic malignancies; strong payload-delivery semantic alignment with Seagen. \\

4 & IMGN & 0.38 & 0.15 & 0.25 & \textbf{0.78} & Identical “antibody-drug conjugate” architecture; payload terminology bridges directly to Seagen’s cell-killing mechanism. \\

5 & MGNX & 0.34 & 0.15 & 0.22 & \textbf{0.71} & HER2 oncology overlap with similar commercialization dynamics and competitive pressures. \\

6 & BPMC & 0.25 & 0.20 & 0.25 & \textbf{0.70} & Shared oral oncology commercialization risks (Medicare Part D, 340B) despite different molecular modality. \\

7 & ZYME & 0.35 & 0.12 & 0.20 & \textbf{0.67} & Direct HER2-targeted ADC overlap but weaker commercial-stage alignment. \\

8 & INCY & 0.15 & 0.25 & 0.26 & \textbf{0.66} & Strong financial structure similarity (royalties, partnerships) despite technical divergence. \\

9 & MRSN & 0.36 & 0.08 & 0.15 & \textbf{0.59} & ADC payload equivalence but large gap due to pre-commercial financial profile. \\

10 & STRO & 0.35 & 0.08 & 0.15 & \textbf{0.58} & Platform-level ADC similarity (“linker-warhead”) but minimal commercialization overlap. \\

\bottomrule
\end{tabular}%
}
\label{tab:stage2_pfizer_seagen}
\end{table}

\section{Per-Event Rank Correlations}
\label{app:spearman}

Table~\ref{tab:spearman} reports the Spearman rank correlation between the LLM similarity
score and the announcement-day abnormal return for each of the 30 events, computed over the
per-event data in Appendices~\ref{app:cases} and~\ref{app:plots}. These statistics are a deterministic function of
the similarity scores and abnormal returns already reported; no additional market data was
collected to produce them.

\begin{table}[h!]
\centering\small
\caption{Per-event Spearman rank correlation between LLM similarity score and announcement-day abnormal return. $\rho>0$ indicates that more similar peers experienced higher abnormal returns (the direction the shadow trading hypothesis predicts). Permutation $p$-values are exact for $n\leq8$ and Monte Carlo ($200{,}000$ draws) otherwise. The final column reports whether the sign of $\rho$ agrees with the holistic label assigned in Table~\ref{tab:results}.}
\begin{tabular}{@{}llcrrl c@{}}
\toprule
\textbf{Industry} & \textbf{Event} & \textbf{Peers} & $\boldsymbol{\rho}$ & \textbf{Perm.\ $p$} & \textbf{Label} & \textbf{Sign agrees} \\
\midrule
Biopharmaceuticals & AbbVie / ImmunoGen & 8 & $+0.810$ & $0.022$ & Supports & \checkmark \\
 & BMS / Karuna & 8 & $-0.238$ & $0.582$ & Supports & $\times$ \\
 & BMS / Mirati & 7 & $+0.179$ & $0.713$ & Contradicts & $\times$ \\
 & BMS / MyoKardia & 8 & $+0.452$ & $0.267$ & Contradicts & $\times$ \\
 & BMS / Turning Point & 9 & $+0.233$ & $0.551$ & Supports & \checkmark \\
 & Gilead / Forty Seven & 7 & $-0.143$ & $0.783$ & Contradicts & \checkmark \\
 & Merck / Acceleron & 7 & $+0.198$ & $0.667$ & Contradicts & $\times$ \\
 & Merck / Pandion & 7 & $+0.214$ & $0.662$ & Supports & \checkmark \\
 & Pfizer / Medivation & 5 & $-0.103$ & $0.900$ & Supports & $\times$ \\
 & Pfizer / Seagen & 7 & $-0.429$ & $0.354$ & Contradicts & \checkmark \\
Consumer Staples & Amazon / Whole Foods & 9 & $+0.417$ & $0.269$ & Supports & \checkmark \\
 & Danone / WhiteWave & 6 & $+0.029$ & $1.000$ & Supports & \checkmark \\
 & J.M. Smucker / Hostess & 5 & $+0.100$ & $0.950$ & Supports & \checkmark \\
 & Kroger / Albertsons & 9 & $+0.233$ & $0.553$ & Contradicts & $\times$ \\
 & Reckitt / Mead Johnson & 9 & $+0.000$ & $1.000$ & Contradicts & --- \\
Technology & AMD / Xilinx & 9 & $-0.117$ & $0.776$ & Contradicts & \checkmark \\
 & Avago / Broadcom & 5 & $-0.800$ & $0.133$ & Contradicts & \checkmark \\
 & Cisco / Splunk & 10 & $+0.188$ & $0.605$ & Mixed & --- \\
 & Microchip / Microsemi & 6 & $-0.143$ & $0.803$ & Mixed & --- \\
 & SS\&C / DST Systems & 7 & $-0.321$ & $0.498$ & Contradicts & \checkmark \\
Automotives & Apollo / Tenneco & 9 & $-0.075$ & $0.852$ & Mixed & --- \\
 & Goodyear / Cooper & 9 & $+0.117$ & $0.775$ & Supports & \checkmark \\
 & Tenneco / Federal-Mogul & 7 & $+0.286$ & $0.556$ & Supports & \checkmark \\
 & Warner / Delphi & 7 & $-0.500$ & $0.267$ & Mixed & --- \\
 & Warner / Remy & 8 & $+0.599$ & $0.125$ & Supports & \checkmark \\
Finance & BB\&T / SunTrust & 7 & $-0.250$ & $0.595$ & Contradicts & \checkmark \\
 & KeyCorp / First Niagara & 5 & $+0.500$ & $0.450$ & Supports & \checkmark \\
 & People's / United & 6 & $-0.371$ & $0.497$ & Contradicts & \checkmark \\
 & Sterling / Astoria & 5 & $+0.100$ & $0.950$ & Supports & \checkmark \\
 & WSFS / Beneficial & 6 & $+0.200$ & $0.714$ & Supports & \checkmark \\
\midrule
\multicolumn{3}{@{}l}{\textbf{Mean}} & $+0.046$ & & & 19/25 \\
\bottomrule
\end{tabular}
\label{tab:spearman}
\end{table}

\FloatBarrier

\section{Case Studies}
\label{app:cases}

\subsection{Reference Case: Pfizer / Medivation (2016)}
\label{app:case-pfizmed}
The Pfizer/Medivation acquisition serves as the factual basis for SEC v. Panuwat. Applying our pipeline ex ante, the model ranks Tesaro, Exelixis, and Incyte as the closest peers, reflecting similarities in oncology pipelines, commercialization approach, and pricing exposure.

All three top-ranked peers exhibit positive abnormal returns (+1.60\%, +1.33\%, +5.04\%). Incyte, the firm actually traded, is ranked third and shows the strongest return \textit{among those three}; the largest abnormal return in the event belongs to Puma Biotechnology (PBYI, $+6.44\%$), ranked ninth. Consequently the rank correlation for this event is slightly negative ($\rho = -0.10$). This case shows that textual similarity can surface a firm the SEC regarded as economically linked, but it does not validate the pipeline: see Section~\ref{sec:groundtruth} for the caveats, including the possibility that the model recalls this widely reported case from pretraining.

\begin{figure}[h!]
\centering
\includegraphics[width=0.75\textwidth]{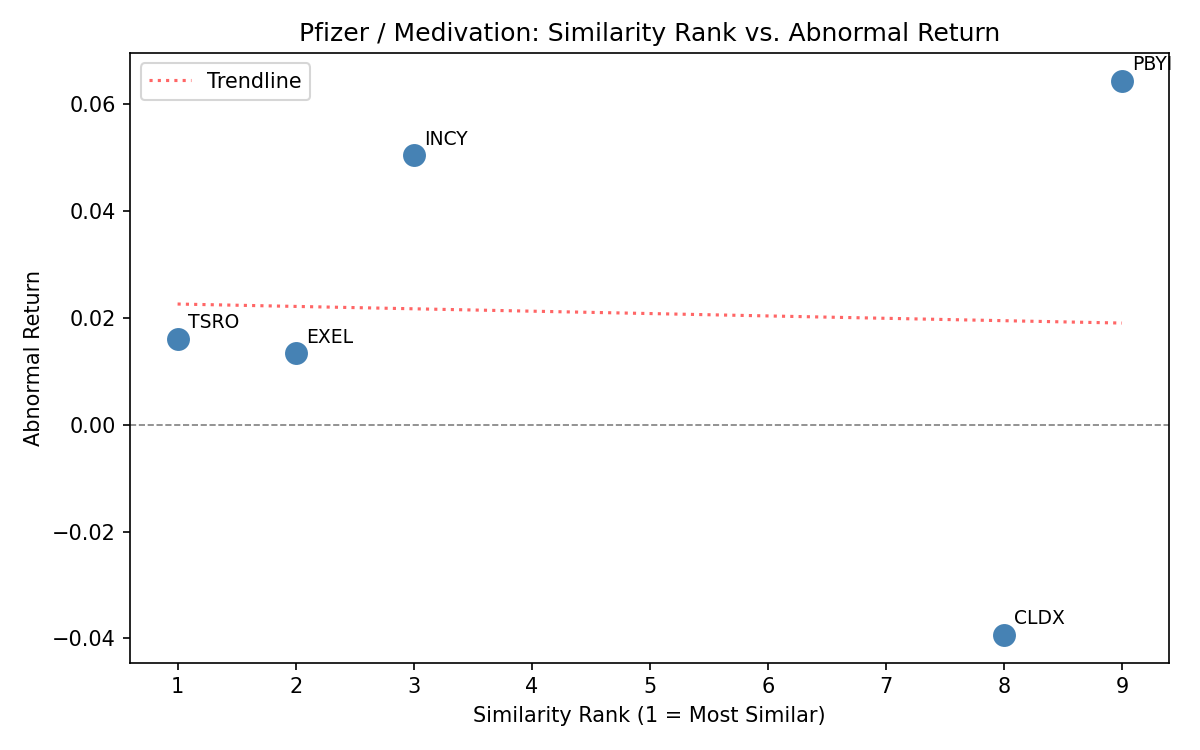}
\caption{Pfizer / Medivation Similarity vs. Abnormal Returns}
\label{fig:pfizer-medivation}
\end{figure}

Tables~\ref{tab:stage1_pfizer_medivation} and~\ref{tab:stage2_pfizer_medivation} show the two-stage LLM outputs. Tables~\ref{tab:medivation-ar} and~\ref{tab:medivation-final} summarize the abnormal return calculations and the combined similarity-ranking analysis, respectively, while Figure~\ref{fig:pfizer-medivation} visualizes the relationship between similarity rank and abnormal returns.

\begin{table}[h!]
\centering
\caption{Abnormal Returns Medivation}
\resizebox{0.65\textwidth}{!}{%
\begin{tabular}{lrrl}
\hline
Ticker & Raw Return & Abnormal Return & Direction \\
\hline
CLDX & -0.012658227848101266 & -0.03933565933364627  & Decreased \\
EXEL &  0.039999961853027344 &  0.013322530367482339 & Increased \\
INCY &  0.07712524897527939  &  0.050447817489734384 & Increased \\
PBYI &  0.09103616209622678  &  0.06435873061068177  & Increased \\
TSRO &  0.04272332537878824  &  0.01604589389324324  & Increased \\
\hline
\end{tabular}
}
\label{tab:medivation-ar}
\end{table}

\begin{table}[h!]
\centering
\caption{Final Analysis Medivation}
\resizebox{0.95\textwidth}{!}{
\begin{tabular}{r l r r r l r}
\hline
Rank & Ticker & Sim. Score & Raw Return & Abnormal Return & Direction & Shadow Signal \\
\hline
1 & TSRO & 0.88 &  0.0427233253787882 &  0.0160458938932432 & Increased &  0.014120386626054016 \\
2 & EXEL & 0.86 &  0.0399999618530273 &  0.0133225303674823 & Increased &  0.011457376116034779 \\
3 & INCY & 0.82 &  0.0771252489752793 &  0.0504478174897343 & Increased &  0.041367210341582125 \\
8 & CLDX & 0.55 & -0.0126582278481012 & -0.0393356593336462 & Decreased & -0.021634612633505412 \\
9 & PBYI & 0.55 &  0.0910361620962267 &  0.0643587306106817 & Increased &  0.03539730183587494 \\
\hline
\end{tabular}
}
\label{tab:medivation-final}
\end{table}

\FloatBarrier

\subsection{Contradicting Case: Pfizer / Seagen (2023)}
\label{app:case-pfizseagen}
The Pfizer/Seagen acquisition provides a direct counterexample in the same industry. Using the identical pipeline, the model again identifies highly similar oncology peers with comparable technical and financial profiles.

However, the market response diverges: the most similar firms show negligible or negative abnormal returns ($\approx$ 0\%, -, -) , and the overall relationship between similarity and returns reverses. Despite strong textual similarity, no clear trading signal emerges, contradicting the predictability assumed by the shadow trading theory.

\begin{figure}[h!]
\centering
\includegraphics[width=0.75\textwidth]{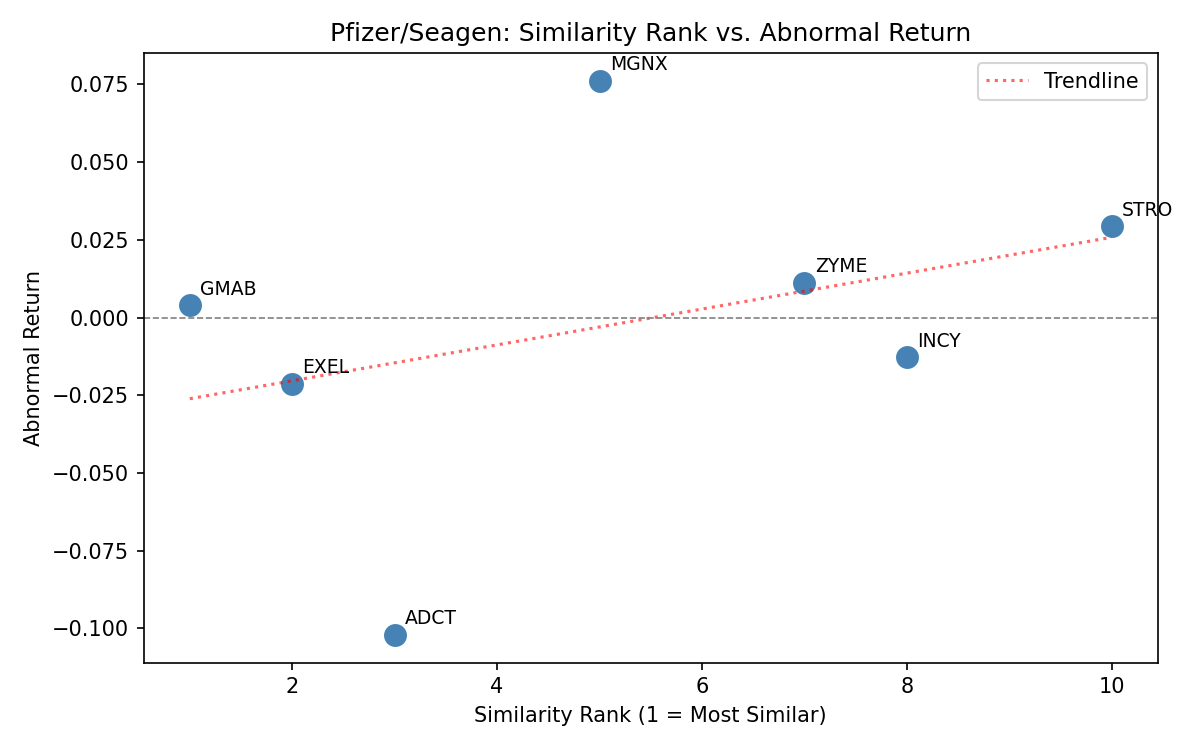}
\caption{Pfizer / Seagen Similarity vs. Abnormal Returns}
\label{fig:pfizer-seagen}
\end{figure}

Tables~\ref{tab:stage1_pfizer_seagen} and~\ref{tab:stage2_pfizer_seagen} show the two-stage LLM outputs. Tables~\ref{tab:seagen-ar} and~\ref{tab:seagen-final} summarize the abnormal return calculations and the combined similarity-ranking analysis, respectively, while Figure~\ref{fig:pfizer-seagen} visualizes the relationship between similarity rank and abnormal returns.

\begin{table}[h!]
\centering
\caption{Abnormal Returns Seagen}
\resizebox{0.65\textwidth}{!}{%
\begin{tabular}{lrrl}
\hline
Ticker & Raw Return & Abnormal Return & Direction \\
\hline
ADCT & -0.07299271787868707 & -0.10207023020129924 & Decreased \\
EXEL &  0.0077937145316635875 & -0.02128379779094858 & Decreased \\
GMAB &  0.03297011685681662  &  0.003892604534204449 & Increased \\
INCY &  0.01624771745919238  & -0.012829794863419789 & Decreased \\
MGNX &  0.10526314374246708  &  0.0761856314198549   & Increased \\
STRO &  0.05847953303345698  &  0.02940202071084481  & Increased \\
ZYME &  0.040353112054028716 &  0.011275599731416546 & Increased \\
\hline
\end{tabular}
}
\label{tab:seagen-ar}
\end{table}

\begin{table}[h!]
\centering
\caption{Final Analysis Seagen}
\resizebox{0.95\textwidth}{!}{
\begin{tabular}{r l r r r l r}
\hline
Rank & Ticker & Sim. Score & Raw Return & Abnormal Return & Direction & Shadow Signal \\
\hline
1  & GMAB & 0.92 &  0.0329701168568166 &  0.0038926045342044 & Increased &  0.003581196171468048 \\
2  & EXEL & 0.86 &  0.0077937145316635 & -0.0212837977909485 & Decreased & -0.01830406610021571 \\
3  & ADCT & 0.82 & -0.0729927178786870 & -0.1020702302012992 & Decreased & -0.08369758876506533 \\
5  & MGNX & 0.71 &  0.1052631437424670 &  0.0761856314198549 & Increased &  0.05409179830809698 \\
7  & ZYME & 0.67 &  0.0403531120540287 &  0.0112755997314165 & Increased &  0.007554651820049055 \\
8  & INCY & 0.66 &  0.0162477174591923 & -0.0128297948634197 & Decreased & -0.008467664609857003 \\
10 & STRO & 0.58 &  0.0584795330334569 &  0.0294020207108448 & Increased &  0.017053172012289983 \\
\hline
\end{tabular}
}
\label{tab:seagen-final}
\end{table}

\FloatBarrier

\section{Remaining Case Studies}
\label{app:plots}
These subsections present the empirical case studies and results of the paper’s analysis. These sections apply the NLP pipeline to real-world acquisition events and examine whether semantic similarity between firms (based on SEC filings) actually predicts stock price reactions—i.e., whether “shadow trading” is empirically supported.

\subsection{Biopharmaceuticals}

\subsubsection{AbbVie / ImmunoGen}

AbbVie's acquisition of ImmunoGen is one of the clearest supporting cases in the dataset. The pipeline identifies ADC-focused peers (ADCT, STRO, MGNX) that show uniformly positive abnormal returns on announcement day, with the top-ranked firm (ADCT, $+5.5\%$ AR) exhibiting the strongest reaction. The negative trendline and positive top-3 AR direction confirm that textual similarity in this case successfully predicts the direction and magnitude of peer stock reactions, consistent with the shadow trading hypothesis.

\begin{figure}[h!]
\centering
\includegraphics[width=0.75\textwidth]{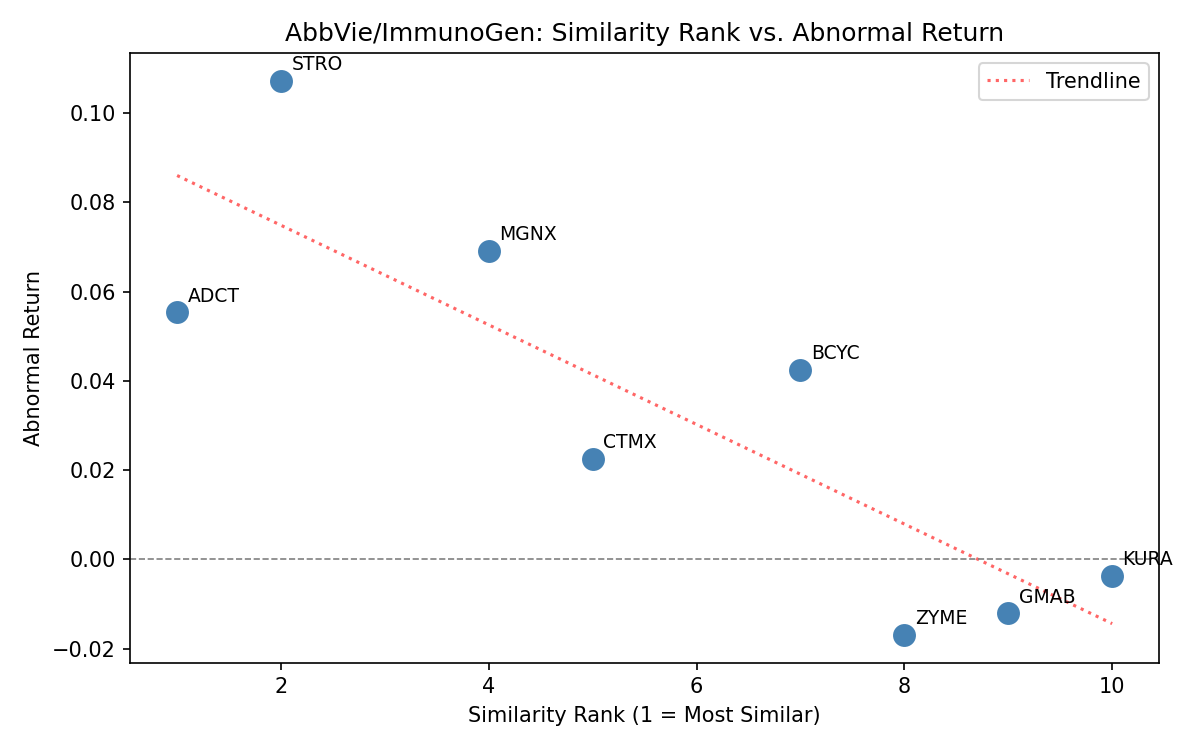}
\caption{AbbVie / ImmunoGen Similarity vs. Abnormal Returns}
\label{fig:abbvie-immunogen}
\end{figure}

\begin{table}[h!]
\centering
\caption{Abnormal Returns ImmunoGen}
\resizebox{0.65\textwidth}{!}{%
\begin{tabular}{lrrl}
\hline
Ticker & Raw Return & Abnormal Return & Direction \\
\hline
ADCT & 0.07792208294861243 &  0.05530851580538361  & Increased \\
BCYC & 0.06508869311918347 &  0.04247512597595465  & Increased \\
CTMX & 0.045112737476325084 & 0.022499170333096268  & Increased \\
GMAB & 0.010549869452110954 & -0.01206369769111786  & Decreased \\
KURA & 0.018967366650902547 & -0.0036462004923262688 & Decreased \\
MGNX & 0.09175532699076502 &  0.0691417598475362   & Increased \\
SNDX & 0.02116572005222625 & -0.0014478470910025652 & Decreased \\
STRO & 0.12987012772547524 &  0.10725656058224642  & Increased \\
ZYME & 0.005668956061779067 & -0.016944611081449748 & Decreased \\
\hline
\end{tabular}
}
\end{table}

\begin{table}[h!]
\centering
\caption{Final Analysis ImmunoGen}
\resizebox{0.95\textwidth}{!}{
\begin{tabular}{r l r r r l r}
\hline
Rank & Ticker & Sim. Score & Raw Return & Abnormal Return & Direction & Shadow Signal \\
\hline
1  & ADCT & 0.93 & 0.0779220829486124 &  0.0553085158053836 & Increased &  0.05143691969900675 \\
2  & STRO & 0.87 & 0.1298701277254752 &  0.1072565605822464 & Increased &  0.09331320770655437 \\
4  & MGNX & 0.74 & 0.0917553269907650 &  0.0691417598475362 & Increased &  0.051164902287176794 \\
5  & CTMX & 0.68 & 0.0451127374763250 &  0.0224991703330962 & Increased &  0.015299435826505416 \\
7  & BCYC & 0.61 & 0.0650886931191834 &  0.0424751259759546 & Increased &  0.025909826845332305 \\
8  & ZYME & 0.52 & 0.0056689560617790 & -0.0169446110814497 & Decreased & -0.008811197762353845 \\
9  & GMAB & 0.40 & 0.0105498694521109 & -0.0120636976911178 & Decreased & -0.00482547907644712 \\
10 & KURA & 0.35 & 0.0189673666509025 & -0.0036462004923262 & Decreased & -0.00127617017231417 \\
\hline
\end{tabular}
}
\end{table}

\FloatBarrier

\subsubsection{BMS / Karuna Therapeutics}

The BMS/Karuna Therapeutics case is classified as supporting, driven primarily by the top-ranked firm RVPH ($+17.0\%$ AR), an extreme positive outlier. The remaining peers show mixed or near-zero reactions. This case illustrates the outlier dominance failure mode: the negative trendline is real but largely attributable to one firm rather than a systematic pattern across the cohort.

\begin{figure}[h!]
\centering
\includegraphics[width=0.75\textwidth]{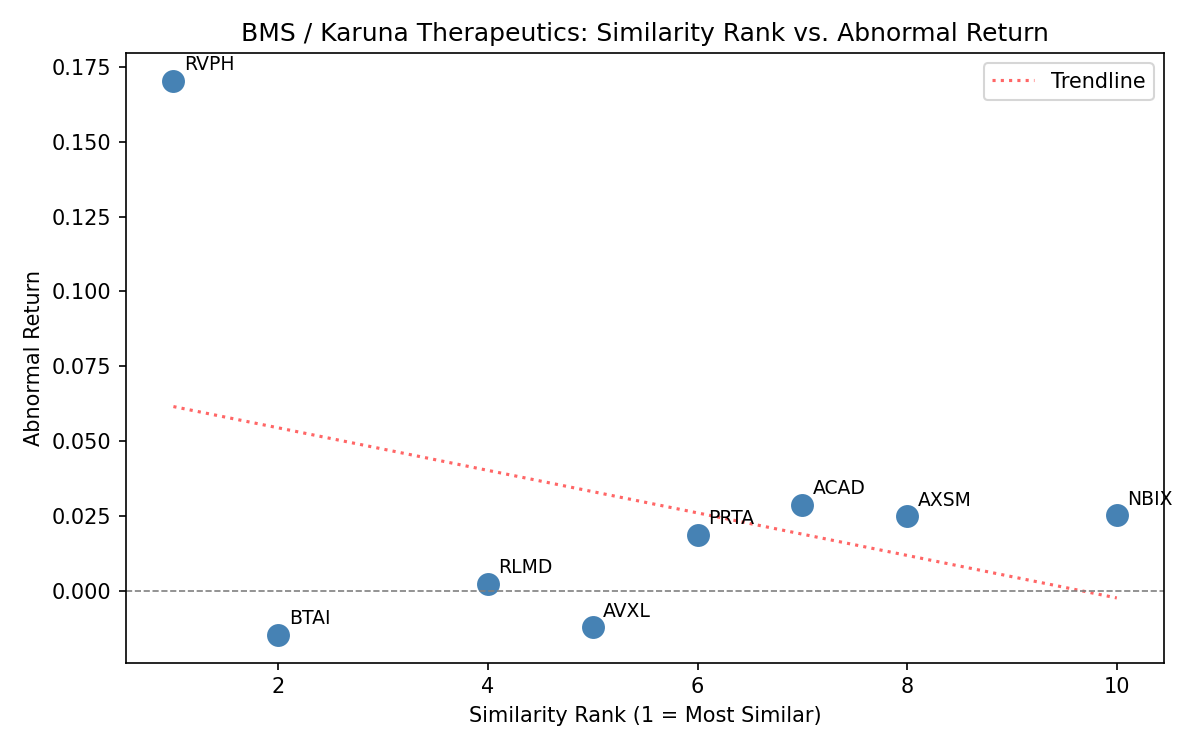}
\caption{BMS / Karuna Therapeutics Similarity vs. Abnormal Returns}
\label{fig:bms-karuna-therapeutics}
\end{figure}

\begin{table}[h!]
\centering
\caption{Abnormal Returns Karuna Therapeutics}
\resizebox{0.65\textwidth}{!}{%
\begin{tabular}{lrrl}
\hline
Ticker & Raw Return & Abnormal Return & Direction \\
\hline
ACAD & 0.0636856074394867  &  0.02884434523941582  & Increased \\
AVXL & 0.02285712105887277 & -0.011984141141198112 & Decreased \\
AXSM & 0.05980958724745921 &  0.02496832504738833  & Increased \\
BTAI & 0.02020200074126503 & -0.014639261458805852 & Decreased \\
NBIX & 0.06016831516828872 &  0.025327052968217842 & Increased \\
PRTA & 0.053662462447954765 & 0.018821200247883883 & Increased \\
RLMD & 0.037288099429759344 & 0.002446837229688463 & Increased \\
RVPH & 0.20519484482802336 &  0.17035358262795247  & Increased \\
\hline
\end{tabular}
}
\end{table}

\begin{table}[h!]
\centering
\caption{Final Analysis Karuna Therapeutics}
\resizebox{0.95\textwidth}{!}{
\begin{tabular}{r l r r r l r}
\hline
Rank & Ticker & Sim. Score & Raw Return & Abnormal Return & Direction & Shadow Signal \\
\hline
1  & RVPH & 0.94 & 0.2051948448280233 &  0.1703535826279524 & Increased &  0.16013236767027522 \\
2  & BTAI & 0.87 & 0.0202020007412650 & -0.0146392614588058 & Decreased & -0.012736157469161045 \\
4  & RLMD & 0.84 & 0.0372880994297593 &  0.0024468372296884 & Increased &  0.002055343272938256 \\
5  & AVXL & 0.81 & 0.0228571210588727 & -0.0119841411411981 & Decreased & -0.009707154324370462 \\
6  & PRTA & 0.76 & 0.0536624624479547 &  0.0188212002478838 & Increased &  0.014304112188391688 \\
7  & ACAD & 0.73 & 0.0636856074394867 &  0.0288443452394158 & Increased &  0.021056372024773534 \\
8  & AXSM & 0.72 & 0.0598095872474592 &  0.0249683250473883 & Increased &  0.017977194034119577 \\
10 & NBIX & 0.63 & 0.0601683151682887 &  0.0253270529682178 & Increased &  0.015956043369977215 \\
\hline
\end{tabular}
}
\end{table}

\FloatBarrier

\subsubsection{BMS / MyoKardia}
Despite a negative trendline, BMS/MyoKardia is classified as strongly contradicting. The top-ranked firm CYTK (+14.3\% AR) is a large positive outlier, while all remaining high-similarity peers show negative or negligible ARs. This is a textbook outlier dominance case: the trendline direction is mechanically driven by a single observation and does not reflect a consistent relationship between similarity and peer stock reactions.

\begin{figure}[h!]
\centering
\includegraphics[width=0.75\textwidth]{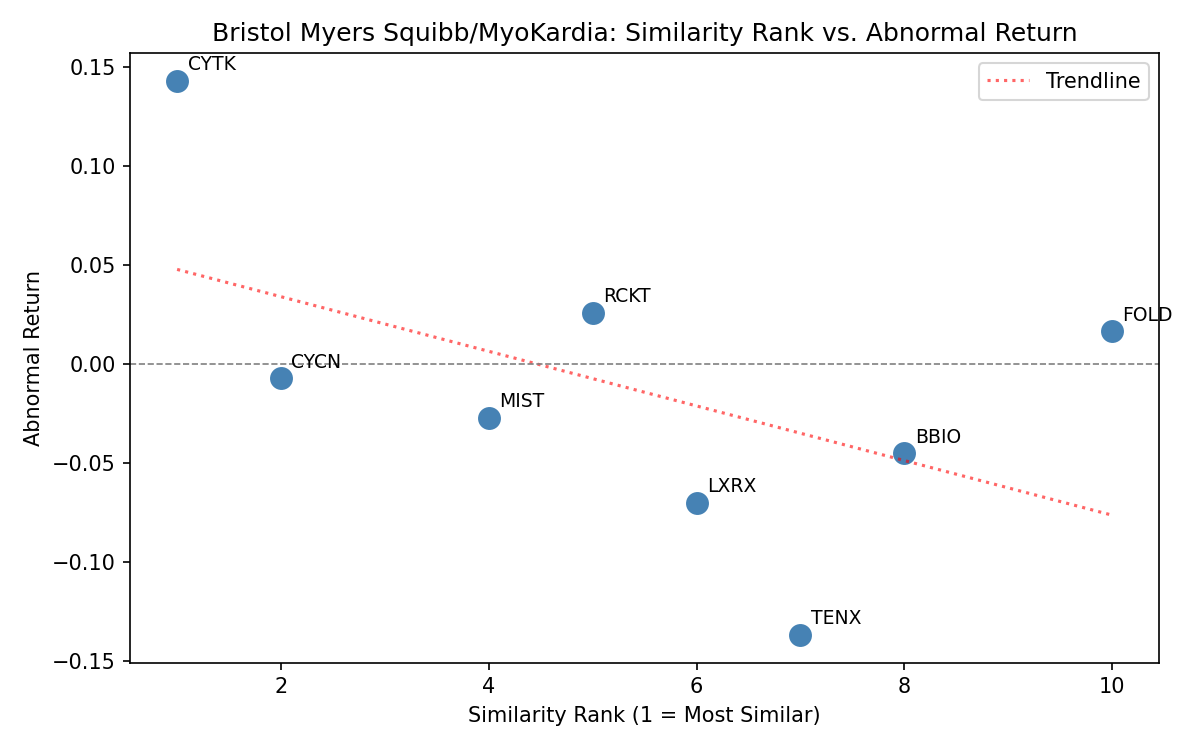}
\caption{BMS / MyoKardia Similarity vs. Abnormal Returns}
\label{fig:bms-myokardia}
\end{figure}

\begin{table}[h!]
\centering
\caption{Abnormal Returns Myokardia}
\resizebox{0.65\textwidth}{!}{%
\begin{tabular}{lrrl}
\hline
Ticker & Raw Return & Abnormal Return & Direction \\
\hline
BBIO &  0.004292979550332126 & -0.04509094642333769  & Decreased \\
CYCN &  0.04213942069154834  & -0.00724450528212147  & Decreased \\
CYTK &  0.19258027858739152  &  0.1431963526137217   & Increased \\
FOLD &  0.06607270764700231  &  0.016688781673332494 & Increased \\
LXRX & -0.02068971697413619  & -0.070073642947806    & Decreased \\
MIST &  0.02215658726588405  & -0.027227338707785764 & Decreased \\
RCKT &  0.075305691387476    &  0.02592176541380619  & Increased \\
TENX & -0.08759124087591241  & -0.13697516684958222  & Decreased \\
\hline
\end{tabular}
}
\end{table}

\begin{table}[h!]
\centering
\caption{Final Analysis Myokardia}
\resizebox{0.95\textwidth}{!}{
\begin{tabular}{r l r r r l r}
\hline
Rank & Ticker & Sim. Score & Raw Return & Abnormal Return & Direction & Shadow Signal \\
\hline
1  & CYTK & 0.94 &  0.1925802785873915 &  0.1431963526137217 & Increased &  0.13460457145689841 \\
2  & CYCN & 0.80 &  0.0421394206915483 & -0.0072445052821214 & Decreased & -0.00579560422569712 \\
4  & MIST & 0.69 &  0.0221565872658840 & -0.0272273387077857 & Decreased & -0.01878686370837213 \\
5  & RCKT & 0.63 &  0.0753056913874760 &  0.0259217654138061 & Increased &  0.016330712210697842 \\
6  & LXRX & 0.58 & -0.0206897169741361 & -0.0700736429478060 & Decreased & -0.04064271290972748 \\
7  & TENX & 0.50 & -0.0875912408759124 & -0.1369751668495822 & Decreased & -0.0684875834247911 \\
8  & BBIO & 0.42 &  0.0042929795503321 & -0.0450909464233376 & Decreased & -0.018938197497801793 \\
10 & FOLD & 0.30 &  0.0660727076470023 &  0.0166887816733324 & Increased &  0.00500663450199972 \\
\hline
\end{tabular}
}
\end{table}

\FloatBarrier

\subsubsection{BMS / Turning Point Therapeutics}
BMS/Turning Point Therapeutics is a clean supporting case. The top three peers (RLAY, RVMD, NUVL) all show positive abnormal returns, and the negative trendline is consistent across the cohort. This case, alongside AbbVie/ImmunoGen and Merck/Pandion, is among the clearest instances in which 10-K similarity does track peer reactions in a targeted oncology context---though only AbbVie/ImmunoGen reaches individual significance (Table~\ref{tab:spearman}).

\begin{figure}[h!]
\centering
\includegraphics[width=0.75\textwidth]{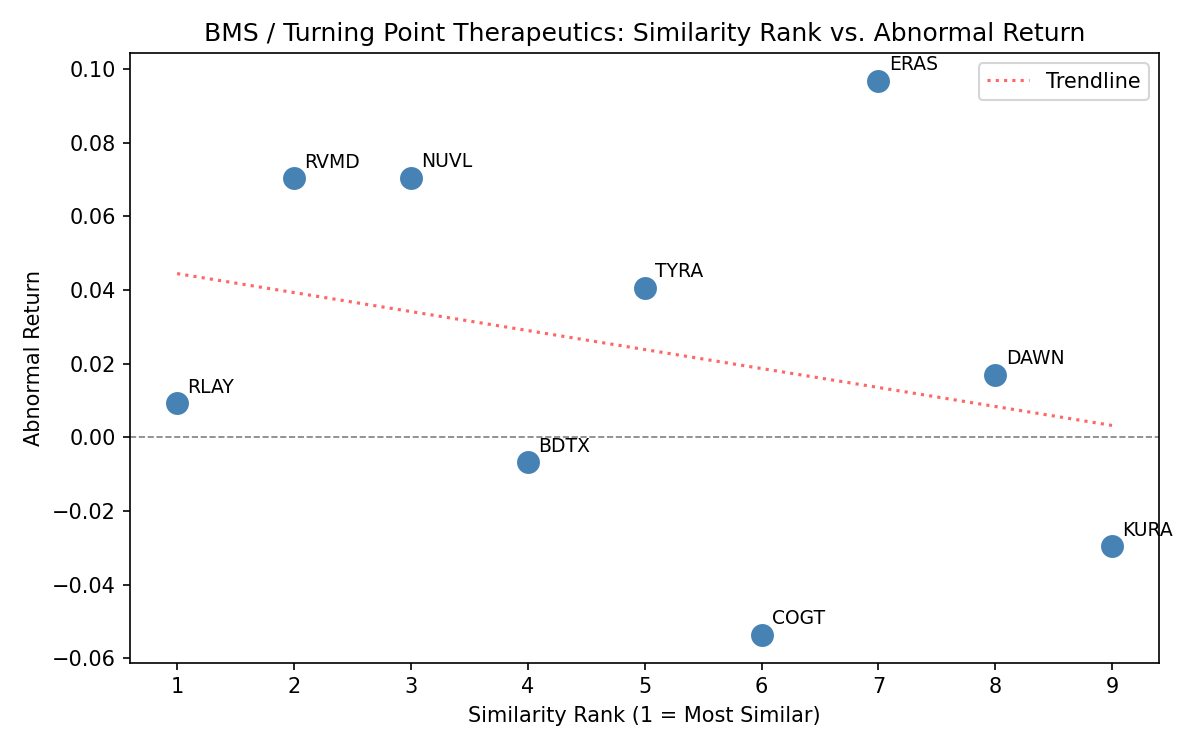}
\caption{BMS / Turning Point Therapeutics Similarity vs. Abnormal Returns}
\label{fig:bms-turning-point-therapeutics}
\end{figure}

\begin{table}[h!]
\centering
\caption{Abnormal Returns Turning Point Therapeutics}
\resizebox{0.65\textwidth}{!}{%
\begin{tabular}{lrrl}
\hline
Ticker & Raw Return & Abnormal Return & Direction \\
\hline
BDTX &  0.02873560462197694  & -0.006745240120843793 & Decreased \\
COGT & -0.018140572896618242 & -0.05362141763943898  & Decreased \\
DAWN &  0.052473748220260734 &  0.01699290347744     & Increased \\
ERAS &  0.13226460010465574  &  0.096783755361835    & Increased \\
KURA &  0.005930313199658817 & -0.029550531543161916 & Decreased \\
NUVL &  0.10589640389057788  &  0.07041555914775714  & Increased \\
RLAY &  0.0446960649171423   &  0.009215220174321567 & Increased \\
RVMD &  0.10578034707572571  &  0.07029950233290498  & Increased \\
TYRA &  0.07613469791143926  &  0.040653853168618526 & Increased \\
\hline
\end{tabular}
}
\end{table}

\begin{table}[h!]
\centering
\caption{Final Analysis Turning Point Therapeutics}
\resizebox{0.95\textwidth}{!}{
\begin{tabular}{r l r r r l r}
\hline
Rank & Ticker & Sim. Score & Raw Return & Abnormal Return & Direction & Shadow Signal \\
\hline
1 & RLAY & 0.92 &  0.0446960649171423 &  0.0092152201743215 & Increased &  0.00847800256037578 \\
2 & RVMD & 0.91 &  0.1057803470757257 &  0.0702995023329049 & Increased &  0.06397254712294347 \\
3 & NUVL & 0.90 &  0.1058964038905778 &  0.0704155591477571 & Increased &  0.0633740032329814 \\
4 & BDTX & 0.87 &  0.0287356046219769 & -0.0067452401208437 & Decreased & -0.005868358905134019 \\
5 & TYRA & 0.83 &  0.0761346979114392 &  0.0406538531686185 & Increased &  0.03374269812995335 \\
6 & COGT & 0.82 & -0.0181405728966182 & -0.0536214176394389 & Decreased & -0.0439695624643399 \\
7 & ERAS & 0.81 &  0.1322646001046557 &  0.0967837553618350 & Increased &  0.07839484184308636 \\
8 & DAWN & 0.77 &  0.0524737482202607 &  0.0169929034774400 & Increased &  0.013084535677628802 \\
9 & KURA & 0.76 &  0.0059303131996588 & -0.0295505315431619 & Decreased & -0.022458403972803045 \\
\hline
\end{tabular}
}
\end{table}

\FloatBarrier

\subsubsection{BMS / Mirati Therapeutics}
BMS/Mirati Therapeutics is classified as contradicting despite a negative trendline. The top-3 AR direction is mixed ($+,-,+$), and the second-ranked firm (TNGX, $-12.5\%$ AR) shows a large negative reaction that disrupts any clean positive signal. The negative trendline is driven by the interaction between RVMD (high similarity, positive AR) and TNGX (high similarity, strongly negative AR), producing a noisy, uninterpretable pattern overall.

\begin{figure}[h!]
\centering
\includegraphics[width=0.75\textwidth]{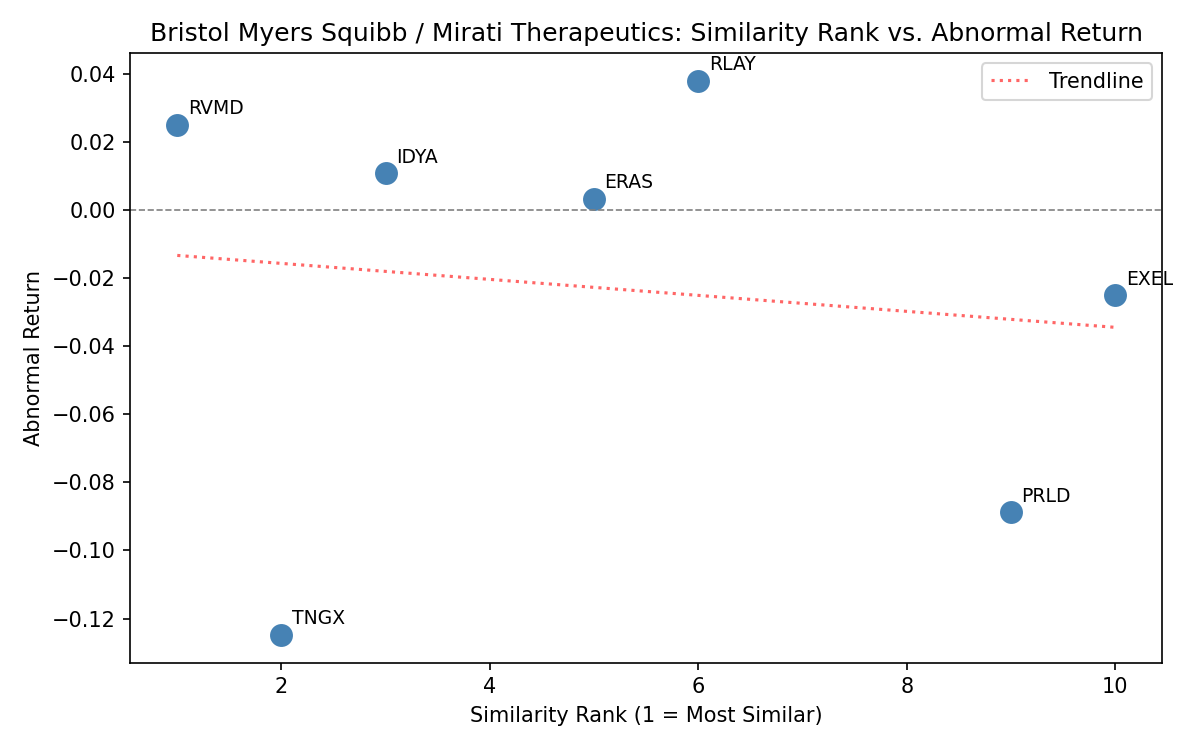}
\caption{BMS / Mirati Therapeutics Similarity vs. Abnormal Returns}
\label{fig:bms-mirati-therapeutics}
\end{figure}

\begin{table}[h!]
\centering
\caption{Abnormal Returns Mirati Therapeutics}
\resizebox{0.65\textwidth}{!}{%
\begin{tabular}{lrrl}
\hline
Ticker & Raw Return & Abnormal Return & Direction \\
\hline
ERAS & -0.004524882484455325 &  0.0033664856347955076 & Increased \\
EXEL & -0.03285326574234348  & -0.024961897623092644  & Decreased \\
IDYA &  0.00299962217762286  &  0.010890990296873692  & Increased \\
PRLD & -0.09665426949169337  & -0.08876290137244254   & Decreased \\
RLAY &  0.030156815579058048 &  0.03804818369830888   & Increased \\
RVMD &  0.017172459349252746 &  0.02506382746850358   & Increased \\
TNGX & -0.13268894672203788  & -0.12479757860278705   & Decreased \\
\hline
\end{tabular}
}
\end{table}

\begin{table}[h!]
\centering
\caption{Final Analysis Mirati Therapeutics}
\resizebox{0.95\textwidth}{!}{
\begin{tabular}{r l r r r l r}
\hline
Rank & Ticker & Sim. Score & Raw Return & Abnormal Return & Direction & Shadow Signal \\
\hline
1  & RVMD & 0.91 &  0.0171724593492527 &  0.0250638274685035 & Increased &  0.022808082996338186 \\
2  & TNGX & 0.85 & -0.1326889467220378 & -0.1247975786027870 & Decreased & -0.10607794181236894 \\
3  & IDYA & 0.84 &  0.0029996221776228 &  0.0108909902968736 & Increased &  0.009148431849373823 \\
5  & ERAS & 0.80 & -0.0045248824844553 &  0.0033664856347955 & Increased &  0.0026931885078364 \\
6  & RLAY & 0.78 &  0.0301568155790580 &  0.0380481836983088 & Increased &  0.029677583284680863 \\
9  & PRLD & 0.58 & -0.0966542694916933 & -0.0887629013724425 & Decreased & -0.051482482796016645 \\
10 & EXEL & 0.45 & -0.0328532657423434 & -0.0249618976230926 & Decreased & -0.01123285393039167 \\
\hline
\end{tabular}
}
\end{table}

\FloatBarrier

\subsubsection{Gilead / Forty Seven}
Gilead/Forty Seven is classified as contradicting. The trendline is positive (higher similarity correlates with lower ARs), and the top-ranked firm shows a negative abnormal return. The case is notable because IPHA, ranked ninth, shows the largest positive AR ($+9.2\%$), inverting the expected pattern entirely.

\begin{figure}[h!]
\centering
\includegraphics[width=0.75\textwidth]{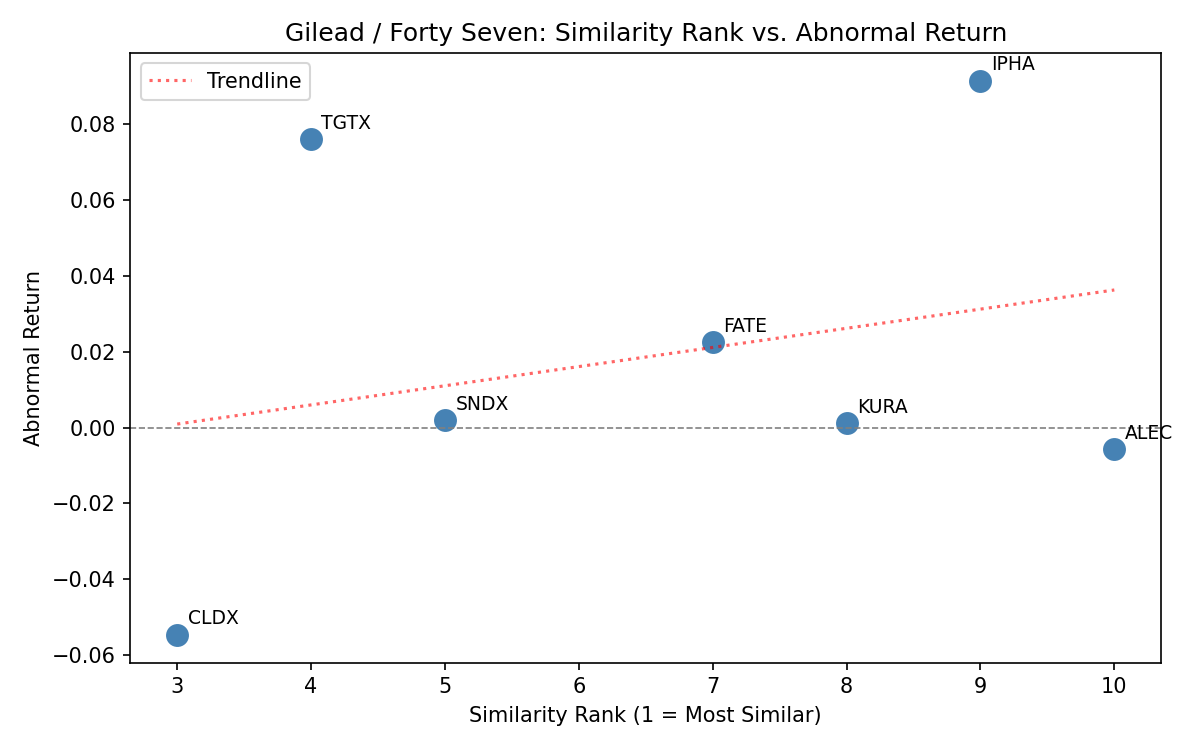}
\caption{Gilead / Forty Seven Similarity vs. Abnormal Returns}
\label{fig:gilead-forty-seven}
\end{figure}

\begin{table}[h!]
\centering
\caption{Abnormal Returns Forty Seven}
\resizebox{0.65\textwidth}{!}{%
\begin{tabular}{lrrl}
\hline
Ticker & Raw Return & Abnormal Return & Direction \\
\hline
ALEC &  0.03384280642895883  & -0.005628199429125465 & Decreased \\
CLDX & -0.015209200962377005 & -0.0546802068204613  & Decreased \\
FATE &  0.06198628146063721  &  0.02251527560255292  & Increased \\
IPHA &  0.1309903617959911   &  0.0915193559379068   & Increased \\
KURA &  0.040596581373783934 &  0.0011255755156996394 & Increased \\
SNDX &  0.041445308145250374 &  0.00197430228716608  & Increased \\
TGTX &  0.11553783165163699  &  0.0760668257935527   & Increased \\
\hline
\end{tabular}
}
\end{table}

\begin{table}[h!]
\centering
\caption{Final Analysis Forty Seven}
\resizebox{0.95\textwidth}{!}{
\begin{tabular}{r l r r r l r}
\hline
Rank & Ticker & Sim. Score & Raw Return & Abnormal Return & Direction & Shadow Signal \\
\hline
3  & CLDX & 0.84 & -0.0152092009623770 & -0.0546802068204613 & Decreased & -0.045931373729187486 \\
4  & TGTX & 0.80 &  0.1155378316516369 &  0.0760668257935527 & Increased &  0.06085346063484216 \\
5  & SNDX & 0.75 &  0.0414453081452503 &  0.0019743022871660 & Increased &  0.0014807267153745 \\
7  & FATE & 0.69 &  0.0619862814606372 &  0.0225152756025529 & Increased &  0.0155355401657615 \\
8  & KURA & 0.59 &  0.0405965813737839 &  0.0011255755156996 & Increased &  0.0006640895542627639 \\
9  & IPHA & 0.58 &  0.1309903617959911 &  0.0915193559379068 & Increased &  0.05308122644398594 \\
10 & ALEC & 0.55 &  0.0338428064289588 & -0.0056281994291254 & Decreased & -0.00309550968601897 \\
\hline
\end{tabular}
}
\end{table}

\FloatBarrier

\subsubsection{Merck / Acceleron Pharma}
Merck/Acceleron Pharma is strongly contradicting. Despite a negative trendline, the top-ranked peers show near-zero or negative ARs, while the two firms with the highest and only positive ARs (SRRK, PTGX) rank in the middle of the similarity distribution. The announcement-day signal is weak across the cohort and does not concentrate in the most similar firms.

\begin{figure}[h!]
\centering
\includegraphics[width=0.75\textwidth]{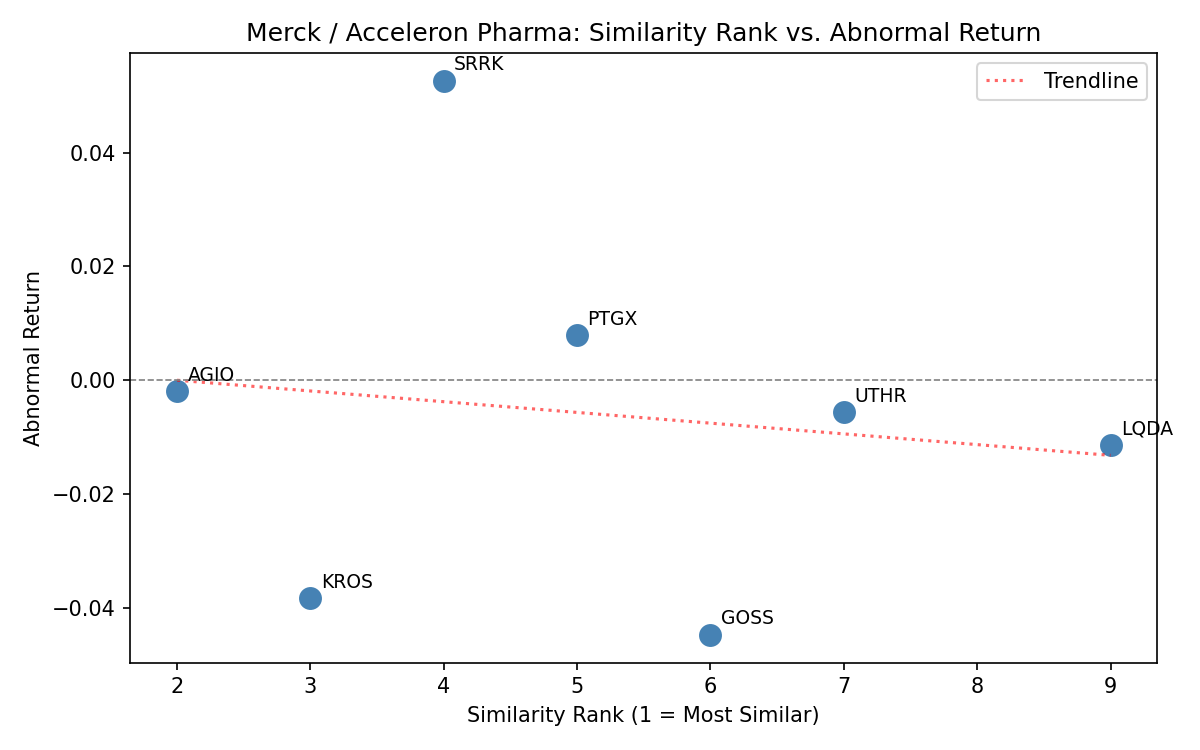}
\caption{Merck / Acceleron Pharma Similarity vs. Abnormal Returns}
\label{fig:merck-acceleron-pharma}
\end{figure}

\begin{table}[h!]
\centering
\caption{Abnormal Returns Acceleron Pharma}
\resizebox{0.65\textwidth}{!}{%
\begin{tabular}{lrrl}
\hline
Ticker & Raw Return & Abnormal Return & Direction \\
\hline
AGIO & -0.0012983674482242133 & -0.0019350482482927741 & Decreased \\
GOSS & -0.04410645935599469   & -0.04474314015606325   & Decreased \\
KROS & -0.03770370260394145   & -0.03834038340401001   & Decreased \\
LQDA & -0.010752678064499305  & -0.011389358864567866  & Decreased \\
PTGX &  0.008537257890474546  &  0.007900577090405985  & Increased \\
SRRK &  0.05326953926549374   &  0.05263285846542518   & Increased \\
UTHR & -0.004959558862238881  & -0.005596239662307442  & Decreased \\
\hline
\end{tabular}
}
\end{table}

\begin{table}[h!]
\centering
\caption{Final Analysis Acceleron Pharma}
\resizebox{0.95\textwidth}{!}{
\begin{tabular}{r l r r r l r}
\hline
Rank & Ticker & Sim. Score & Raw Return & Abnormal Return & Direction & Shadow Signal \\
\hline
2  & AGIO & 0.70 & -0.0012983674482242 & -0.0019350482482927 & Decreased & -0.00135453377380489 \\
3  & KROS & 0.68 & -0.0377037026039414 & -0.0383403834040100 & Decreased & -0.0260714607147268 \\
4  & SRRK & 0.65 &  0.0532695392654937 &  0.0526328584654251 & Increased &  0.03421135800252632 \\
5  & PTGX & 0.60 &  0.0085372578904745 &  0.0079005770904059 & Increased &  0.00474034625424354 \\
6  & GOSS & 0.60 & -0.0441064593559946 & -0.0447431401560632 & Decreased & -0.026845884093637917 \\
7  & UTHR & 0.50 & -0.0049595588622388 & -0.0055962396623074 & Decreased & -0.0027981198311537 \\
9  & LQDA & 0.40 & -0.0107526780644993 & -0.0113893588645678 & Decreased & -0.00455574354582712 \\
\hline
\end{tabular}
}
\end{table}

\FloatBarrier

\subsubsection{Merck / Pandion Therapeutics}
Merck/Pandion Therapeutics is a strongly supporting case. All top-ranked peers show positive abnormal returns, and the pattern holds across the full cohort: six of seven firms show positive ARs, with the negative case (PTGX) ranking sixth. This is one of the most consistent results in the dataset and suggests that in immuno-oncology acquisitions with tightly defined peer groups, textual similarity can reliably predict announcement-day reactions.

\begin{figure}[h!]
\centering
\includegraphics[width=0.75\textwidth]{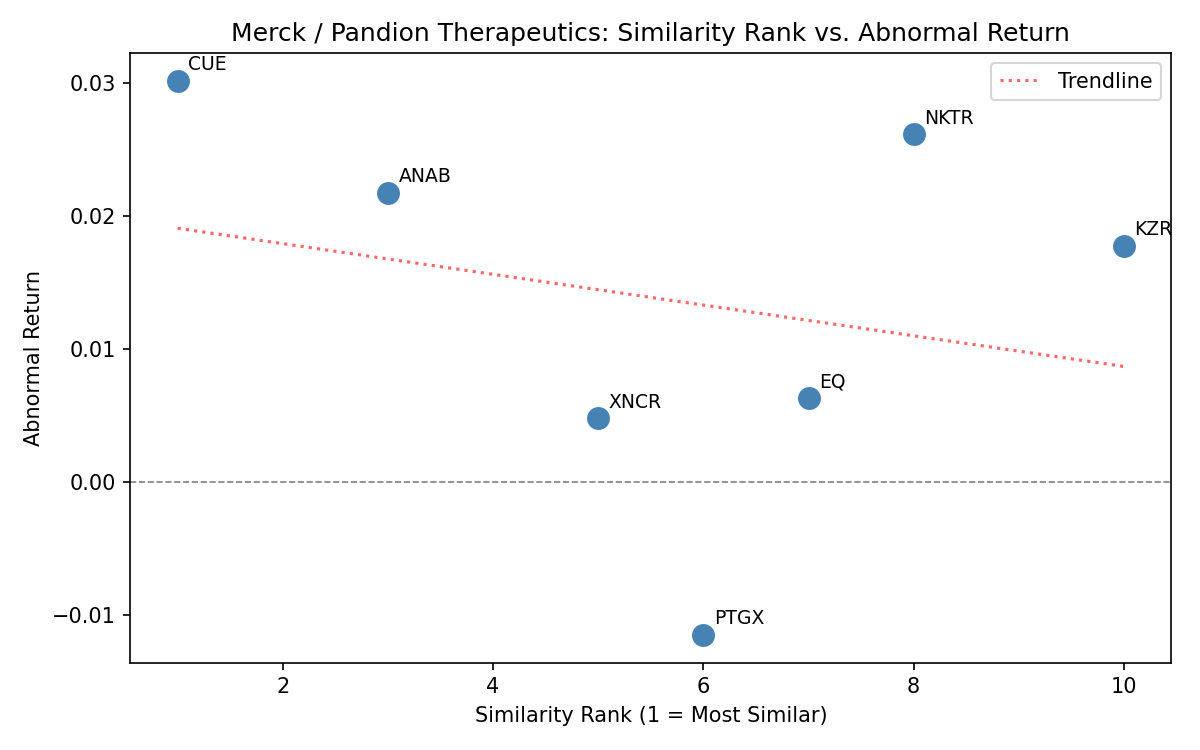}
\caption{Merck / Pandion Therapeutics Similarity vs. Abnormal Returns}
\label{fig:merck-pandion-therapeutics}
\end{figure}

\begin{table}[h!]
\centering
\caption{Abnormal Returns Pandion Therapeutics}
\resizebox{0.65\textwidth}{!}{%
\begin{tabular}{lrrl}
\hline
Ticker & Raw Return & Abnormal Return & Direction \\
\hline
ANAB & -0.01888217555300506  &  0.021744613828727936 & Increased \\
CUE  & -0.010471229956700903 &  0.03015555942503209  & Increased \\
EQ   & -0.03435114202800781  &  0.006275647353725183 & Increased \\
KZR  & -0.022887310819052452 &  0.017739478562680544 & Increased \\
NKTR & -0.014492700531008559 &  0.026134088850724435 & Increased \\
PTGX & -0.052153288134218864 & -0.011526498752485868 & Decreased \\
XNCR & -0.03587632326735664  &  0.004750466114376357 & Increased \\
\hline
\end{tabular}
}
\end{table}

\begin{table}[h!]
\centering
\caption{Final Analysis Pandion Therapeutics}
\resizebox{0.95\textwidth}{!}{
\begin{tabular}{r l r r r l r}
\hline
Rank & Ticker & Sim. Score & Raw Return & Abnormal Return & Direction & Shadow Signal \\
\hline
1  & CUE  & 0.94 & -0.0104712299567009 &  0.0301555594250320 & Increased &  0.028346225859530078 \\
3  & ANAB & 0.83 & -0.0188821755530050 &  0.0217446138287279 & Increased &  0.018048029477844157 \\
5  & XNCR & 0.80 & -0.0358763232673566 &  0.0047504661143763 & Increased &  0.0038003728915010403 \\
6  & PTGX & 0.79 & -0.0521532881342188 & -0.0115264987524858 & Decreased & -0.009105934014463782 \\
7  & EQ   & 0.78 & -0.0343511420280078 &  0.0062756473537251 & Increased &  0.004895004935905578 \\
8  & NKTR & 0.76 & -0.0144927005310085 &  0.0261340888507244 & Increased &  0.019861907526550544 \\
10 & KZR  & 0.72 & -0.0228873108190524 &  0.0177394785626805 & Increased &  0.01277242456512996 \\
\hline
\end{tabular}
}
\end{table}

\FloatBarrier

\subsection{Consumer Staples}

\subsubsection{Amazon / Whole Foods}
Amazon/Whole Foods is classified as supporting despite all peers showing negative ARs. The key is the trendline: higher-similarity grocery peers (SFM, NGVC) decline more than lower-similarity peers, which is consistent with the shadow trading mechanism in reverse---the acquisition of a major competitor signals competitive pressure rather than acquisition premium for peers. The negative trendline remains directionally consistent with the hypothesis.

\begin{figure}[h!]
\centering
\includegraphics[width=0.75\textwidth]{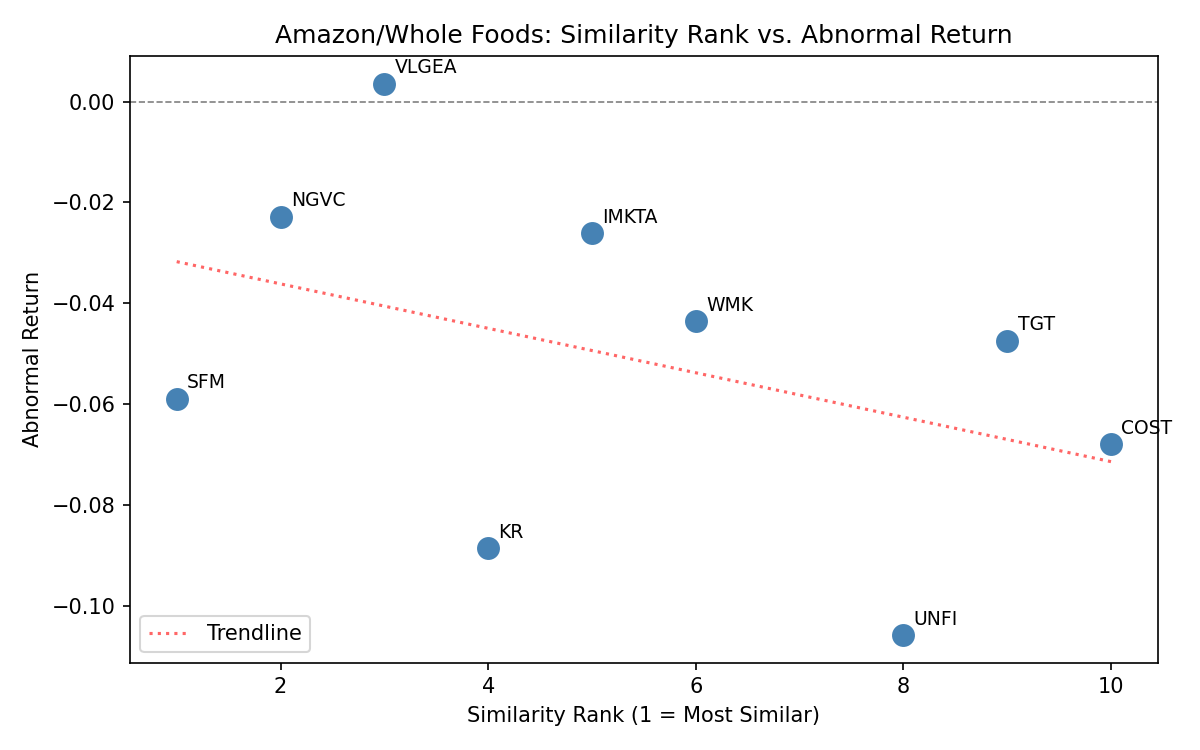}
\caption{Amazon / Whole Foods Similarity vs. Abnormal Returns}
\label{fig:amazon-whole-foods}
\end{figure}

\begin{table}[h!]
\centering
\caption{Abnormal Returns Whole Foods}
\resizebox{0.65\textwidth}{!}{
\begin{tabular}{lrrl}
\hline
Ticker & Raw Return & Abnormal Return & Direction \\
\hline
COST  & -0.07192040374481146  & -0.06796522838659612  & Decreased \\
IMKTA & -0.03000031493783293  & -0.026045139579617585 & Decreased \\
KR    & -0.09242663874926776  & -0.08847146339105243  & Decreased \\
NGVC  & -0.026755852092367225 & -0.022800676734151884 & Decreased \\
SFM   & -0.06289026951890986  & -0.058935094160694514 & Decreased \\
TGT   & -0.05138819614229597  & -0.047433020784080626 & Decreased \\
UNFI  & -0.10968554994595126  & -0.10573037458773592  & Decreased \\
VLGEA & -0.00039950769994962914 & 0.003555667658265714 & Increased \\
WMK   & -0.04752578168203486  & -0.04357060632381952  & Decreased \\
\hline
\end{tabular}
}
\end{table}

\begin{table}[h!]
\centering
\caption{Final Analysis Whole Foods}
\resizebox{0.95\textwidth}{!}{
\begin{tabular}{r l r r r l r}
\hline
Rank & Ticker & Sim. Score & Raw Return & Abnormal Return & Direction & Shadow Signal \\
\hline
1  & SFM   & 0.92 & -0.0628902695189098 & -0.0589350941606945 & Decreased & -0.05422028662783894 \\
2  & NGVC  & 0.90 & -0.0267558520923672 & -0.0228006767341518 & Decreased & -0.020520609060736623 \\
3  & VLGEA & 0.83 & -0.0003995076999496 &  0.0035556676582657 & Increased &  0.0029512041563605307 \\
4  & KR    & 0.76 & -0.0924266387492677 & -0.0884714633910524 & Decreased & -0.06723831217719982 \\
5  & IMKTA & 0.65 & -0.0300003149378329 & -0.0260451395796175 & Decreased & -0.016929340726751375 \\
6  & WMK   & 0.60 & -0.0475257816820348 & -0.0435706063238195 & Decreased & -0.026142363794291697 \\
8  & UNFI  & 0.50 & -0.1096855499459512 & -0.1057303745877359 & Decreased & -0.05286518729386795 \\
9  & TGT   & 0.45 & -0.0513881961422959 & -0.0474330207840806 & Decreased & -0.02134485935283627 \\
10 & COST  & 0.35 & -0.0719204037448114 & -0.0679652283865961 & Decreased & -0.023787829935308636 \\
\hline
\end{tabular}
}
\end{table}

\FloatBarrier

\subsubsection{Danone / WhiteWave Foods}
Danone/WhiteWave Foods is a supporting case. The top-ranked firm (HAIN, $+6.9\%$ AR) shows a clear positive reaction, and the negative trendline holds moderately across the cohort. Several mid- and lower-ranked firms show slightly negative but negligible reactions, consistent with a weaker similarity-return relationship in consumer staples compared to biopharmaceuticals.

\begin{figure}[h!]
\centering
\includegraphics[width=0.75\textwidth]{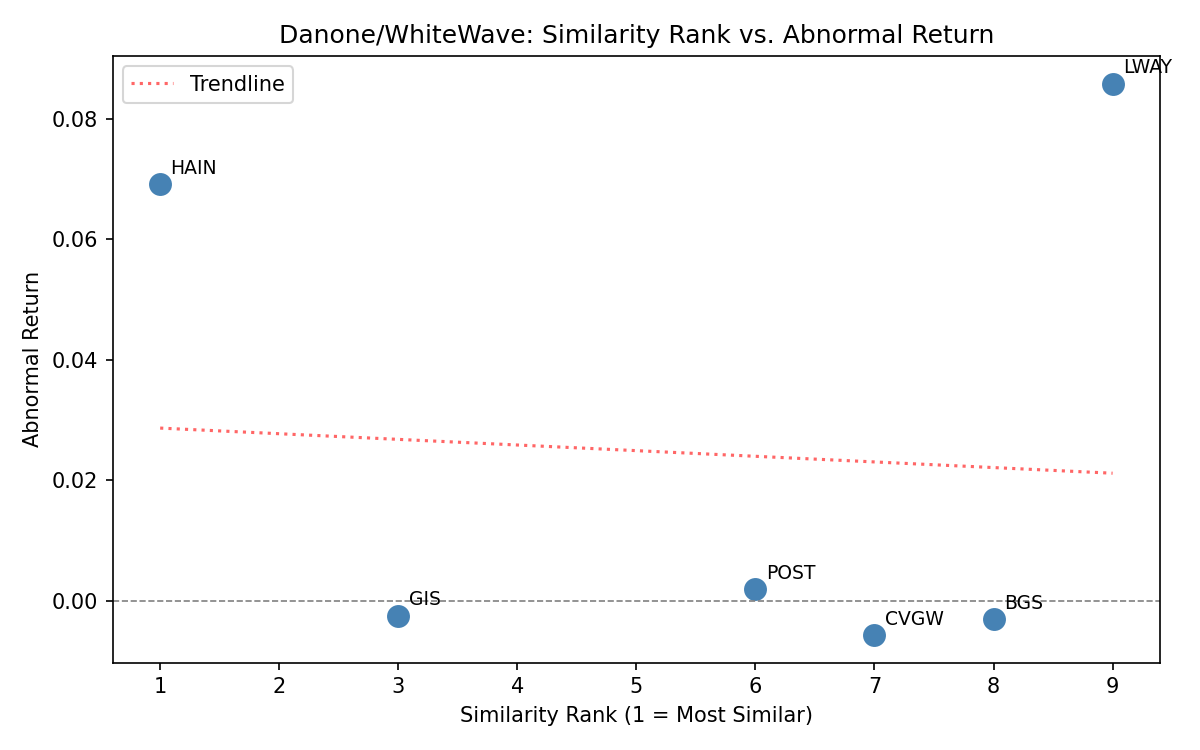}
\caption{Danone / WhiteWave Foods Similarity vs. Abnormal Returns}
\label{fig:danone-whitewave-foods}
\end{figure}

\begin{table}[h!]
\centering
\caption{Abnormal Returns WhiteWave Foods}
\resizebox{0.65\textwidth}{!}{%
\begin{tabular}{lrrl}
\hline
Ticker & Raw Return & Abnormal Return & Direction \\
\hline
BGS  & -0.001464230372955497 & -0.0031091979285966117 & Decreased \\
CVGW & -0.004072587271339847 & -0.005717554826980962  & Decreased \\
DFP  &  0.00041192300467115514 & -0.0012330445509699595 & Decreased \\
GIS  & -0.0008324021591602719 & -0.0024773697148013866 & Decreased \\
HAIN &  0.07085453416477798  &  0.06920956660913687   & Increased \\
LWAY &  0.08752332905866998  &  0.08587836150302887   & Increased \\
POST &  0.0035473528668403583 & 0.0019023853111992436 & Increased \\
\hline
\end{tabular}
}
\end{table}

\begin{table}[h!]
\centering
\caption{Final Analysis WhiteWave Foods}
\resizebox{0.95\textwidth}{!}{
\begin{tabular}{r l r r r l r}
\hline
Rank & Ticker & Sim. Score & Raw Return & Abnormal Return & Direction & Shadow Signal \\
\hline
1 & HAIN & 0.87 &  0.0708545341647779 &  0.0692095666091368 & Increased &  0.060212322949949014 \\
3 & GIS  & 0.76 & -0.0008324021591602 & -0.0024773697148013 & Decreased & -0.0018828009832489879 \\
6 & POST & 0.69 &  0.0035473528668403 &  0.0019023853111992 & Increased &  0.001312645864727448 \\
7 & CVGW & 0.68 & -0.0040725872713398 & -0.0057175548269809 & Decreased & -0.0038879372823470123 \\
8 & BGS  & 0.64 & -0.0014642303729554 & -0.0031091979285966 & Decreased & -0.001989886674301824 \\
9 & LWAY & 0.63 &  0.0875233290586699 &  0.0858783615030288 & Increased &  0.054103367746908146 \\
\hline
\end{tabular}
}
\end{table}

\FloatBarrier

\subsubsection{JM Smucker / Hostess}
JM Smucker/Hostess is a clean supporting case. All five peers for which data was available show positive abnormal returns, and the top three (FLO, JJSF, THS) show the largest gains, consistent with the shadow trading hypothesis. This is one of the strongest results in the consumer staples subsample.

\begin{figure}[h!]
\centering
\includegraphics[width=0.75\textwidth]{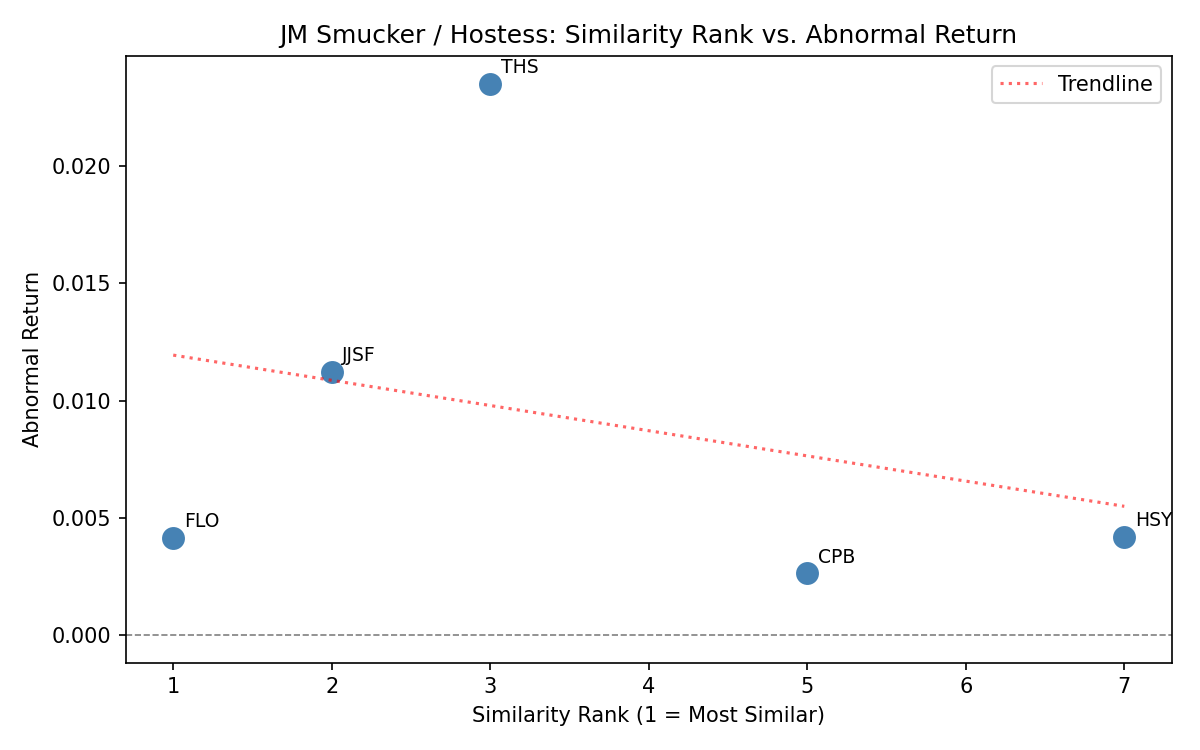}
\caption{JM Smucker / Hostess Similarity vs. Abnormal Returns}
\label{fig:jm-smucker-hostess}
\end{figure}

\begin{table}[h!]
\centering
\caption{Abnormal Returns Hostess}
\resizebox{0.65\textwidth}{!}{%
\begin{tabular}{lrrl}
\hline
Ticker & Raw Return & Abnormal Return & Direction \\
\hline
CPB  &  0.012802546888889169 &  0.00263678770741119  & Increased \\
FLO  &  0.014310556528568134 &  0.004144797347090156 & Increased \\
GIS  & -0.003930505467713301 & -0.01409626464919128  & Decreased \\
HSY  &  0.01435192480057275  &  0.004186165619094772 & Increased \\
JJSF &  0.021397606391074405 &  0.011231847209596427 & Increased \\
POST &  0.004609905973885611 & -0.005555853207592367 & Decreased \\
THS  &  0.033685659526077175 &  0.023519900344599196 & Increased \\
\hline
\end{tabular}
}
\end{table}

\begin{table}[h!]
\centering
\caption{Final Analysis Hostess}
\resizebox{0.95\textwidth}{!}{
\begin{tabular}{r l r r r l r}
\hline
Rank & Ticker & Sim. Score & Raw Return & Abnormal Return & Direction & Shadow Signal \\
\hline
1 & FLO  & 0.92 & 0.0143105565285681 & 0.0041447973470901 & Increased & 0.0038132135593228924 \\
2 & JJSF & 0.86 & 0.0213976063910744 & 0.0112318472095964 & Increased & 0.009659388600252904 \\
3 & THS  & 0.82 & 0.0336856595260771 & 0.0235199003445991 & Increased & 0.019286318282571262 \\
5 & CPB  & 0.78 & 0.0128025468888891 & 0.0026367877074111 & Increased & 0.002056694411780658 \\
7 & HSY  & 0.72 & 0.0143519248005727 & 0.0041861656190947 & Increased & 0.003014039245748184 \\
\hline
\end{tabular}
}
\end{table}

\FloatBarrier

\subsubsection{Kroger / Albertsons}
Kroger/Albertsons is classified as contradicting. Despite a negative trendline, the top-3 AR direction is mixed ($+,-,+$), and the overall distribution shows no consistent relationship between similarity rank and return magnitude. Regulatory uncertainty surrounding the proposed merger likely introduced idiosyncratic noise that overwhelmed any fundamental peer signal.

\begin{figure}[h!]
\centering
\includegraphics[width=0.75\textwidth]{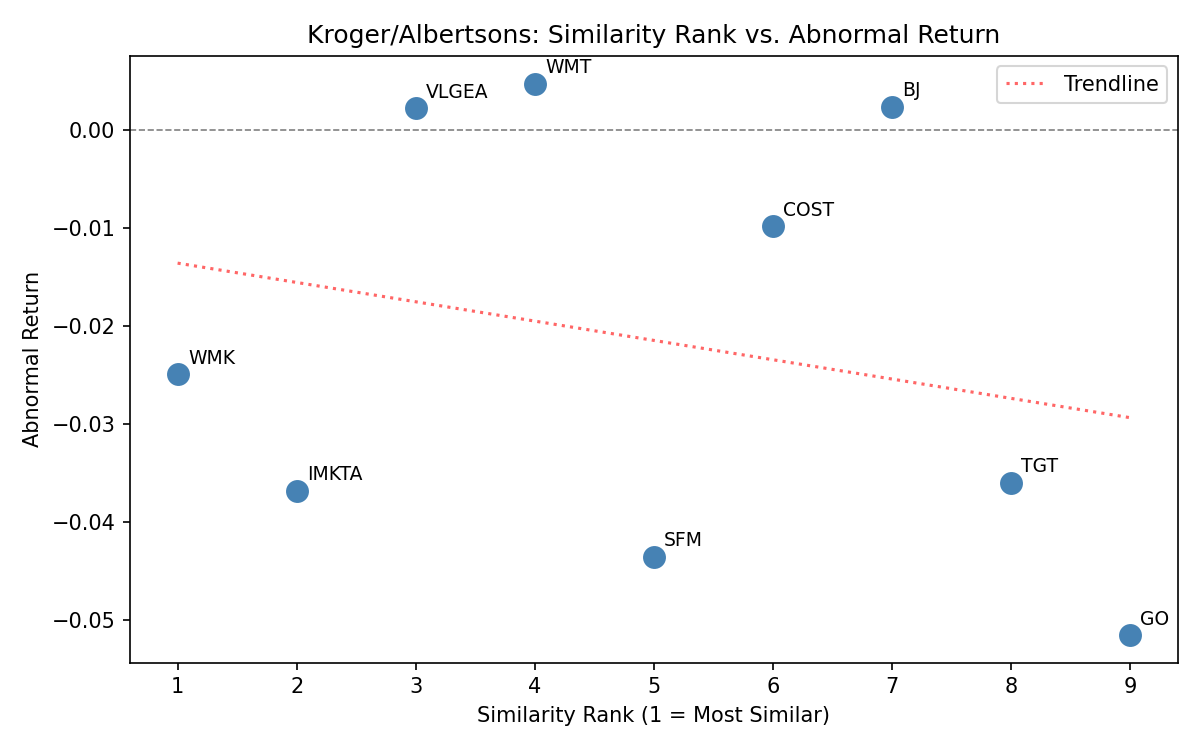}
\caption{Kroger / Albertsons Similarity vs. Abnormal Returns}
\label{fig:kroger-albertsons}
\end{figure}

\begin{table}[h!]
\centering
\caption{Abnormal Returns Albertsons}
\resizebox{0.65\textwidth}{!}{%
\begin{tabular}{lrrl}
\hline
Ticker & Raw Return & Abnormal Return & Direction \\
\hline
BJ    & -0.016341231410284245 &  0.002392864995054235  & Increased \\
COST  & -0.02850488629098535  & -0.00977078988564687   & Decreased \\
GO    & -0.07023213268824618  & -0.05149803628290771   & Decreased \\
IMKTA & -0.055493940265775386 & -0.0367598438604369    & Decreased \\
SFM   & -0.06223925247850891  & -0.04350515607317043   & Decreased \\
TGT   & -0.05469759121142679  & -0.035963494806088306  & Decreased \\
VLGEA & -0.01649676540765968  &  0.0022373309976788    & Increased \\
WMK   & -0.04363704397423403  & -0.024902947568895547  & Decreased \\
WMT   & -0.013985591718419774 &  0.0047485046869187066 & Increased \\
\hline
\end{tabular}
}
\end{table}

\begin{table}[h!]
\centering
\caption{Final Analysis Albertsons}
\resizebox{0.95\textwidth}{!}{
\begin{tabular}{r l r r r l r}
\hline
Rank & Ticker & Sim. Score & Raw Return & Abnormal Return & Direction & Shadow Signal \\
\hline
1 & WMK   & 0.94 & -0.0436370439742340 & -0.0249029475688955 & Decreased & -0.023408770714761766 \\
2 & IMKTA & 0.89 & -0.0554939402657753 & -0.0367598438604369 & Decreased & -0.03271626103578884 \\
3 & VLGEA & 0.88 & -0.0164967654076596 &  0.0022373309976788 & Increased &  0.001968851277957344 \\
4 & WMT   & 0.76 & -0.0139855917184197 &  0.0047485046869187 & Increased &  0.003608863562058212 \\
5 & SFM   & 0.68 & -0.0622392524785089 & -0.0435051560731704 & Decreased & -0.029583506129755875 \\
6 & COST  & 0.65 & -0.0285048862909853 & -0.0097707898856468 & Decreased & -0.00635101342567042 \\
7 & BJ    & 0.63 & -0.0163412314102842 &  0.0023928649950542 & Increased &  0.001507504946884146 \\
8 & TGT   & 0.58 & -0.0546975912114267 & -0.0359634948060883 & Decreased & -0.020858826987531214 \\
9 & GO    & 0.45 & -0.0702321326882461 & -0.0514980362829077 & Decreased & -0.023174116327308467 \\
\hline
\end{tabular}
}
\end{table}

\FloatBarrier

\subsubsection{Reckitt Benckiser / Mead Johnson Nutrition}
Reckitt Benckiser/Mead Johnson is contradicting. The trendline is positive (primarily due to the outlier), but the top-ranked firm (ABT) shows essentially zero abnormal return. The acquisition of an infant nutrition company does not appear to have generated meaningful peer firm reactions, suggesting limited economic linkage within the consumer staples peer group as identified from 10-K text.

\begin{figure}[h!]
\centering
\includegraphics[width=0.75\textwidth]{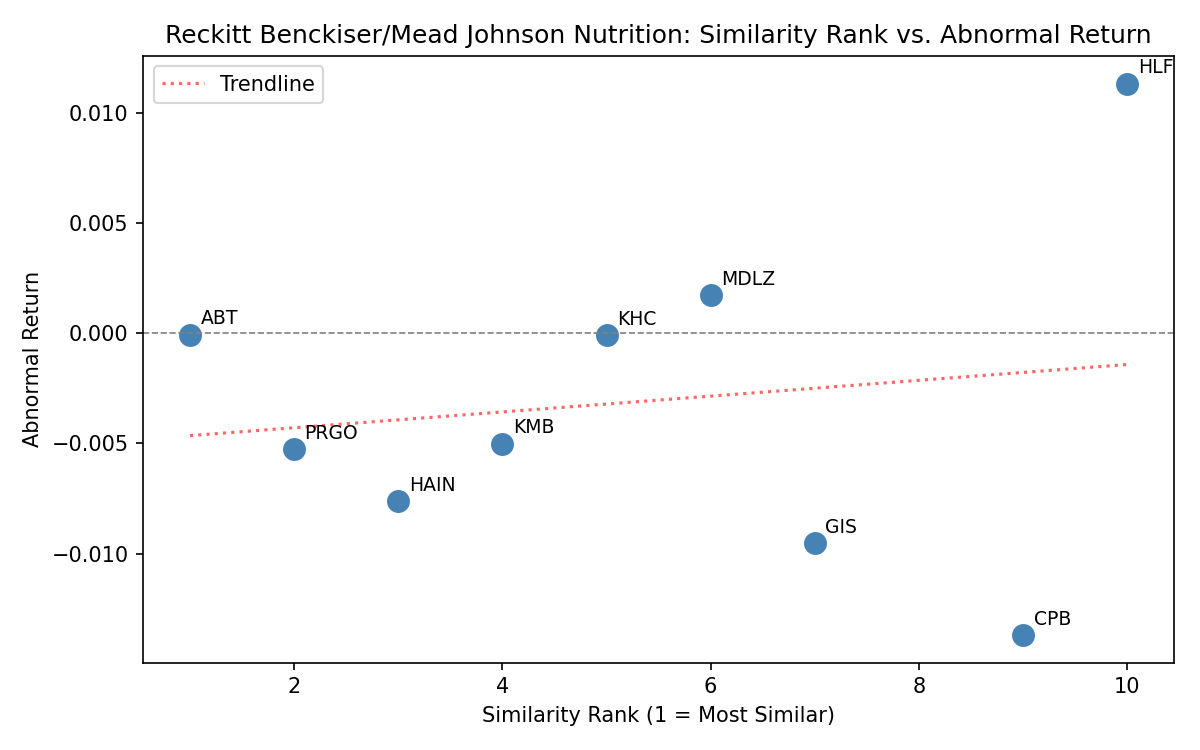}
\caption{Reckitt Benckiser / Mead Johnson Nutrition Similarity vs. Abnormal Returns}
\label{fig:reckitt-benckiser-mead-johnson-nutrition}
\end{figure}

\begin{table}[h!]
\centering
\caption{Abnormal Returns Mead Johnson Nutrition}
\resizebox{0.65\textwidth}{!}{%
\begin{tabular}{lrrl}
\hline
Ticker & Raw Return & Abnormal Return & Direction \\
\hline
ABT  &  0.00470164327620931   & -6.045667372484714e-05 & Decreased \\
CPB  & -0.008928184876190674  & -0.013690284826124832  & Decreased \\
GIS  & -0.004757503745379787  & -0.009519603695313943  & Decreased \\
HAIN & -0.0028468067310845843 & -0.007608906681018742  & Decreased \\
HLF  &  0.016090798580756894  &  0.011328698630822736  & Increased \\
KHC  &  0.004677904408272617  & -8.419554166153986e-05 & Decreased \\
KMB  & -0.00024270730659042898 & -0.005004807256524586 & Decreased \\
MDLZ &  0.006487801983811118  &  0.001725702033876961  & Increased \\
PRGO & -0.0005037224269522213 & -0.005265822376886378  & Decreased \\
\hline
\end{tabular}
}
\end{table}

\begin{table}[h!]
\centering
\caption{Final Analysis Mead Johnson Nutrition}
\resizebox{0.95\textwidth}{!}{
\begin{tabular}{r l r r r l r}
\hline
Rank & Ticker & Sim. Score & Raw Return & Abnormal Return & Direction & Shadow Signal \\
\hline
1  & ABT  & 0.88 &  0.0047016432762093 & -6.045667372484714e-05 & Decreased & -5.320187287786549e-05 \\
2  & PRGO & 0.80 & -0.0005037224269522 & -0.0052658223768863    & Decreased & -0.00421265790150904 \\
3  & HAIN & 0.60 & -0.0028468067310845 & -0.0076089066810187    & Decreased & -0.00456534400861122 \\
4  & KMB  & 0.55 & -0.0002427073065904 & -0.0050048072565245    & Decreased & -0.0027526439910884755 \\
5  & KHC  & 0.49 &  0.0046779044082726 & -8.419554166153986e-05 & Decreased & -4.125581541415453e-05 \\
6  & MDLZ & 0.42 &  0.0064878019838111 &  0.0017257020338769    & Increased &  0.000724794854228298 \\
7  & GIS  & 0.38 & -0.0047575037453797 & -0.0095196036953139    & Decreased & -0.003617449404219282 \\
9  & CPB  & 0.30 & -0.0089281848761906 & -0.0136902848261248    & Decreased & -0.00410708544783744 \\
10 & HLF  & 0.27 &  0.0160907985807568 &  0.0113286986308227    & Increased &  0.003058748630322129 \\
\hline
\end{tabular}
}
\end{table}

\FloatBarrier

\subsection{Technology}

\subsubsection{AMD / Xilinx}
AMD/Xilinx is contradicting. Most semiconductor peers show negative ARs and the top three peers are inconsistent. The positive trendline indicates that lower-similarity firms slightly outperformed higher-similarity ones, inverting the shadow trading prediction. Semiconductor M\&A events may reflect competitive consolidation rather than acquisition premium contagion.

\begin{figure}[h!]
\centering
\includegraphics[width=0.75\textwidth]{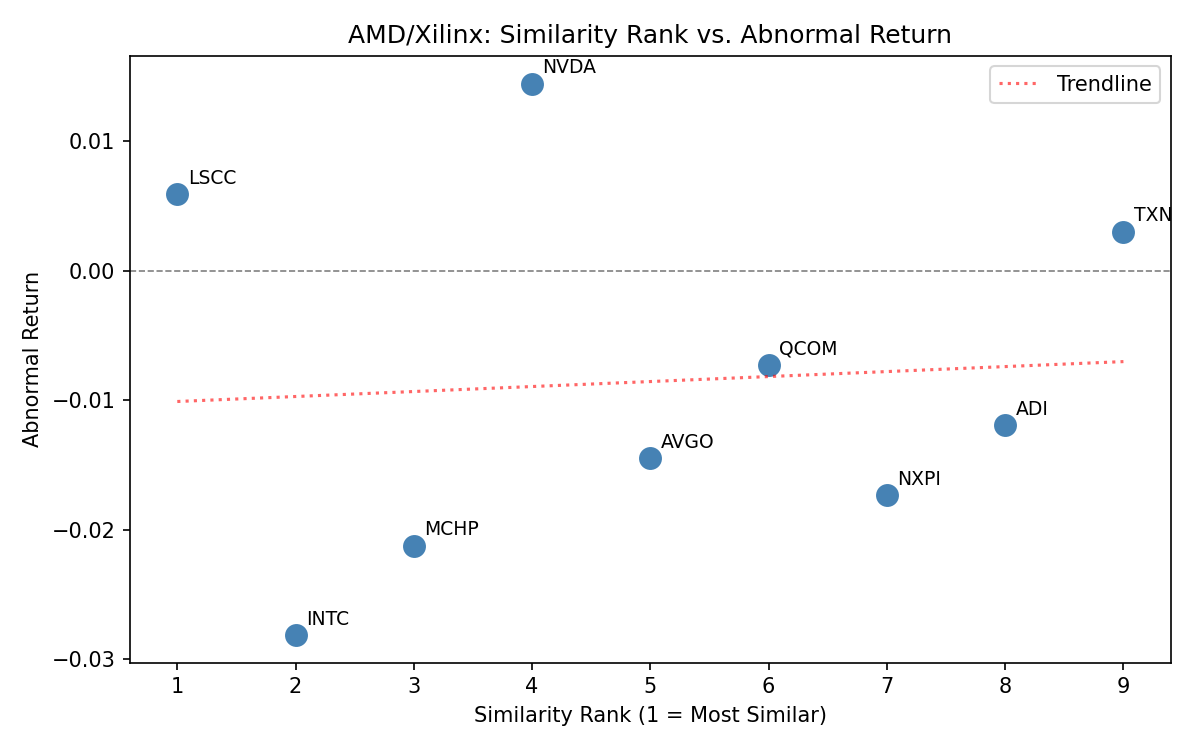}
\caption{AMD / Xilinx Similarity vs. Abnormal Returns}
\label{fig:amd-xilinx}
\end{figure}

\begin{table}[h!]
\centering
\caption{Abnormal Returns Xilinx}
\resizebox{0.65\textwidth}{!}{%
\begin{tabular}{lrrl}
\hline
Ticker & Raw Return & Abnormal Return & Direction \\
\hline
ADI  & -0.006934434890752879  & -0.011939806696974281  & Decreased \\
AVGO & -0.009475118768035064  & -0.014480490574256464  & Decreased \\
INTC & -0.02311685008217791   & -0.028122221888399312  & Decreased \\
LSCC &  0.01090167240875663   &  0.005896300602535229  & Increased \\
MCHP & -0.016221798231416912  & -0.021227170037638313  & Decreased \\
NVDA &  0.019442558707294864  &  0.014437186901073464  & Increased \\
NXPI & -0.012306568731181536  & -0.017311940537402936  & Decreased \\
QCOM & -0.0022979824548389656 & -0.007303354261060367  & Decreased \\
TXN  &  0.008016157780864525  &  0.0030107859746431242 & Increased \\
\hline
\end{tabular}
}
\end{table}

\begin{table}[h!]
\centering
\caption{Final Analysis Xilinx}
\resizebox{0.95\textwidth}{!}{
\begin{tabular}{r l r r r l r}
\hline
Rank & Ticker & Sim. Score & Raw Return & Abnormal Return & Direction & Shadow Signal \\
\hline
1 & LSCC & 0.86 &  0.0109016724087566 &  0.0058963006025352 & Increased &  0.005070818518180272 \\
2 & INTC & 0.75 & -0.0231168500821779 & -0.0281222218883993 & Decreased & -0.021091666416299476 \\
3 & MCHP & 0.72 & -0.0162217982314169 & -0.0212271700376383 & Decreased & -0.015283562427099575 \\
4 & NVDA & 0.65 &  0.0194425587072948 &  0.0144371869010734 & Increased &  0.00938417148569771 \\
5 & AVGO & 0.60 & -0.0094751187680350 & -0.0144804905742564 & Decreased & -0.00868829434455384 \\
6 & QCOM & 0.55 & -0.0022979824548389 & -0.0073033542610603 & Decreased & -0.0040168448435831654 \\
7 & NXPI & 0.50 & -0.0123065687311815 & -0.0173119405374029 & Decreased & -0.00865597026870145 \\
8 & ADI  & 0.48 & -0.0069344348907528 & -0.0119398066969742 & Decreased & -0.005731107214547616 \\
9 & TXN  & 0.35 &  0.0080161577808645 &  0.0030107859746431 & Increased &  0.001053775091125085 \\
\hline
\end{tabular}
}
\end{table}

\FloatBarrier

\subsubsection{Avago / Broadcom}
Avago/Broadcom is contradicting. The trendline is positive, and the top two peers (QCOM, SWKS) both show negative ARs, while the fourth and fifth peers show extremely high ARs relative to the cohort. This case, along with AMD/Xilinx, suggests that in semiconductor acquisitions, the announcement-day signal does not propagate to textually similar peers in a manner consistent with the shadow trading hypothesis.

\begin{figure}[h!]
\centering
\includegraphics[width=0.75\textwidth]{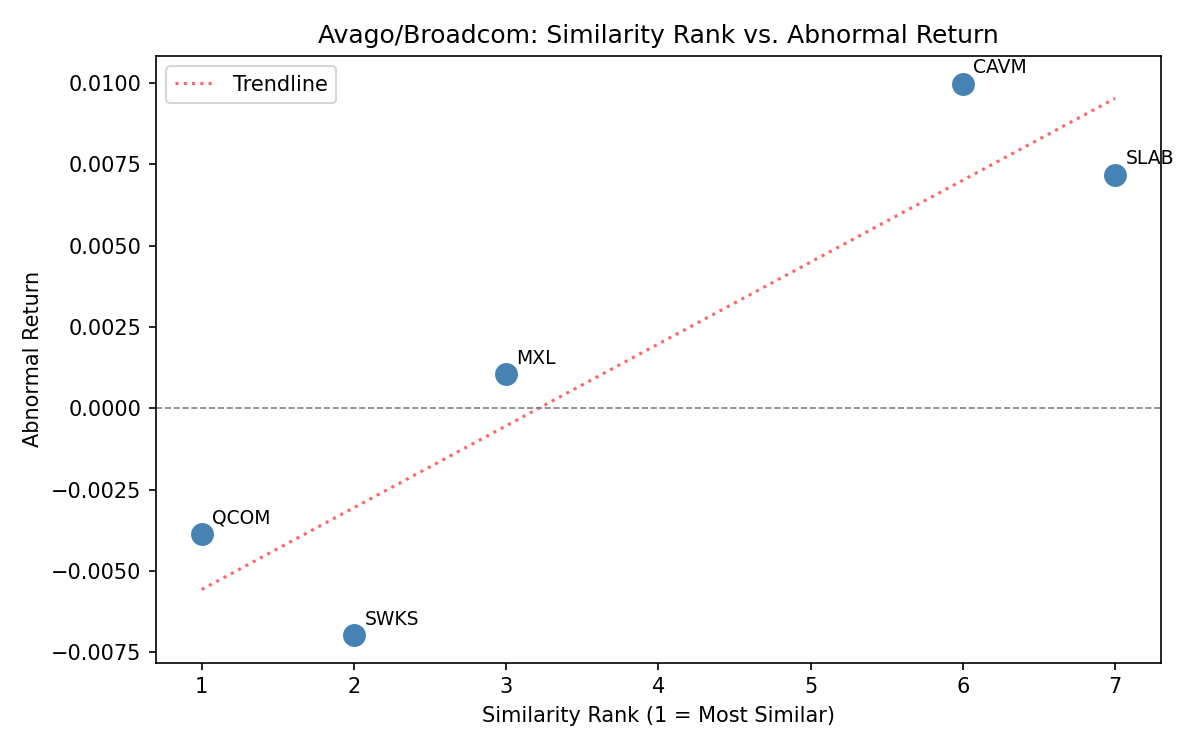}
\caption{Avago / Broadcom Similarity vs. Abnormal Returns}
\label{fig:avago-broadcom}
\end{figure}

\begin{table}[h!]
\centering
\caption{Abnormal Returns Broadcom}
\resizebox{0.65\textwidth}{!}{%
\begin{tabular}{lrrl}
\hline
Ticker & Raw Return & Abnormal Return & Direction \\
\hline
CAVM &  0.007932084987134312  &  0.009987993662528285  & Increased \\
MXL  & -0.0010019314294756854 &  0.001053977245918287  & Increased \\
QCOM & -0.0059139995490219    & -0.003858090873627927  & Decreased \\
SLAB &  0.005126374944520318  &  0.007182283619914291  & Increased \\
SWKS & -0.009021781906765464  & -0.006965873231371492  & Decreased \\
\hline
\end{tabular}
}
\end{table}

\begin{table}[h!]
\centering
\caption{Final Analysis Broadcom}
\resizebox{0.95\textwidth}{!}{
\begin{tabular}{r l r r r l r}
\hline
Rank & Ticker & Sim. Score & Raw Return & Abnormal Return & Direction & Shadow Signal \\
\hline
1 & QCOM & 0.91 & -0.0059139995490219 & -0.0038580908736279 & Decreased & -0.003510862695001389 \\
2 & SWKS & 0.83 & -0.0090217819067654 & -0.0069658732313714 & Decreased & -0.005781674782038262 \\
3 & MXL  & 0.74 & -0.0010019314294756 &  0.0010539772459182 & Increased &  0.0007799431619794681 \\
6 & CAVM & 0.60 &  0.0079320849871343 &  0.0099879936625282 & Increased &  0.005992796197516919 \\
7 & SLAB & 0.58 &  0.0051263749445203 &  0.0071822836199142 & Increased &  0.004165724499550235 \\
\hline
\end{tabular}
}
\end{table}

\FloatBarrier

\subsubsection{Cisco / Splunk}
Cisco/Splunk is classified as mixed. The top-3 AR direction is ($-,+,+$). The trendline is negative, and the distribution is bimodal: some peers show meaningful positive reactions (ESTC, DT) while others decline sharply (DDOG, SNOW). The mixed classification reflects genuine ambiguity rather than a clean contradicting signal.

\begin{figure}[h!]
\centering
\includegraphics[width=0.75\textwidth]{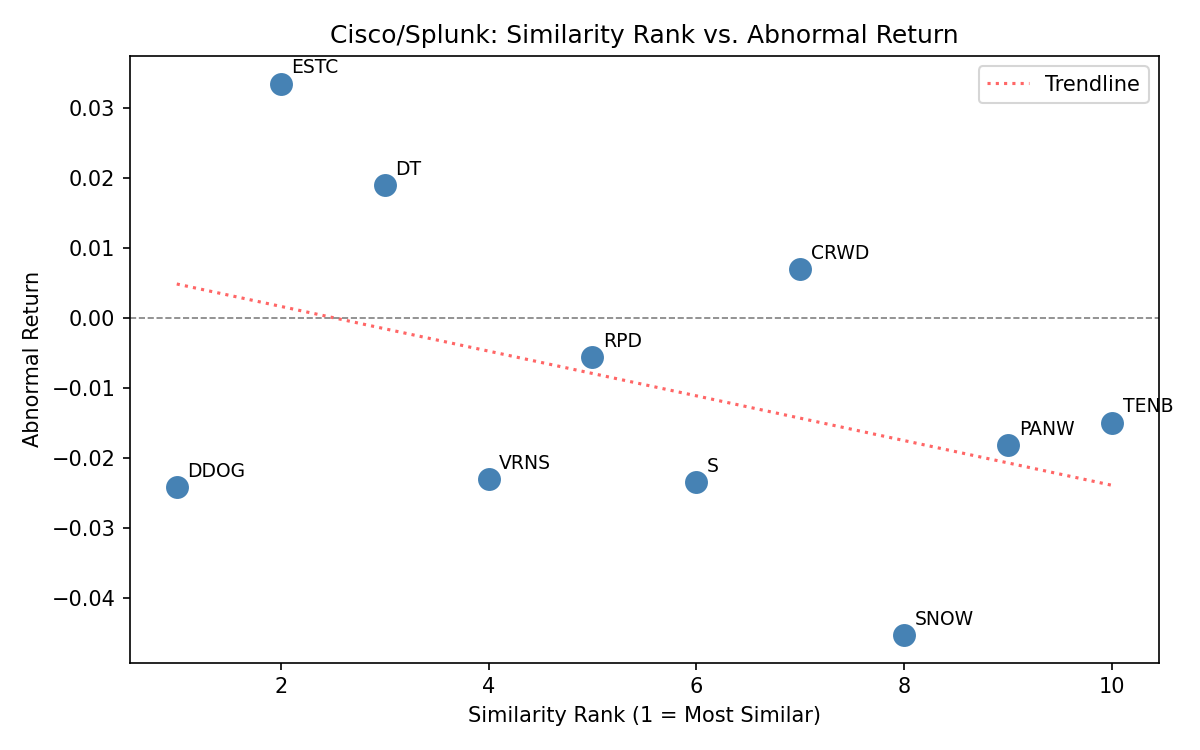}
\caption{Cisco / Splunk Similarity vs. Abnormal Returns}
\label{fig:cisco-splunk}
\end{figure}

\begin{table}[h!]
\centering
\caption{Abnormal Returns Splunk}
\resizebox{0.65\textwidth}{!}{%
\begin{tabular}{lrrl}
\hline
Ticker & Raw Return & Abnormal Return & Direction \\
\hline
CRWD & -0.008298252580489437 &  0.006947863262951159 & Increased \\
DDOG & -0.03943514136301994  & -0.024189025519579344 & Decreased \\
DT   &  0.003675636085304054 &  0.01892175192874465  & Increased \\
ESTC &  0.01819609112127831  &  0.033442206964718904 & Increased \\
PANW & -0.03340416705656666  & -0.018158051213126063 & Decreased \\
RPD  & -0.02088660657461326  & -0.005640490731172662 & Decreased \\
SNOW & -0.060518223589588895 & -0.0452721077461483   & Decreased \\
S    & -0.038670659767480936 & -0.02342454392404034  & Decreased \\
TENB & -0.030199183741365668 & -0.014953067897925072 & Decreased \\
VRNS & -0.038279056262499216 & -0.02303294041905862  & Decreased \\
\hline
\end{tabular}
}
\end{table}

\begin{table}[h!]
\centering
\caption{Final Analysis Splunk}
\resizebox{0.95\textwidth}{!}{
\begin{tabular}{r l r r r l r}
\hline
Rank & Ticker & Sim. Score & Raw Return & Abnormal Return & Direction & Shadow Signal \\
\hline
1  & DDOG & 0.92 & -0.0394351413630199 & -0.0241890255195793 & Decreased & -0.022253903478012958 \\
2  & ESTC & 0.89 &  0.0181960911212783 &  0.0334422069647189 & Increased &  0.02976356419859982 \\
3  & DT   & 0.87 &  0.0036756360853040 &  0.0189217519287446 & Increased &  0.016461924178007802 \\
4  & VRNS & 0.76 & -0.0382790562624992 & -0.0230329404190586 & Decreased & -0.017505034718484535 \\
5  & RPD  & 0.74 & -0.0208866065746132 & -0.0056404907311726 & Decreased & -0.004173963141067724 \\
6  & S    & 0.73 & -0.0386706597674809 & -0.0234245439240403 & Decreased & -0.01709991706454942 \\
7  & CRWD & 0.71 & -0.0082982525804894 &  0.0069478632629511 & Increased &  0.004932982916695281 \\
8  & SNOW & 0.70 & -0.0605182235895888 & -0.0452721077461483 & Decreased & -0.03169047542230381 \\
9  & PANW & 0.63 & -0.0334041670565666 & -0.0181580512131260 & Decreased & -0.011439572264269381 \\
10 & TENB & 0.56 & -0.0301991837413656 & -0.0149530678979250 & Decreased & -0.008373718022838001 \\
\hline
\end{tabular}
}
\end{table}

\FloatBarrier

\subsubsection{Microchip / Microsemi}
Microchip/Microsemi is also classified as mixed. All identified peers show small negative or positive ARs close to zero, and the trendline is positive but relatively flat when considering the scale. This case is notable for its uniformly low magnitude reactions, suggesting the acquisition was broadly anticipated by the market and generated no meaningful announcement-day peer spillover.

\begin{figure}[h!]
\centering
\includegraphics[width=0.75\textwidth]{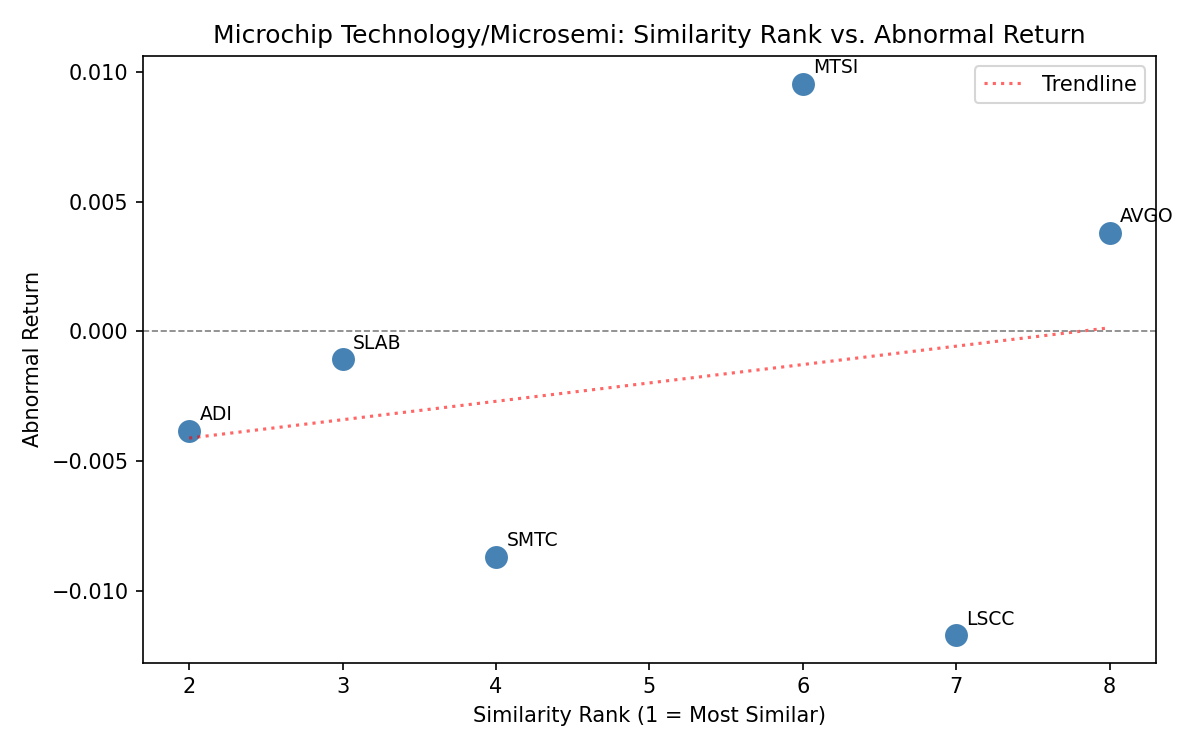}
\caption{Microchip / Microsemi Similarity vs. Abnormal Returns}
\label{fig:microchip-microsemi}
\end{figure}

\begin{table}[h!]
\centering
\caption{Abnormal Returns Microsemi}
\resizebox{0.65\textwidth}{!}{%
\begin{tabular}{lrrl}
\hline
Ticker & Raw Return & Abnormal Return & Direction \\
\hline
ADI  & -0.020410672883296514 & -0.003834261127835862  & Decreased \\
AVGO & -0.012781175422850264 &  0.0037952363326103876 & Increased \\
LSCC & -0.028286201301122735 & -0.011709789545662083  & Decreased \\
MTSI & -0.007029035494738882 &  0.00954737626072177   & Increased \\
SLAB & -0.017647075143089906 & -0.0010706633876292546 & Decreased \\
SMTC & -0.025260096590624392 & -0.00868368483516374   & Decreased \\
\hline
\end{tabular}
}
\end{table}

\begin{table}[h!]
\centering
\caption{Final Analysis Microsemi}
\resizebox{0.95\textwidth}{!}{
\begin{tabular}{r l r r r l r}
\hline
Rank & Ticker & Sim. Score & Raw Return & Abnormal Return & Direction & Shadow Signal \\
\hline
2 & ADI  & 0.87 & -0.0204106728832965 & -0.0038342611278358 & Decreased & -0.003335807181217146 \\
3 & SLAB & 0.84 & -0.0176470751430899 & -0.0010706633876292 & Decreased & -0.0008993572456085279 \\
4 & SMTC & 0.81 & -0.0252600965906243 & -0.0086836848351637 & Decreased & -0.007033784716482597 \\
6 & MTSI & 0.75 & -0.0070290354947388 &  0.0095473762607217 & Increased &  0.007160532195541275 \\
7 & LSCC & 0.72 & -0.0282862013011227 & -0.0117097895456620 & Decreased & -0.008431048472876639 \\
8 & AVGO & 0.68 & -0.0127811754228502 &  0.0037952363326103 & Increased &  0.002580760706175004 \\
\hline
\end{tabular}
}
\end{table}

\FloatBarrier

\subsubsection{SS\&C / DST Systems}
SS\&C/DST Systems is contradicting. Despite most peers showing positive ARs, the trendline is inverted: lower-similarity peers (MMS, CNDT) show the largest positive reactions, while higher-similarity peers (SEIC, FIS) show negligible returns. This is a case where broad sector sentiment lifted the cohort but failed to concentrate in the most textually similar firms.

\begin{figure}[h!]
\centering
\includegraphics[width=0.75\textwidth]{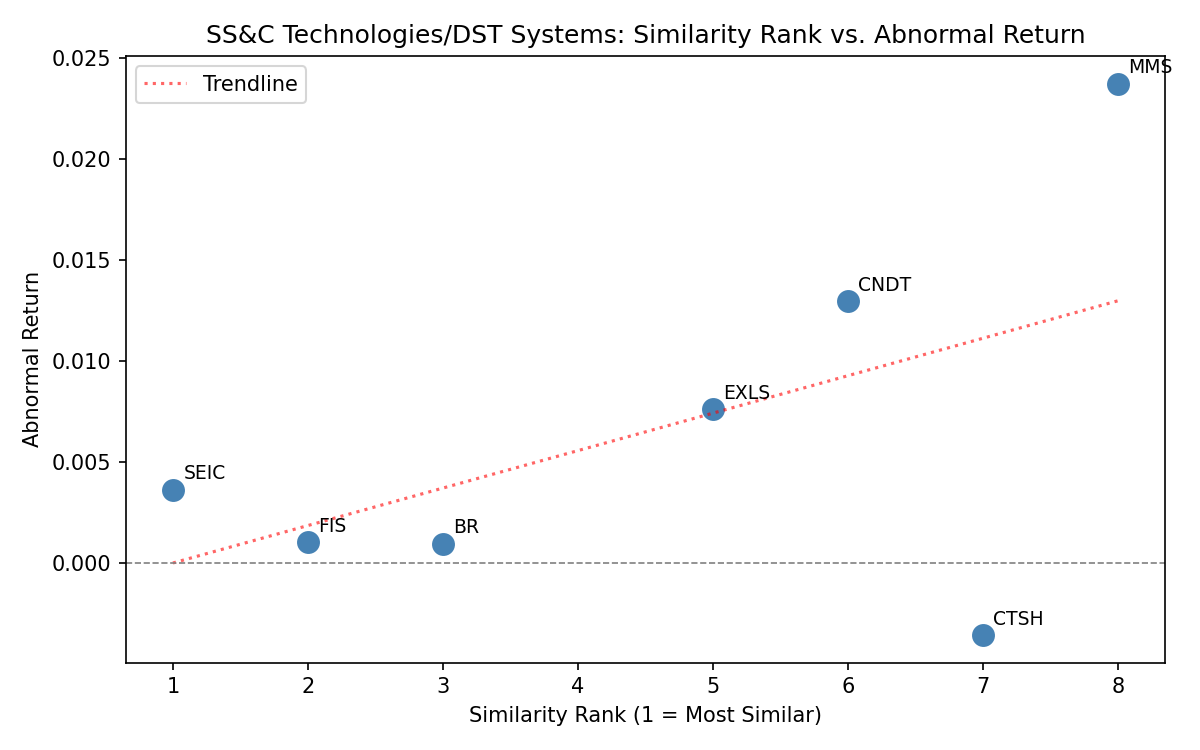}
\caption{SS\&C / DST Systems Similarity vs. Abnormal Returns}
\label{fig:ssandc-dst-systems}
\end{figure}

\begin{table}[h!]
\centering
\caption{Abnormal Returns DST Systems}
\resizebox{0.65\textwidth}{!}{%
\begin{tabular}{lrrl}
\hline
Ticker & Raw Return & Abnormal Return & Direction \\
\hline
BR   & 0.005054569298770341  & 0.0009727415307561648 & Increased \\
CNDT & 0.017041921734393988  & 0.012960093966379812  & Increased \\
CTSH & 0.000540932537022394  & -0.003540895230991782 & Decreased \\
DXC  & -0.0013830794533293202 & -0.005464907221343496 & Decreased \\
EXLS & 0.011720661765818713  & 0.007638833997804536  & Increased \\
FISV & 0.00598501829863103   & 0.0019031905306168542 & Increased \\
FIS  & 0.0051402877822758035 & 0.0010584600142616273 & Increased \\
MMS  & 0.02782137306097962   & 0.023739545292965444  & Increased \\
SEIC & 0.007733366430757337  & 0.003651538662743161  & Increased \\
\hline
\end{tabular}
}
\end{table}

\begin{table}[h!]
\centering
\caption{Final Analysis DST Systems}
\resizebox{0.95\textwidth}{!}{
\begin{tabular}{r l r r r l r}
\hline
Rank & Ticker & Sim. Score & Raw Return & Abnormal Return & Direction & Shadow Signal \\
\hline
1 & SEIC & 0.87 & 0.0077333664307573 & 0.0036515386627431 & Increased & 0.003176838636586497 \\
2 & FIS  & 0.79 & 0.0051402877822758 & 0.0010584600142616 & Increased & 0.000836183411266664 \\
3 & BR   & 0.77 & 0.0050545692987703 & 0.0009727415307561 & Increased & 0.0007490109786821971 \\
5 & EXLS & 0.64 & 0.0117206617658187 & 0.0076388339978045 & Increased & 0.00488885375859488 \\
6 & CNDT & 0.54 & 0.0170419217343939 & 0.0129600939663798 & Increased & 0.006998450741845092 \\
7 & CTSH & 0.45 & 0.0005409325370223 & -0.0035408952309917 & Decreased & -0.0015934028539462649 \\
8 & MMS  & 0.36 & 0.0278213730609796 & 0.0237395452929654 & Increased & 0.008546236305467544 \\
\hline
\end{tabular}
}
\end{table}

\FloatBarrier

\subsection{Automobiles}

\subsubsection{Apollo / Tenneco}
Apollo/Tenneco is classified as mixed. The trendline is positive and the top-3 direction is ($-,+,-$), reflecting the split between DAN (top-ranked, strongly negative AR) and BWA (second-ranked, strongly positive AR). The two most similar firms diverge sharply in their announcement-day reactions, making the case uninterpretable under either hypothesis.

\begin{figure}[h!]
\centering
\includegraphics[width=0.75\textwidth]{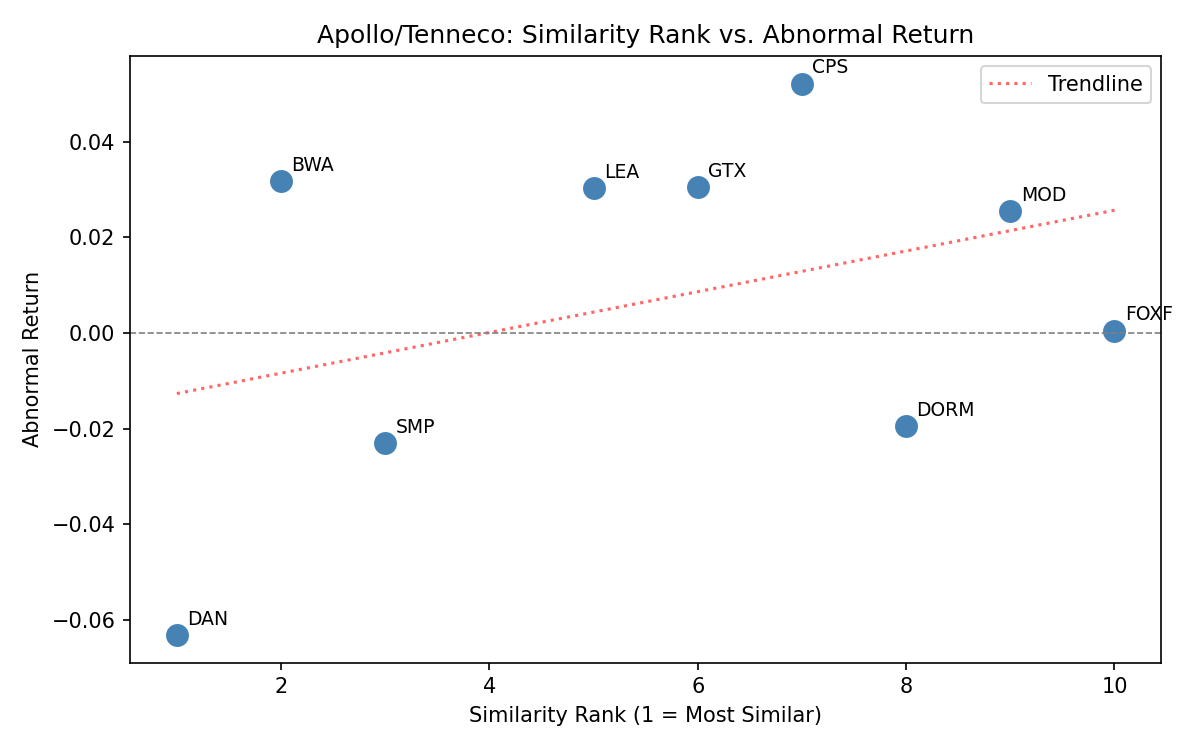}
\caption{Apollo / Tenneco Similarity vs. Abnormal Returns}
\label{fig:apollo-tenneco}
\end{figure}

\begin{table}[h!]
\centering
\caption{Abnormal Returns Tenneco}
\resizebox{0.65\textwidth}{!}{%
\begin{tabular}{lrrl}
\hline
Ticker & Raw Return & Abnormal Return & Direction \\
\hline
SMP  & -0.05646922230037127 & -0.023050145168253267 & Decreased \\
GTX  & -0.00291541343558791 &  0.03050366369653009  & Increased \\
DCH  &  0.06953226312827365 &  0.10295134026039165  & Increased \\
BWA  & -0.001651781076413566 & 0.031767296055704435 & Increased \\
MOD  & -0.00795237316535015 &  0.025466703966767854 & Increased \\
DORM & -0.052984267629254074 & -0.01956519049713607 & Decreased \\
FOXF & -0.032919795754358626 & 0.0004992813777593766 & Increased \\
CPS  &  0.018749981839210065 & 0.052169058971328064 & Increased \\
DAN  & -0.0966093471696384  & -0.0631902700375204  & Decreased \\
LEA  & -0.0031639338837505583 & 0.030255143248367443 & Increased \\
\hline
\end{tabular}
}
\end{table}

\begin{table}[h!]
\centering
\caption{Final Analysis Tenneco}
\resizebox{0.95\textwidth}{!}{
\begin{tabular}{r l r r r l r}
\hline
Rank & Ticker & Sim. Score & Raw Return & Abnormal Return & Direction & Shadow Signal \\
\hline
1  & DAN  & 0.90 & -0.0966093471696384 & -0.0631902700375204 & Decreased & -0.05687124303376836 \\
2  & BWA  & 0.90 & -0.0016517810764135 &  0.0317672960557044 & Increased &  0.02859056645013396 \\
3  & SMP  & 0.84 & -0.0564692223003712 & -0.0230501451682532 & Decreased & -0.01936212194133269 \\
5  & LEA  & 0.74 & -0.0031639338837505 &  0.0302551432483674 & Increased &  0.022388806003791877 \\
6  & GTX  & 0.70 & -0.0029154134355879 &  0.03050366369653   & Increased &  0.021352564587571 \\
7  & CPS  & 0.69 &  0.01874998183921   &  0.052169058971328  & Increased &  0.03599665069021632 \\
8  & DORM & 0.62 & -0.052984267629254  & -0.019565190497136  & Decreased & -0.012130418108224319 \\
9  & MOD  & 0.55 & -0.0079523731653501 &  0.0254667039667678 & Increased &  0.014006687181722292 \\
10 & FOXF & 0.50 & -0.0329197957543586 &  0.0004992813777593 & Increased &  0.00024964068887965 \\
\hline
\end{tabular}
}
\end{table}

\FloatBarrier

\subsubsection{Goodyear / Cooper}
Goodyear/Cooper is a strongly supporting case. Eight of nine peers show positive abnormal returns, and the negative trendline holds clearly across the cohort. This is the most uniformly positive result in the Automotive subsample and suggests that in tire and rubber supply chains, textual similarity from 10-K disclosures reliably identifies economically linked firms.

\begin{figure}[h!]
\centering
\includegraphics[width=0.75\textwidth]{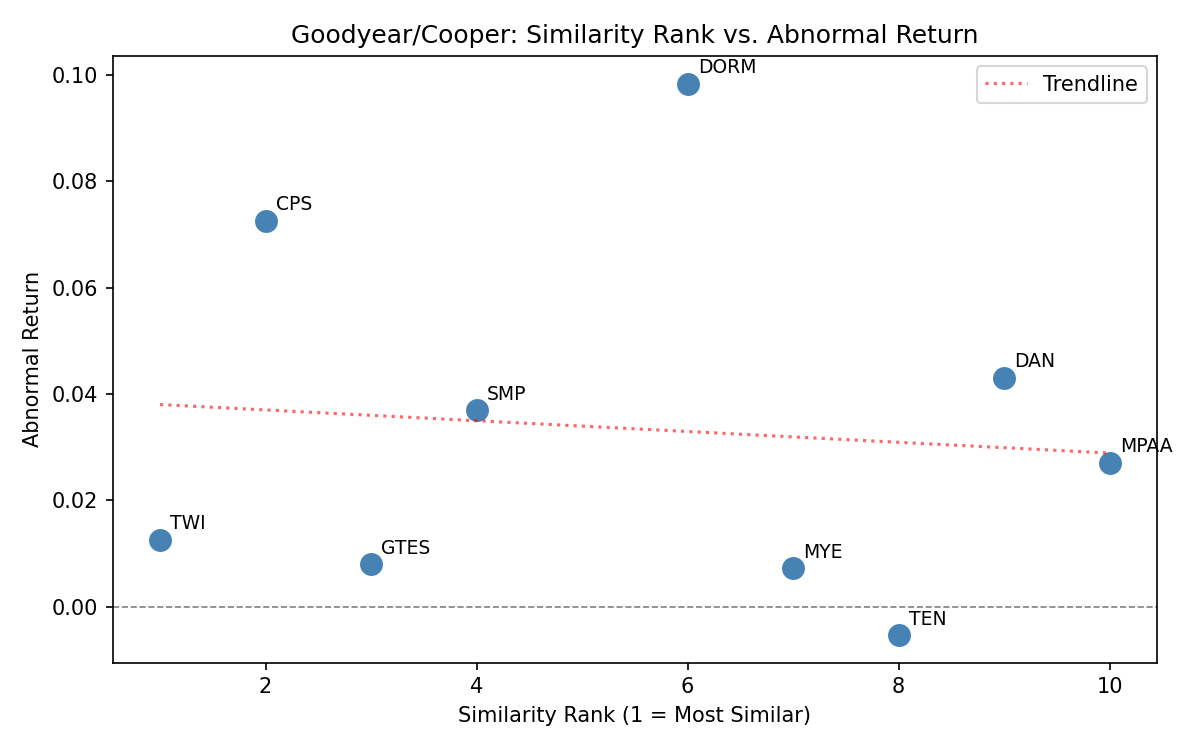}
\caption{Goodyear / Cooper Similarity vs. Abnormal Returns}
\label{fig:goodyear-cooper}
\end{figure}

\begin{table}[h!]
\centering
\caption{Abnormal Returns Cooper}
\resizebox{0.65\textwidth}{!}{%
\begin{tabular}{lrrl}
\hline
Ticker & Raw Return & Abnormal Return & Direction \\
\hline
DORM &  0.10152553461907389 & 0.09833821671472792  & Increased \\
TWI  &  0.01581509856664611 & 0.012627780662300143 & Increased \\
DAN  &  0.04622995957741803 & 0.04304264167307206  & Increased \\
CPS  &  0.07574440359430722 & 0.07255708568996125  & Increased \\
GTES &  0.01118308043656393 & 0.007995762532217964 & Increased \\
MYE  &  0.010512060818193676 & 0.007324742913847709 & Increased \\
TEN  & -0.0020920732042427684 & -0.005279391108588736 & Decreased \\
MPAA &  0.030302988601413512 & 0.027115670697067544 & Increased \\
SMP  &  0.04016662089667492 & 0.036979302992328954 & Increased \\
\hline
\end{tabular}
}
\end{table}

\begin{table}[h!]
\centering
\caption{Final Analysis Cooper}
\resizebox{0.95\textwidth}{!}{
\begin{tabular}{r l r r r l r}
\hline
Rank & Ticker & Sim. Score & Raw Return & Abnormal Return & Direction & Shadow Signal \\
\hline
1  & TWI  & 0.86 &  0.0158150985666461 & 0.0126277806623001 & Increased &  0.010859891369578085 \\
2  & CPS  & 0.77 &  0.0757444035943072 & 0.0725570856899612 & Increased &  0.05586895598127013 \\
3  & GTES & 0.75 &  0.0111830804365639 & 0.0079957625322179 & Increased &  0.005996821899163425 \\
4  & SMP  & 0.60 &  0.0401666208966749 & 0.0369793029923289 & Increased &  0.02218758179539734 \\
6  & DORM & 0.56 &  0.1015255346190738 & 0.0983382167147279 & Increased &  0.05506940136024763 \\
7  & MYE  & 0.52 &  0.0105120608181936 & 0.0073247429138477 & Increased &  0.0038088663152008044 \\
8  & TEN  & 0.50 & -0.0020920732042427 & -0.0052793911085887 & Decreased & -0.00263969555429435 \\
9  & DAN  & 0.48 &  0.046229959577418  & 0.043042641673072  & Increased &  0.02066046800307456 \\
10 & MPAA & 0.40 &  0.0303029886014135 & 0.0271156706970675 & Increased &  0.010846268278827 \\
\hline
\end{tabular}
}
\end{table}

\FloatBarrier

\subsubsection{Tenneco / Federal-Mogul}
Tenneco/Federal-Mogul is classified as supporting despite a mixed top-3 direction ($-,+,-$). The classification reflects the overall pattern: the two firms with positive reactions (BWA, rank~4; SMP, rank~8) are outweighed by larger negative reactions concentrated in the lowest-ranked peers (MOD, LKQ), producing the negative trendline. The holistic analysis supports the hypothesis even though the top-ranked firm (DAN) shows a small negative AR.

\begin{figure}[h!]
\centering
\includegraphics[width=0.75\textwidth]{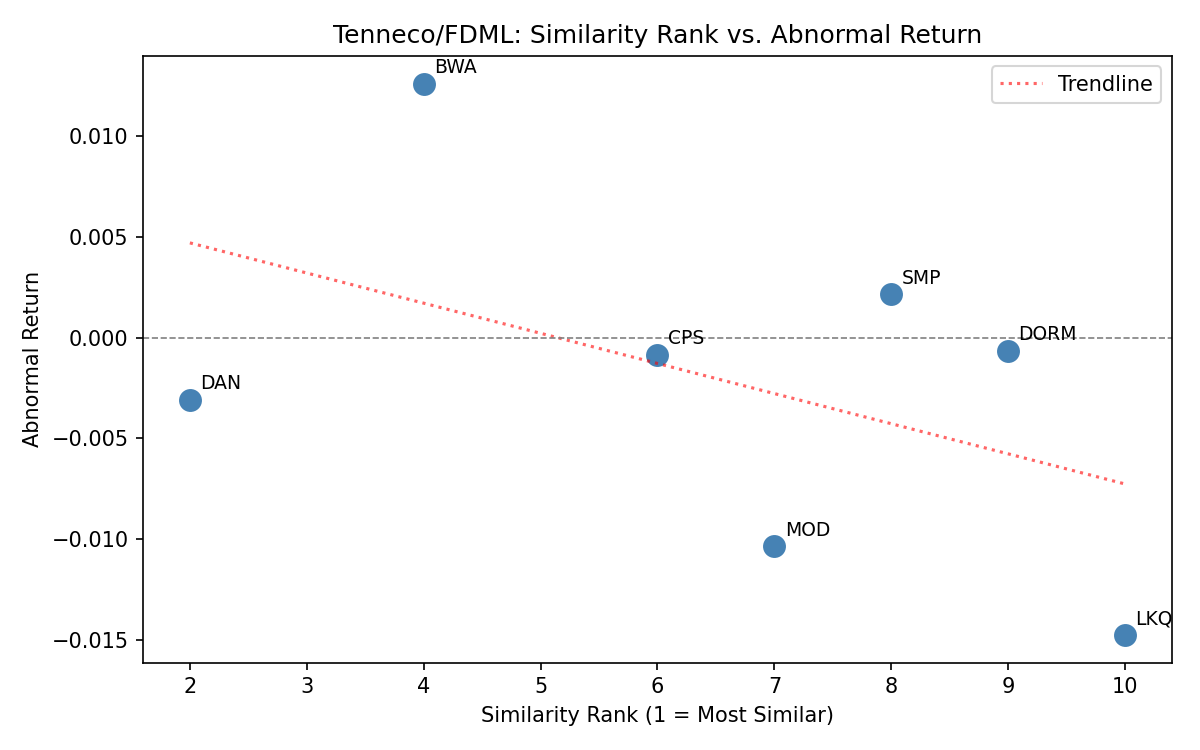}
\caption{Tenneco / Federal-Mogul Similarity vs. Abnormal Returns}
\label{fig:tenneco-federal-mogul}
\end{figure}

\begin{table}[h!]
\centering
\caption{Abnormal Returns Federal}
\resizebox{0.65\textwidth}{!}{%
\begin{tabular}{lrrl}
\hline
Ticker & Raw Return & Abnormal Return & Direction \\
\hline
BWA  & 0.027923368972255665  &  0.012623729497841695  & Increased \\
CMI  & 0.03463282681896851   &  0.01933318734455454   & Increased \\
DORM & 0.01464721921591628   & -0.000652420258497691  & Decreased \\
LKQ  & 0.0005245657554827761 & -0.014775073718931194  & Decreased \\
SMP  & 0.017457329721931526  &  0.0021576902475175554 & Increased \\
DAN  & 0.012209253473614552  & -0.0030903860007994183 & Decreased \\
DCH  & 0.0006250143051147461 & -0.014674625169299224  & Decreased \\
CPS  & 0.014450176569453304  & -0.0008494629049606659 & Decreased \\
MOD  & 0.004926127351136306  & -0.010373512123277664  & Decreased \\
\hline
\end{tabular}
}
\end{table}

\begin{table}[h!]
\centering
\caption{Final Analysis Federal}
\resizebox{0.95\textwidth}{!}{
\begin{tabular}{r l r r r l r}
\hline
Rank & Ticker & Sim. Score & Raw Return & Abnormal Return & Direction & Shadow Signal \\
\hline
2  & DAN  & 0.91 &  0.0122092534736145 & -0.0030903860007994 & Decreased & -0.0028122512607274544 \\
4  & BWA  & 0.85 &  0.0279233689722556 &  0.0126237294978416 & Increased &  0.01073017007316536 \\
6  & CPS  & 0.82 &  0.0144501765694533 & -0.0008494629049606 & Decreased & -0.0006965595820676919 \\
7  & MOD  & 0.74 &  0.0049261273511363 & -0.0103735121232776 & Decreased & -0.007676398971225424 \\
8  & SMP  & 0.60 &  0.0174573297219315 &  0.0021576902475175 & Increased &  0.0012946141485104998 \\
9  & DORM & 0.51 &  0.0146472192159162 & -0.0006524202584976 & Decreased & -0.000332734331833776 \\
10 & LKQ  & 0.37 &  0.0005245657554827 & -0.0147750737189311 & Decreased & -0.005466777276004507 \\
\hline
\end{tabular}
}
\end{table}

\FloatBarrier

\subsubsection{Warner / Delphi}
Warner/Delphi is classified as mixed. The trendline is positive and the top-3 direction is ($-,\approx0,+$), with the top-ranked firm (TEN) showing a negative AR and the third-ranked (DAN) showing a positive one. The lack of a consistent pattern across either direction prevents a clear supporting or contradicting classification.

\begin{figure}[h!]
\centering
\includegraphics[width=0.75\textwidth]{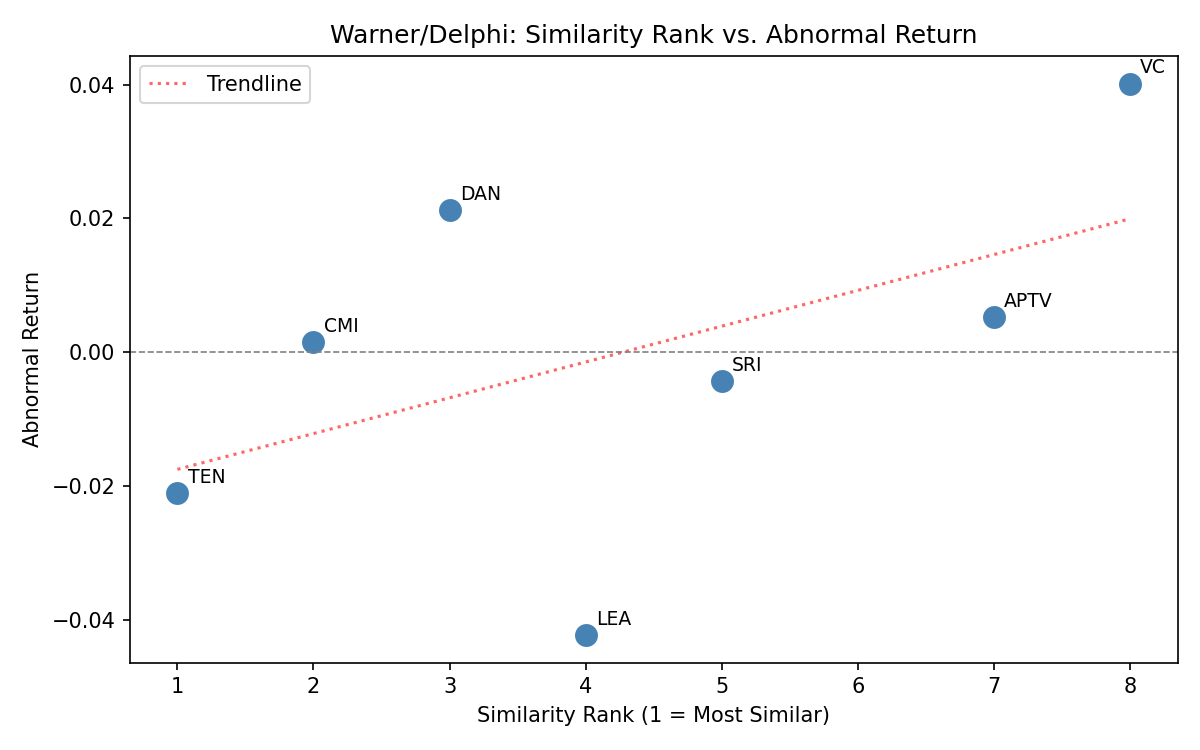}
\caption{Warner / Delphi Similarity vs. Abnormal Returns}
\label{fig:warner-delphi}
\end{figure}

\begin{table}[h!]
\centering
\caption{Abnormal Returns Delphi}
\resizebox{0.65\textwidth}{!}{%
\begin{tabular}{lrrl}
\hline
Ticker & Raw Return & Abnormal Return & Direction \\
\hline
VC   &  0.04662139122157009  &  0.04015559590695077  & Increased \\
DCH  &  0.018104373810255786 &  0.011638578495636463 & Increased \\
SMP  & -0.0027740168695917916 & -0.009239812184211115 & Decreased \\
APTV &  0.011738143800527612 &  0.00527234848590829  & Increased \\
TEN  & -0.014577270889797867 & -0.02104306620441719  & Decreased \\
ST   &  0.010699377862144956 &  0.004233582547525633 & Increased \\
SRI  &  0.0021413085282714493 & -0.004324486786347873 & Decreased \\
DAN  &  0.027690593095965572 &  0.021224797781346248 & Increased \\
CMI  &  0.007950362187841791 &  0.0014845668732224688 & Increased \\
LEA  & -0.03579390798828836  & -0.04225970330290768  & Decreased \\
\hline
\end{tabular}
}
\end{table}

\begin{table}[h!]
\centering
\caption{Final Analysis Delphi}
\resizebox{0.95\textwidth}{!}{
\begin{tabular}{r l r r r l r}
\hline
Rank & Ticker & Sim. Score & Raw Return & Abnormal Return & Direction & Shadow Signal \\
\hline
1 & TEN  & 0.94 & -0.0145772708897978 & -0.0210430662044171 & Decreased & -0.019780482232152074 \\
2 & CMI  & 0.87 &  0.0079503621878417 &  0.0014845668732224 & Increased &  0.001291573179703488 \\
3 & DAN  & 0.85 &  0.0276905930959655 &  0.0212247977813462 & Increased &  0.01804107811414427 \\
4 & LEA  & 0.81 & -0.0357939079882883 & -0.0422597033029076 & Decreased & -0.034230359675355156 \\
5 & SRI  & 0.78 &  0.0021413085282714 & -0.0043244867863478 & Decreased & -0.003373099693351284 \\
7 & APTV & 0.73 &  0.0117381438005276 &  0.0052723484859082 & Increased &  0.0038488143947129857 \\
8 & VC   & 0.65 &  0.04662139122157   &  0.0401555959069507 & Increased &  0.026101137339517955 \\
\hline
\end{tabular}
}
\end{table}

\FloatBarrier

\subsubsection{Warner / Remy}
Warner/Remy is a strongly supporting case. Seven of eight peers show positive abnormal returns, and the negative trendline holds across the cohort. The top-ranked firm (MPAA, $+3.9\%$ AR) and second-ranked firm (SMP, $+1.0\%$ AR) both show positive reactions consistent with the shadow trading hypothesis.

\begin{figure}[h!]
\centering
\includegraphics[width=0.75\textwidth]{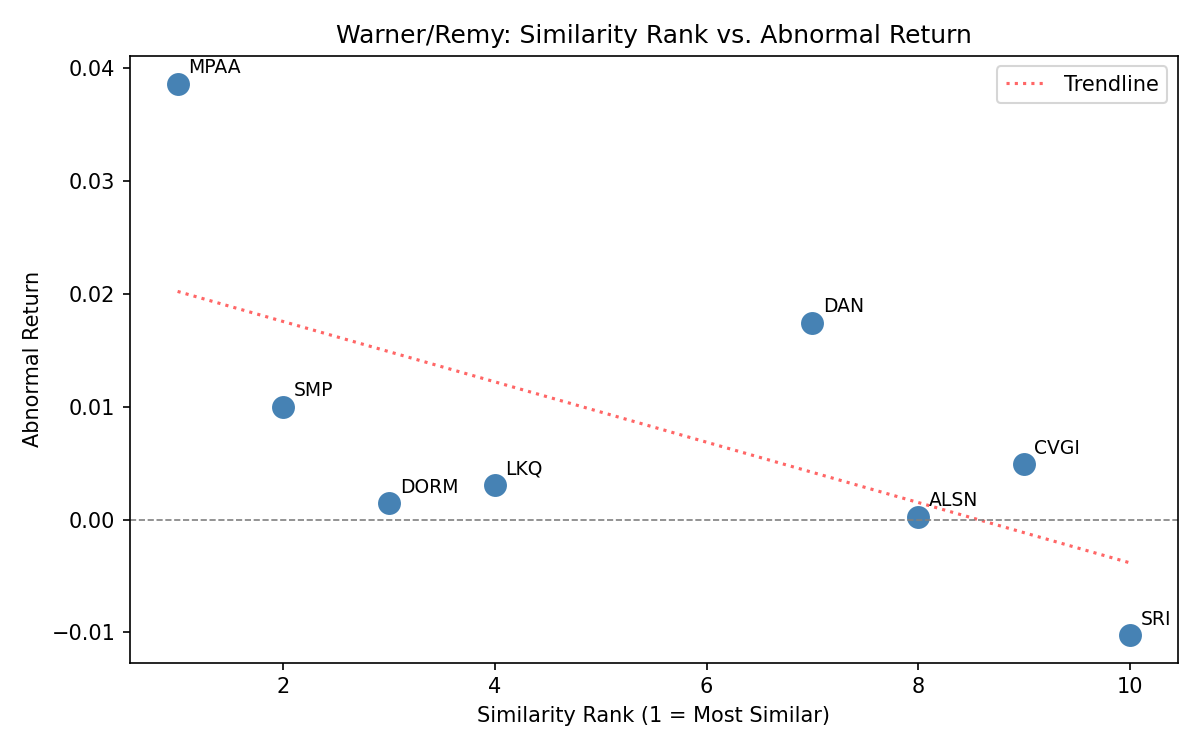}
\caption{Warner / Remy Similarity vs. Abnormal Returns}
\label{fig:warner-remy}
\end{figure}

\begin{table}[h!]
\centering
\caption{Abnormal Returns Remy}
\resizebox{0.65\textwidth}{!}{%
\begin{tabular}{lrrl}
\hline
Ticker & Raw Return & Abnormal Return & Direction \\
\hline
ALSN & 0.011316857070665666 &  0.00024048661547815438 & Increased \\
DAN  & 0.02852791101708753  &  0.017451540561900018   & Increased \\
LKQ  & 0.014177439899237226 &  0.003101069444049715   & Increased \\
MPAA & 0.04971055644378255  &  0.03863418598859504    & Increased \\
SRI  & 0.0008417700997898144 & -0.010234600355397697  & Decreased \\
DORM & 0.01250525832727829  &  0.0014288878720907783  & Increased \\
CVGI & 0.01597442507608828  &  0.004898054620900767   & Increased \\
SMP  & 0.021061745727400483 &  0.009985375272212972   & Increased \\
\hline
\end{tabular}
}
\end{table}

\begin{table}[h!]
\centering
\caption{Final Analysis Remy}
\resizebox{0.95\textwidth}{!}{
\begin{tabular}{r l r r r l r}
\hline
Rank & Ticker & Sim. Score & Raw Return & Abnormal Return & Direction & Shadow Signal \\
\hline
1  & MPAA & 0.94 &  0.0497105564437825 &  0.0386341859885950 & Increased &  0.0363161348292793 \\
2  & SMP  & 0.87 &  0.0210617457274004 &  0.0099853752722129 & Increased &  0.008687276486825221 \\
3  & DORM & 0.71 &  0.0125052583272782 &  0.0014288878720907 & Increased &  0.0010145103891843969 \\
4  & LKQ  & 0.70 &  0.0141774398992372 &  0.0031010694440497 & Increased &  0.00217074861083479 \\
7  & DAN  & 0.59 &  0.0285279110170875 &  0.0174515405619000 & Increased &  0.0102964089315210 \\
8  & ALSN & 0.56 &  0.0113168570706656 &  0.0002404866154781 & Increased &  0.00013467250466773603 \\
9  & CVGI & 0.48 &  0.0159744250760882 &  0.0048980546209007 & Increased &  0.0023510662180323363 \\
10 & SRI  & 0.48 &  0.0008417700997898 & -0.0102346003553976 & Decreased & -0.004912608170590848 \\
\hline
\end{tabular}
}
\end{table}

\FloatBarrier

\subsection{Finance}

\subsubsection{BB\&T / SunTrust}

BB\&T/SunTrust presents a nuanced case. All peer firms show positive abnormal returns on announcement day, which is consistent with the weak form of the shadow trading hypothesis---that MNPI about one bank would affect similar banks generally. However, the positive trendline indicates that the magnitude of those gains completely decoupled from semantic similarity rank: lower-similarity peers outperformed higher-similarity peers. This is the sector-wide contagion failure mode: the merger signaled regulatory approval of large-bank consolidation broadly, lifting all regional bank peers regardless of their textual similarity to SunTrust. Because our test evaluates the strong form of the hypothesis--- whether similarity rank predicts return magnitude, which is what the SEC needs to identify specific trading targets---this case is classified as contradicting, even though all peers moved in the predicted direction.

\begin{figure}[h!]
\centering
\includegraphics[width=0.75\textwidth]{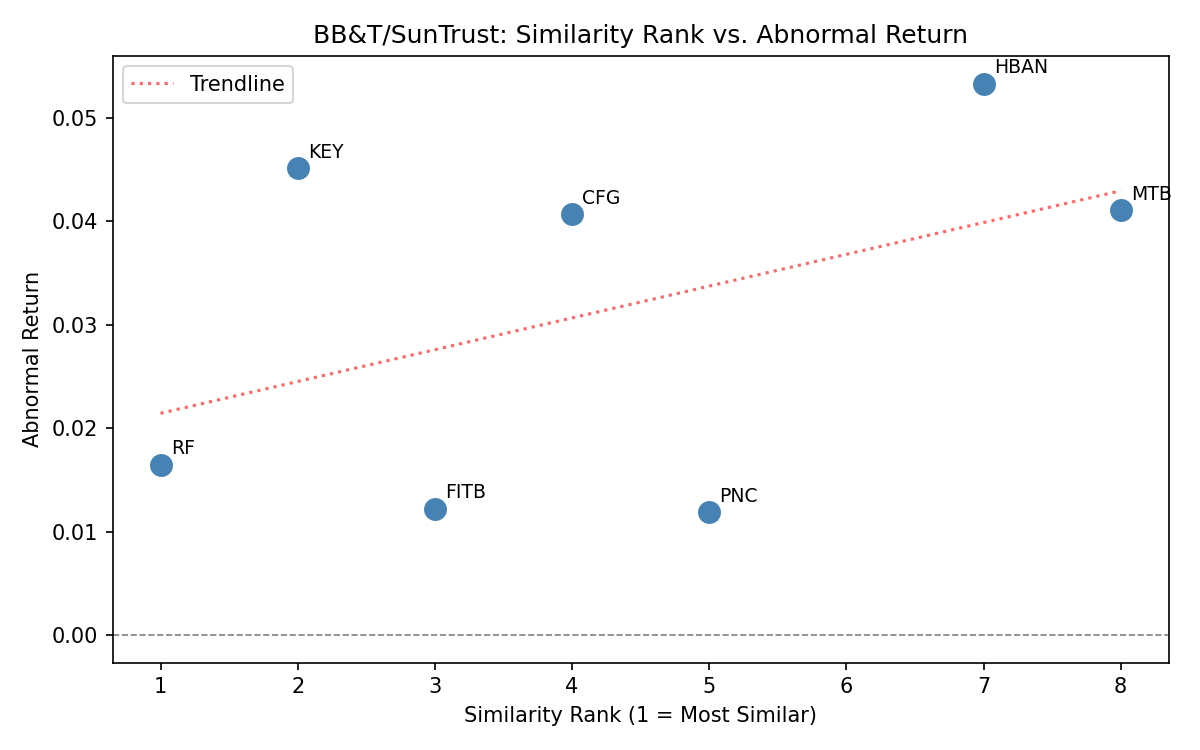}
\caption{BB\&T / SunTrust Similarity vs. Abnormal Returns}
\label{fig:bbandt-suntrust}
\end{figure}

\begin{table}[h!]
\centering
\caption{Abnormal Returns SunTrust}
\resizebox{0.65\textwidth}{!}{%
\begin{tabular}{lrrl}
\hline
Ticker & Raw Return & Abnormal Return & Direction \\
\hline
MTB  & 0.031114360274174226 & 0.041091210736052076 & Increased \\
RF   & 0.006489173391255033 & 0.016466023853132884 & Increased \\
FITB & 0.002245720098404834 & 0.012222570560282685 & Increased \\
CFG  & 0.030707060645199617 & 0.04068391110707747  & Increased \\
KEY  & 0.03517316578329104  & 0.04515001624516889  & Increased \\
PNC  & 0.0018759089574348529 & 0.011852759419312704 & Increased \\
PNFP & 0.02699211079993078  & 0.03696896126180863  & Increased \\
FHN  & 0.02747994594638716  & 0.037456796408265006 & Increased \\
HBAN & 0.0433482958726529   & 0.053325146334530754 & Increased \\
\hline
\end{tabular}
}
\end{table}

\begin{table}[h!]
\centering
\caption{Final Analysis SunTrust}
\resizebox{0.95\textwidth}{!}{
\begin{tabular}{r l r r r l r}
\hline
Rank & Ticker & Sim. Score & Raw Return & Abnormal Return & Direction & Shadow Signal \\
\hline
1 & RF   & 0.91 & 0.0064891733912550 & 0.0164660238531328 & Increased & 0.014984081706350849 \\
2 & KEY  & 0.88 & 0.0351731657832910 & 0.0451500162451688 & Increased & 0.03973201429574855 \\
3 & FITB & 0.87 & 0.0022457200984048 & 0.0122225705602826 & Increased & 0.010633636387445862 \\
4 & CFG  & 0.86 & 0.0307070606451996 & 0.0406839111070774 & Increased & 0.03498816355208656 \\
5 & PNC  & 0.85 & 0.0018759089574348 & 0.0118527594193127 & Increased & 0.010074845506415795 \\
7 & HBAN & 0.78 & 0.0433482958726529 & 0.0533251463345307 & Increased & 0.041593614140933946 \\
8 & MTB  & 0.74 & 0.0311143602741742 & 0.0410912107360520 & Increased & 0.03040749594467848 \\
\hline
\end{tabular}
}
\end{table}

\FloatBarrier

\subsubsection{KeyCorp / First Niagara}
KeyCorp/First Niagara is classified as strongly supporting. The top-ranked firm (CFG) shows a positive AR while all lower-ranked firms show negative reactions, producing a clear negative trendline that aligns with the shadow trading hypothesis. This is the cleanest supporting case in the Finance subsample.

\begin{figure}[h!]
\centering
\includegraphics[width=0.75\textwidth]{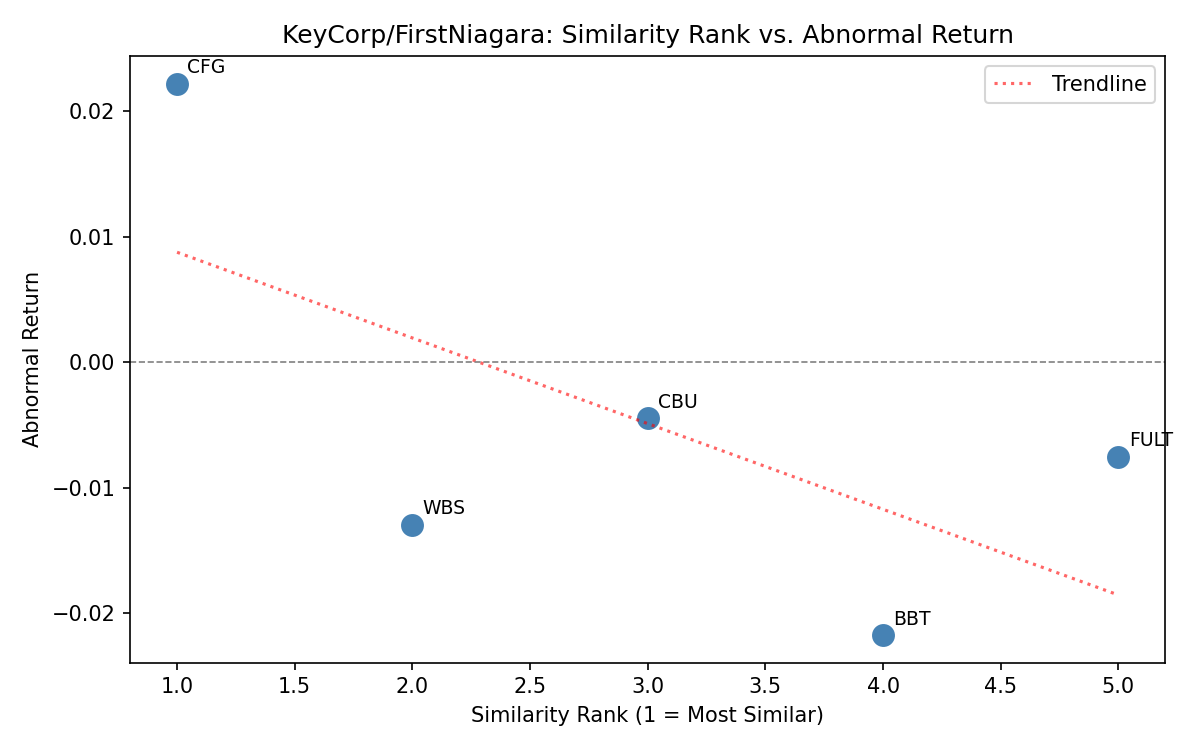}
\caption{KeyCorp / First Niagara Similarity vs. Abnormal Returns}
\label{fig:keycorp-first-niagara}
\end{figure}

\begin{table}[h!]
\centering
\caption{Abnormal Returns First Niagara}
\resizebox{0.65\textwidth}{!}{%
\begin{tabular}{lrrl}
\hline
Ticker & Raw Return & Abnormal Return & Direction \\
\hline
BBT  & -0.03606357708124112  & -0.02173759961246282  & Decreased \\
CFG  &  0.007880570014119051 &  0.02220654748289735  & Increased \\
WBS  & -0.027267811913015836 & -0.012941834444237537 & Decreased \\
FNB  & -0.023912975220883568 & -0.009586997752105269 & Decreased \\
CBU  & -0.0187772386177642   & -0.0044512611489859   & Decreased \\
VLY  & -0.018691419065318966 & -0.004365441596540667 & Decreased \\
FULT & -0.021865289671713632 & -0.007539312202935333 & Decreased \\
MTB  & -0.025768358038338268 & -0.011442380569559969 & Decreased \\
\hline
\end{tabular}
}
\end{table}

\begin{table}[h!]
\centering
\caption{Final Analysis First Niagara}
\resizebox{0.95\textwidth}{!}{
\begin{tabular}{r l r r r l r}
\hline
Rank & Ticker & Sim. Score & Raw Return & Abnormal Return & Direction & Shadow Signal \\
\hline
1 & CFG  & 0.92 &  0.0078805700141190 &  0.0222065474828973 & Increased &  0.020430023684265514 \\
2 & WBS  & 0.85 & -0.0272678119130158 & -0.0129418344442375 & Decreased & -0.011000559277601876 \\
3 & CBU  & 0.79 & -0.0187772386177642 & -0.0044512611489859 & Decreased & -0.0035164963076988608 \\
4 & BBT  & 0.76 & -0.0360635770812411 & -0.0217375996124628 & Decreased & -0.01652057570547173 \\
5 & FULT & 0.71 & -0.0218652896717136 & -0.0075393122029353 & Decreased & -0.005352911664084063 \\
\hline
\end{tabular}
}
\end{table}

\FloatBarrier

\subsubsection{People's / United}
People's/United is classified as contradicting. The trendline is positive and most peers show negative or near-zero ARs. The one strongly positive outlier (MRBK, $+4.4\%$ AR) ranks last in similarity, directly inverting the expected pattern.

\begin{figure}[h!]
\centering
\includegraphics[width=0.75\textwidth]{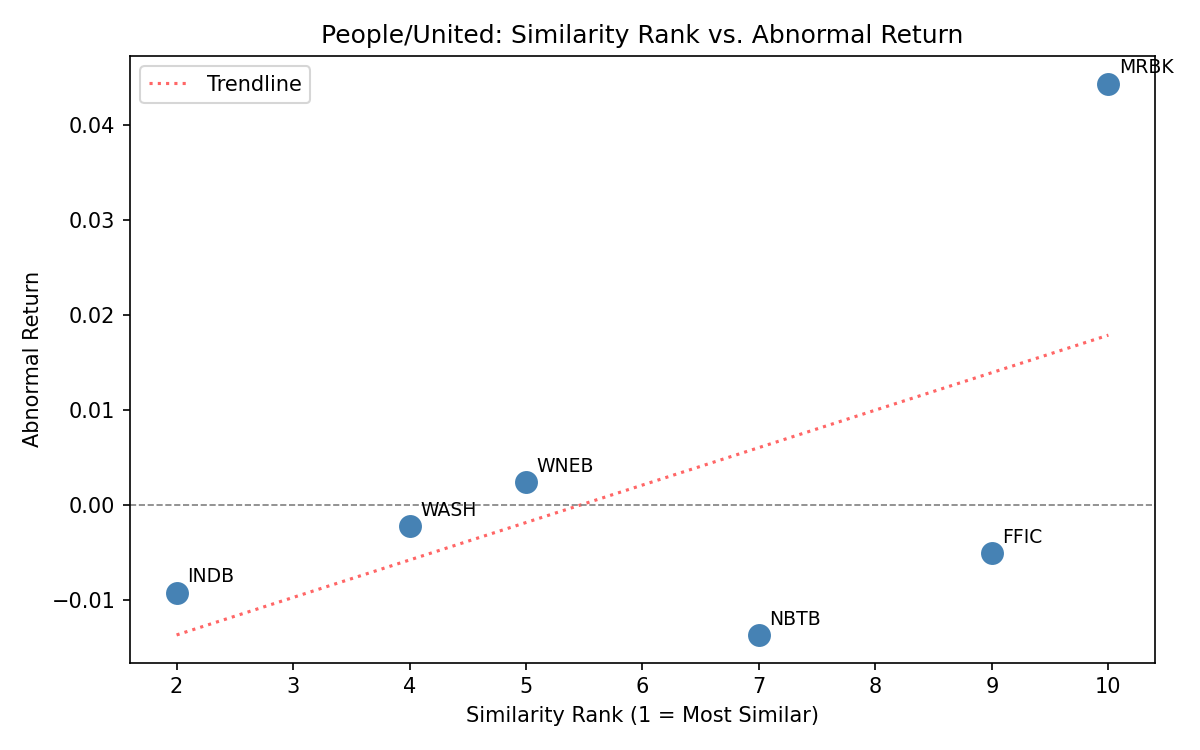}
\caption{People's / United Similarity vs. Abnormal Returns}
\label{fig:people-s-united}
\end{figure}

\begin{table}[h!]
\centering
\caption{Abnormal Returns United}
\resizebox{0.65\textwidth}{!}{%
\begin{tabular}{lrrl}
\hline
Ticker & Raw Return & Abnormal Return & Direction \\
\hline
NBTB & -0.019291550895895102 & -0.013641987240980431 & Decreased \\
WASH & -0.007836123577671349 & -0.0021865599227566777 & Decreased \\
BBT  & -0.0287973922743519   & -0.02314782861943723  & Decreased \\
MRBK &  0.038690462622618854 &  0.04434002627753353  & Increased \\
WNEB & -0.0032019992606314137 & 0.0024475643942832572 & Increased \\
FFIC & -0.010700167840185815 & -0.005050604185271144 & Decreased \\
INDB & -0.014858426109558339 & -0.009208862454643668 & Decreased \\
\hline
\end{tabular}
}
\end{table}

\begin{table}[h!]
\centering
\caption{Final Analysis United}
\resizebox{0.95\textwidth}{!}{
\begin{tabular}{r l r r r l r}
\hline
Rank & Ticker & Sim. Score & Raw Return & Abnormal Return & Direction & Shadow Signal \\
\hline
2  & INDB & 0.88 & -0.0148584261095583 & -0.0092088624546436 & Decreased & -0.008103798960086369 \\
4  & WASH & 0.81 & -0.0078361235776713 & -0.0021865599227566 & Decreased & -0.0017711135374328462 \\
5  & WNEB & 0.73 & -0.0032019992606314 &  0.0024475643942832 & Increased &  0.001786722007826736 \\
7  & NBTB & 0.66 & -0.0192915508958951 & -0.0136419872409804 & Decreased & -0.009003711579047065 \\
9  & FFIC & 0.52 & -0.0107001678401858 & -0.0050506041852711 & Decreased & -0.002626314176340972 \\
10 & MRBK & 0.42 &  0.0386904626226188 &  0.0443400262775335 & Increased &  0.01862281103656407 \\
\hline
\end{tabular}
}
\end{table}

\FloatBarrier

\subsubsection{Sterling / Astoria}
Sterling/Astoria is classified as supporting. The top-ranked firm (FFIC, $+0.79\%$ AR) shows a positive reaction while lower-ranked firms show negative reactions, producing a modest negative trendline. The magnitude of reactions is small throughout, reflecting the limited market impact of a regional bank acquisition.

\begin{figure}[h!]
\centering
\includegraphics[width=0.75\textwidth]{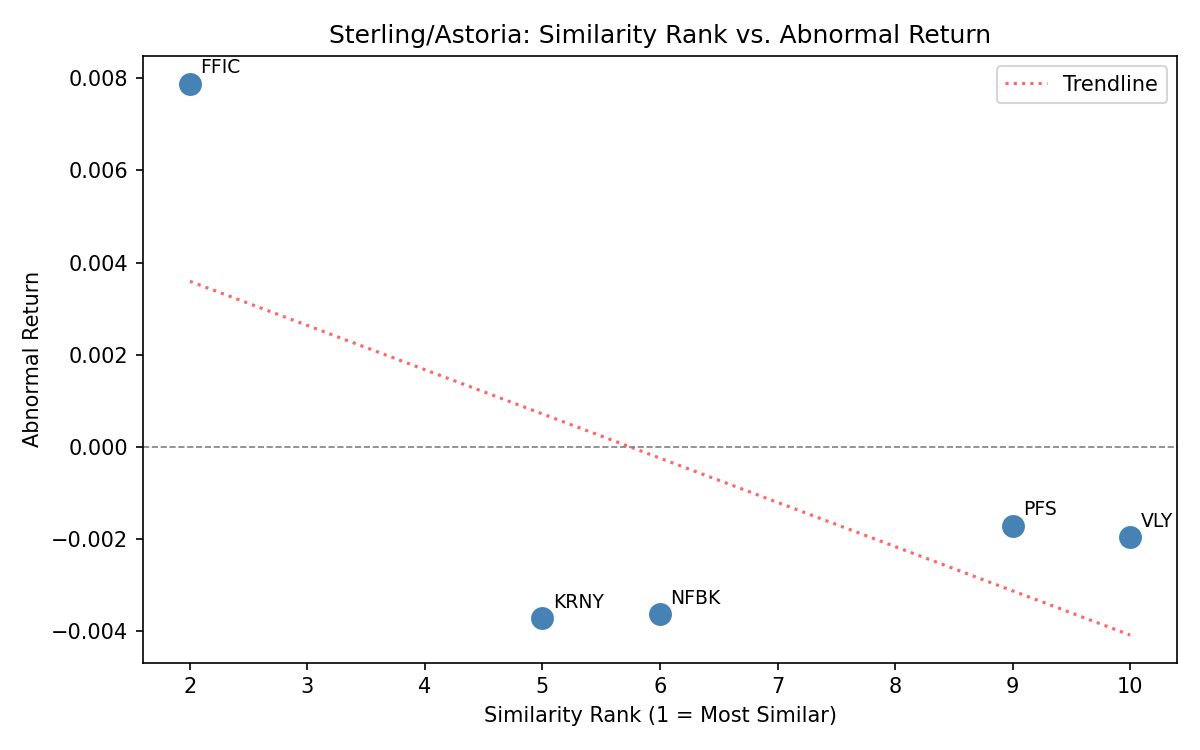}
\caption{Sterling / Astoria Similarity vs. Abnormal Returns}
\label{fig:sterling-astoria}
\end{figure}

\begin{table}[h!]
\centering
\caption{Abnormal Returns Astoria}
\resizebox{0.65\textwidth}{!}{%
\begin{tabular}{lrrl}
\hline
Ticker & Raw Return & Abnormal Return & Direction \\
\hline
VLY  & -0.004796628053415628 & -0.0019719103377265033 & Decreased \\
CNOB & -0.008080841000498168 & -0.005256123284809043  & Decreased \\
FFIC &  0.005063261483582687 &  0.00788797919927181   & Increased \\
FLG  & -0.007487942790061821 & -0.004663225074372696  & Decreased \\
NFBK & -0.006468928281572891 & -0.003644210565883767  & Decreased \\
CFG  & -0.007839123578677706 & -0.005014405862988582  & Decreased \\
DCOM &  0.002816869865432588 &  0.005641587581121712  & Increased \\
PFS  & -0.004540318237324739 & -0.0017156005216356143 & Decreased \\
KRNY & -0.006557784922778603 & -0.003733067207089479  & Decreased \\
\hline
\end{tabular}
}
\end{table}

\begin{table}[h!]
\centering
\caption{Final Analysis Astoria}
\resizebox{0.95\textwidth}{!}{
\begin{tabular}{r l r r r l r}
\hline
Rank & Ticker & Sim. Score & Raw Return & Abnormal Return & Direction & Shadow Signal \\
\hline
2  & FFIC & 0.93 &  0.0050632614835826 &  0.0078879791992718 & Increased &  0.007335820655322775 \\
5  & KRNY & 0.83 & -0.0065577849227786 & -0.0037330672070894 & Decreased & -0.003098445781884202 \\
6  & NFBK & 0.82 & -0.0064689282815728 & -0.0036442105658837 & Decreased & -0.0029882526640246337 \\
9  & PFS  & 0.77 & -0.0045403182373247 & -0.0017156005216356 & Decreased & -0.001321012401659412 \\
10 & VLY  & 0.70 & -0.0047966280534156 & -0.0019719103377265 & Decreased & -0.0013803372364085498 \\
\hline
\end{tabular}
}
\end{table}

\FloatBarrier

\subsubsection{WSFS / Beneficial}
WSFS/Beneficial is a supporting case. All peers show positive abnormal returns, and the top-ranked firm (UVSP) together with PGC (rank~3) show among the largest positive ARs. The negative trendline holds across the cohort, making this one of the more consistent supporting cases in the Finance subsample. PKBK's similarity score is recorded as $0.00$, which is a transcription error in our Stage~2 output rather than a model judgment; because PKBK ranks last under any plausible score below $0.65$, the event's rank correlation ($\rho = +0.20$) is unchanged by the error.

\begin{figure}[H]
\centering
\includegraphics[width=0.75\textwidth]{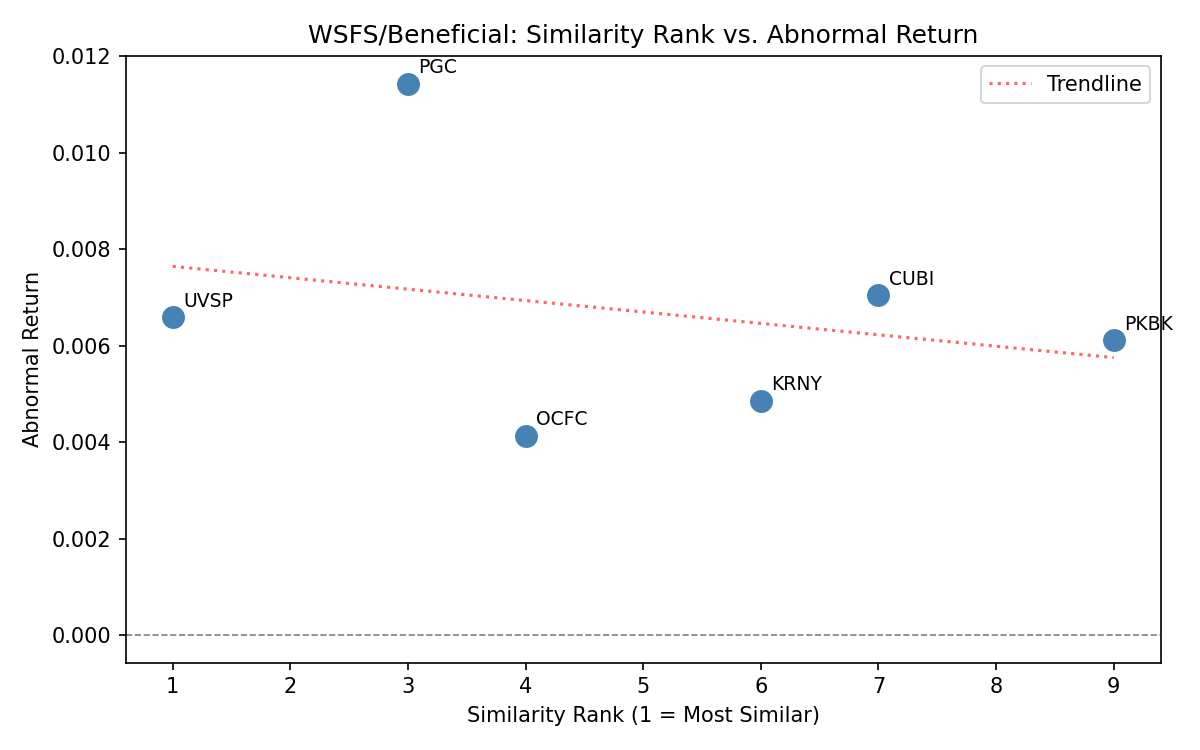}
\caption{WSFS / Beneficial Similarity vs. Abnormal Returns}
\label{fig:wsfs-beneficial}
\end{figure}

\begin{table}[H]
\centering
\caption{Abnormal Returns Beneficial}
\resizebox{0.65\textwidth}{!}{%
\begin{tabular}{lrrl}
\hline
Ticker & Raw Return & Abnormal Return & Direction \\
\hline
KRNY  &  0.007326169006966355 &  0.004861293886339441 & Increased \\
UVSP  &  0.009057941329852869 &  0.006593066209225955 & Increased \\
FRBA  &  0.017985763611004544 &  0.015520888490377631 & Increased \\
PKBK  &  0.00858361590995938  &  0.006118740789332466 & Increased \\
OCFC  &  0.006599403266681618 &  0.004134528146054704 & Increased \\
CUBI  &  0.009516558646504194 &  0.00705168352587728  & Increased \\
FRBKQ &  0.0                  & -0.002464875120626914 & Decreased \\
FULT  &  0.017192299737995555 &  0.014727424617368642 & Increased \\
WSFS  & -0.08130069889002943  & -0.08376557401065635  & Decreased \\
PGC   &  0.013901568834827368 &  0.011436693714200453 & Increased \\
\hline
\end{tabular}
}
\end{table}

\begin{table}[H]
\centering
\caption{Final Analysis Beneficial}
\resizebox{0.95\textwidth}{!}{
\begin{tabular}{r l r r r l r}
\hline
Rank & Ticker & Sim. Score & Raw Return & Abnormal Return & Direction & Shadow Signal \\
\hline
1 & UVSP & 0.94 & 0.0090579413298528 & 0.0065930662092259 & Increased & 0.006197482236672346 \\
3 & PGC  & 0.87 & 0.0139015688348273 & 0.0114366937142004 & Increased & 0.009949923531354347 \\
4 & OCFC & 0.84 & 0.0065994032666816 & 0.0041345281460547 & Increased & 0.0034730036426859475 \\
6 & KRNY & 0.70 & 0.0073261690069663 & 0.0048612938863394 & Increased & 0.00340290572043758 \\
7 & CUBI & 0.65 & 0.0095165586465041 & 0.0070516835258772 & Increased & 0.00458359429182018 \\
9 & PKBK & 0.00 & 0.0085836159099593 & 0.0061187407893324 & Increased & 0.0 \\
\hline
\end{tabular}
}
\end{table}

\FloatBarrier

\end{document}